\documentclass[10pt,twocolumn,letterpaper]{article}

\usepackage{cvpr}              
\usepackage[dvipsnames]{xcolor}

\usepackage{graphbox}
\usepackage{multirow}

\usepackage[column=0]{cellspace}
\newcommand{\underl}[1]{\underline{#1}}
\newcommand{\smallf}[2][6]{{\fontsize{#1}{20}\selectfont #2}}

\usepackage{bm}
\usepackage{pifont}
\newcommand{\cmark}{\ding{51}}
\newcommand{\xmark}{\ding{55}}

\definecolor{cvprblue}{rgb}{0.21,0.49,0.74}
\usepackage[pagebackref,breaklinks,colorlinks,citecolor=cvprblue]{hyperref}

\def\paperID{4550} 
\def\confName{CVPR}
\def\confYear{2024}

\title{SCE-MAE: Selective Correspondence Enhancement with Masked Autoencoder for Self-Supervised Landmark Estimation}

\author{
Kejia Yin\textsuperscript{1,2\dag *}\quad
Varshanth Rao\textsuperscript{2*}\quad
Ruowei Jiang\textsuperscript{2}\quad
Xudong Liu\textsuperscript{1,2\dag}\quad
Parham Aarabi\textsuperscript{1}\quad
David B. Lindell\textsuperscript{1}\\
\textsuperscript{1}University of Toronto \quad \textsuperscript{2}ModiFace}

\begin{document}
\maketitle
\def\thefootnote{*}\footnotetext{Equal contribution.}
\def\thefootnote{\dag}\footnotetext{This work was done during the internship at ModiFace.}
\begin{abstract}

Self-supervised landmark estimation is a challenging task that demands the formation of locally distinct feature representations to identify sparse facial landmarks in the absence of annotated data. To tackle this task, existing state-of-the-art (SOTA) methods (1) extract coarse features from backbones that are trained with instance-level self-supervised learning (SSL) paradigms, which neglect the dense prediction nature of the task, (2) aggregate them into memory-intensive hypercolumn formations, and (3) supervise lightweight projector networks to na\"{\i}vely establish full local correspondences among all pairs of spatial features. In this paper, we introduce SCE-MAE, a framework that (1) leverages the MAE \cite{MAE}, a region-level SSL method that naturally better suits the landmark prediction task, (2) operates on the vanilla feature map instead of on expensive hypercolumns, and (3) employs a Correspondence Approximation and Refinement Block (CARB) that utilizes a simple density peak clustering algorithm and our proposed Locality-Constrained Repellence Loss to directly hone only select local correspondences. We demonstrate through extensive experiments that SCE-MAE is highly effective and robust, outperforming existing SOTA methods by large margins of \textbf{$\sim$20\%-44\%} on the landmark matching and \textbf{$\sim$9\%-15\%} on the landmark detection tasks.

\end{abstract}
\section{Introduction}
\label{sec:intro}

Facial landmark detection is a computer vision task involving the identification and localization of specific keypoints corresponding to particular positions on a human face. Facial landmarks form the crux for many classical downstream tasks such as 3D face reconstruction \cite{3dFaceReconstruction_1, 3dFaceReconstruction_2}, face recognition \cite{FaceRecog_1, FaceRecog_2}, face emotion/expression recognition \cite{FER_1, FER_2}, and more contemporary applications such as facial beauty prediction \cite{FBP_1, FBP_2} and face make-up try on \cite{Makeup_Tryon_1, Makeup_Tryon_2, Makeup_Tryon_3}.

Albeit extremely useful, training facial landmark detectors requires numerous precise annotations per sample, making it a laborious and expensive ordeal.  Furthermore, landmarks are not always semantically well-defined, making their annotations prone to inconsistencies and biases \cite{Luvli, adnet, starloss}, which can severely limit the development of accurate landmark models. Motivated to avoid these demerits, recent works \cite{DVE, ContrastLandmark, LEAD} have incorporated the unsupervised \cite{DVE} and self-supervised learning (SSL) paradigms \cite{BYOL, MOCO, DenseCL} into their methods. SSL-pretrained models have shown to yield highly effective feature representations without the use of labeled data and, at many times, outperform their supervised counterparts on the target tasks \cite{DINO, MAE, NEURIPS2022_aa56c745}.

\begin{figure}[!t]
\centering
\includegraphics[width=1\linewidth]{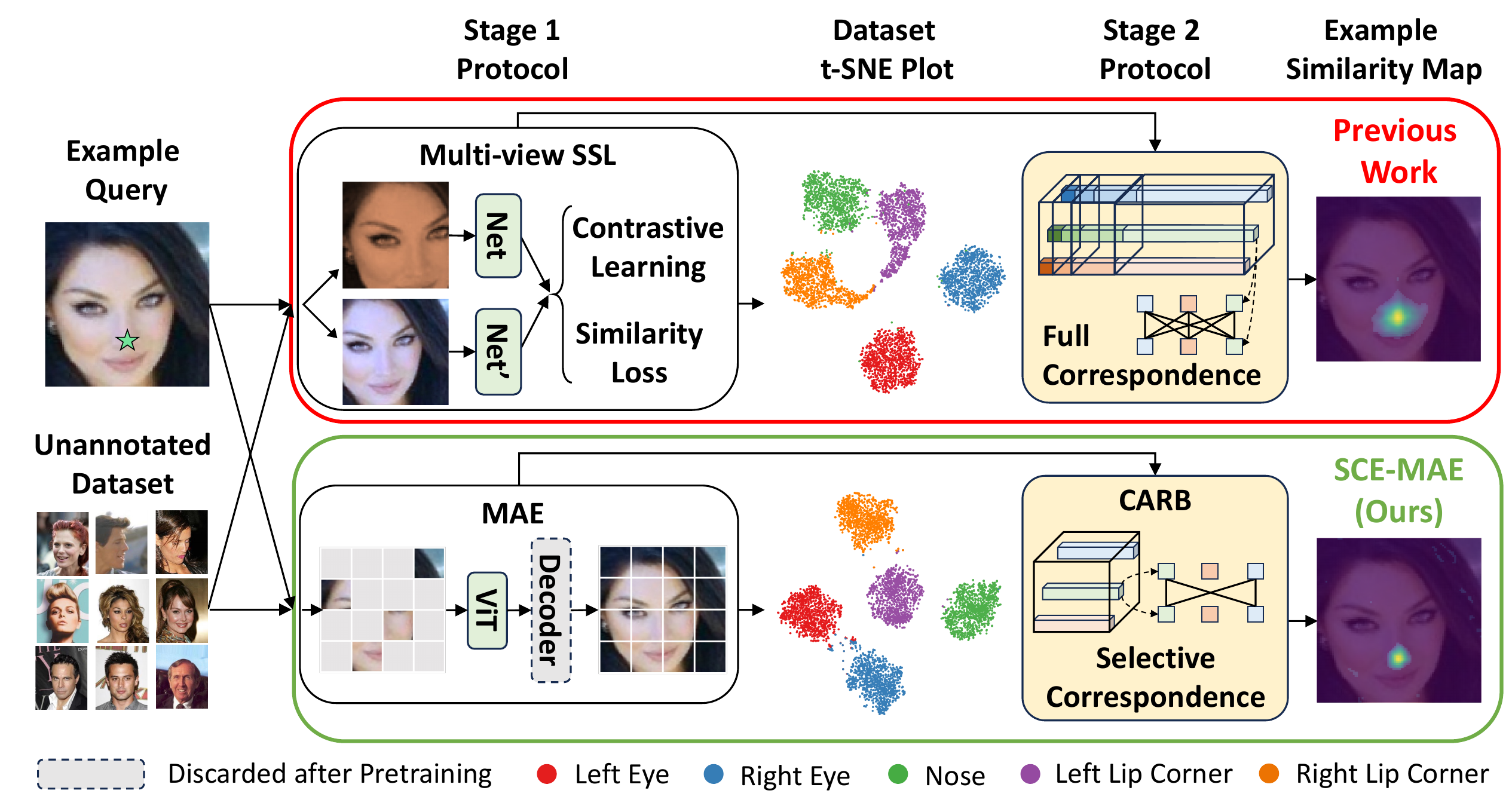}
\caption{SCE-MAE vs prior self-supervised facial landmark detection methods. \textbf{Stage 1:} Prior works (top) use instance-level multi-view SSL paradigms that output less distinct initial local features. Our framework (bottom) leverages MAE to naturally form better initial features that result in \emph{well-defined boundaries} between facial landmarks (see t-SNE plots). \textbf{Stage 2:} Prior works operate on \emph{memory-intensive hypercolumns} and supervise \emph{each feature pair} to achieve correspondence. Our framework employs a Correspondence Approximation and Refinement Block (CARB) that operates on the original MAE output and directly hones \emph{only the selected correspondence pairs}. For the example query, SCE-MAE outputs a more-focused and sharper similarity map, demonstrating the superiority of the final features.}
\label{figure1}
\end{figure}

Facial landmark detection and matching tasks rely on the formation of locally distinct features to differentiate between (1) the facial regions (e.g., eye vs.\ lip), (2) the components of face parts (e.g., left vs.\ right corners of the lip), and finally, (3) the specific pixels of each landmark. In the setting where annotations are severely limited, the recent methods \cite{LEAD, ContrastLandmark} follow a two-stage training protocol. During the first stage, the backbone is trained with a typical SSL objective. In the second stage, the backbone is frozen and a separate light-weight projector network is trained to encode \emph{local correspondences, i.e., the relationships between the different regions within the same image}.

Prior work adopted multi-view SSL protocols \cite{MOCO, BYOL}, which may be less effective on the landmark estimation tasks due to several factors. Firstly, these augment-and-compare pretext tasks prompt the network to learn category-specific signals, but we operate only on a single category, i.e., the human face. Secondly, contrastive learning requires a large and diverse set of negative samples to avoid collapse \cite{DCL, SimMoCo}. Lastly, the training objectives might not directly encourage the model to learn the intricate facial cues within the positive face samples to differentiate between facial regions, which are required for dense tasks \cite{SCRL, DenseCL} such as landmark detection and matching.

On the other hand, the Masked Image Modeling (MIM) protocol \cite{MAE, iBOT, BEiT, SimMIM}, which requires the network to reconstruct the masked regions from limited context, \emph{intrinsically} suits our downstream task objective. Based on the observation that the non-landmark regions (e.g., cheeks and foreheads) are larger and more uniform than the sparse and distinctive landmark regions (e.g., the eyes and lip corners), we hypothesize that the reconstruction of the masked landmark regions leads to the formation of effective representations of the facial landmarks. Hence, we choose to adopt the Masked Autoencoder (MAE) \cite{MAE} as our backbone in the first stage of our framework.

For the second stage, both CL \cite{ContrastLandmark} and LEAD \cite{LEAD} utilize objectives to establish correspondences between \emph{each pair} of feature descriptors within the same image. Based on the earlier observation that non-landmark regions are larger and more uniform, we ask the question: \emph{is it necessary to establish correspondences between all feature descriptor pairs?} We hypothesize that the selective refinement of the \emph{important} correspondences utilizes the network's parameters more effectively. To this end, we employ a novel Correspondence Approximation and Refinement Block. Here, we first differentiate the MAE's output into attentive (landmark and important facial regions) and inattentive (insignificant facial regions or background) tokens using the first-stage correspondence signals. Next, a clustering algorithm operates on the inattentive tokens and approximates the member tokens using the cluster center. Finally, we supervise a light-weight projector network using a novel Locality-Constrained Repellence Loss that penalizes the erroneous strong correspondences between the different token types weighted by spatial proximity. Here, only the \emph{select correspondences} are directly refined since the loss operates only on the attentive tokens and inattentive cluster center proxies. 

In order to highlight the above stage-wise merits of our approach, we visually compare, at a high level, our framework, which we term Selective Correspondence Enhancement with MAE (SCE-MAE), with prior works in Fig.\ \ref{figure1}. Our approach not only produces more distinguishable first-stage features but also outputs sharper similarity maps corresponding to the example query, testifying to superior final landmark representations.  

In this paper, we show that by leveraging MAE \cite{MAE} during the first stage and systematically eliminating redundant correspondence learning during the second stage, SCE-MAE can output locally distinct facial landmark representations without the use of labeled data. As a result, it outperforms the previous SOTA methods by large margins. We summarize our contributions below:

\begin{enumerate}
    \item We are the first to adopt an MIM-trained SSL backbone for the first-stage training of self-supervised facial landmark detection and matching methods. We demonstrate using MAE \cite{MAE} that the mask-and-predict pretext task more naturally suits the downstream objective and delivers highly potent initial landmark representations.
    
    \item We introduce the Correspondence Approximation and Refinement Block (CARB) during the second-stage to identify and approximate the features of unimportant non-landmark regions, and subsequently operate a novel Locality-Constrained Repellence (LCR) Loss to directly hone only the salient correspondences.
    \item We demonstrate the effectiveness and robustness of our framework, SCE-MAE, as it surpasses existing SOTA methods on the landmark matching (${\sim}$20\%-44\%) and detection (${\sim}$9\%-15\%) tasks under various annotation conditions for several challenging datasets.
    
\end{enumerate}
\newcommand{\Pone}[1]{\textcolor{Maroon}{#1}}
\newcommand{\Ptwo}[1]{\textcolor{BurntOrange}{#1}}
\newcommand{\Pthree}[1]{\textcolor{ForestGreen}{#1}}
\newcommand{\Pfour}[1]{\textcolor{SkyBlue}{#1}}

\section{Related Works}
\label{sec:related_works}
\textbf{Self-Supervised Learning (SSL).} By solving unique pretext tasks, SSL methods are able to learn discriminative feature representations from unlabeled data. Early works explored pretext tasks such as  predicting the rotation angle \cite{RotNet} and recovering the original image from random permuted patches \cite{Jigsaw, Jigsaw++}. Recently, invariant and contrastive learning based SSL methods \cite{BYOL, DINO, DINOv2, SimCLR, MOCO, MOCOv2, MOCOv3} have gained popularity due to their ability to capture high-level semantic concepts from the data. Invariant learning aims to learn transformation invariant features by forcing the representations of two randomly augmented views of the same image to be similar. Contrastive learning defines different views of an anchor image as positives and views of different images as negatives. Here, the objective is to pull the representations of the anchor and positives together while pushing apart those of the anchor and negatives. To achieve this, MOCO \cite{MOCO} and SimCLR \cite{SimCLR} adopted the InfoNCE \cite{InfoNCE} and NT-Xent \cite{NT-Xent} losses respectively. These methods operate at the encoded image or \emph{instance-level} and can be categorized as augment-and-compare SSL methods \cite{li2023correlational}.

Recently, the Masked Image Modeling (MIM) protocol has gained significant momentum \cite{MAE, BEiT, iBOT, SimMIM, MixMAE}. These methods operate at the \emph{region-level} and learn to recover the masked regions from the contextual information contained in the unmasked patches. It has been empirically shown that by using non-extreme masking ratios or patch sizes in Masked Autoencoders (MAE) \cite{MAE}, the representation abstractions capture robust high-level information, while extreme masking ratios capture more low-level information \cite{understandingMAE}. With higher masking ratios as the norm, MAE executes dense reconstruction, making them intrinsically suitable for dense prediction tasks \cite{MAE, MixMAE}. 

For the first stage of self-supervised face landmark detectors, ContrastLandmark (CL) \cite{ContrastLandmark} and LEAD \cite{LEAD} utilize MOCO \cite{MOCO} and BYOL \cite{BYOL} pretrained backbones respectively. Since neither MOCO nor BYOL operate explicitly at the sub-image (region/pixel) level, the representations used for the second stage of CL and LEAD are potentially sub-optimal. On the other hand, the sparse nature of facial landmarks perfectly matches the MIM objective to reconstruct the whole view from unmasked patches, resulting in higher fidelity coarse local features. Hence, in our framework, we adopt the MAE \cite{MAE} as our first stage SSL protocol.

\noindent \textbf{Unsupervised Landmark Prediction.} 
To tackle landmark prediction without annotated data, there have been several approaches. Equivalence learning leverages transformation equivalence as a free supervision signal to learn landmark embeddings \cite{Dense3D}. Since an undesirable constant vector output would satisfy the objective, adding a diversity loss or enabling similarity enforcement through intermediate auxiliary images are proposed to tackle the issue \cite{Sparse, DVE}. Another approach is through generative modeling where landmarks are discovered by training networks with a reconstruction objective \cite{DeformAE, ImGen, mallis2020unsupervised, StructuralRepr, xu2020unsupervised, lorenz2019unsupervised} such as reconstructing the human image with a different pose \cite{ImGen}.

Recent works such as ContrastLandmark (CL) \cite{ContrastLandmark} and LEAD \cite{LEAD} have adopted SSL methods to extract coarse features that capture the broad semantic concept and further process them to establish regional/local correspondences. CL and LEAD construct hypercolumns and compact them using proximity-guided and correspondence guided reduction objectives respectively. While both methods reduce the final representation size, hypercolumns are memory-wise enormous structures and operating on them is a computationally intensive process. Furthermore, each spatial feature pair is subject to the optimization objective, neglecting the possibility that some local correspondences do not contribute as much to the downstream task. On the contrary, using our SCE-MAE framework, we do not operate on expensive hypercolumns, and we identify and directly process only salient local correspondences.
 
\section{Method}
\label{sec:method}

\begin{figure*}[!t]
\centering
\includegraphics[width=0.8\linewidth]{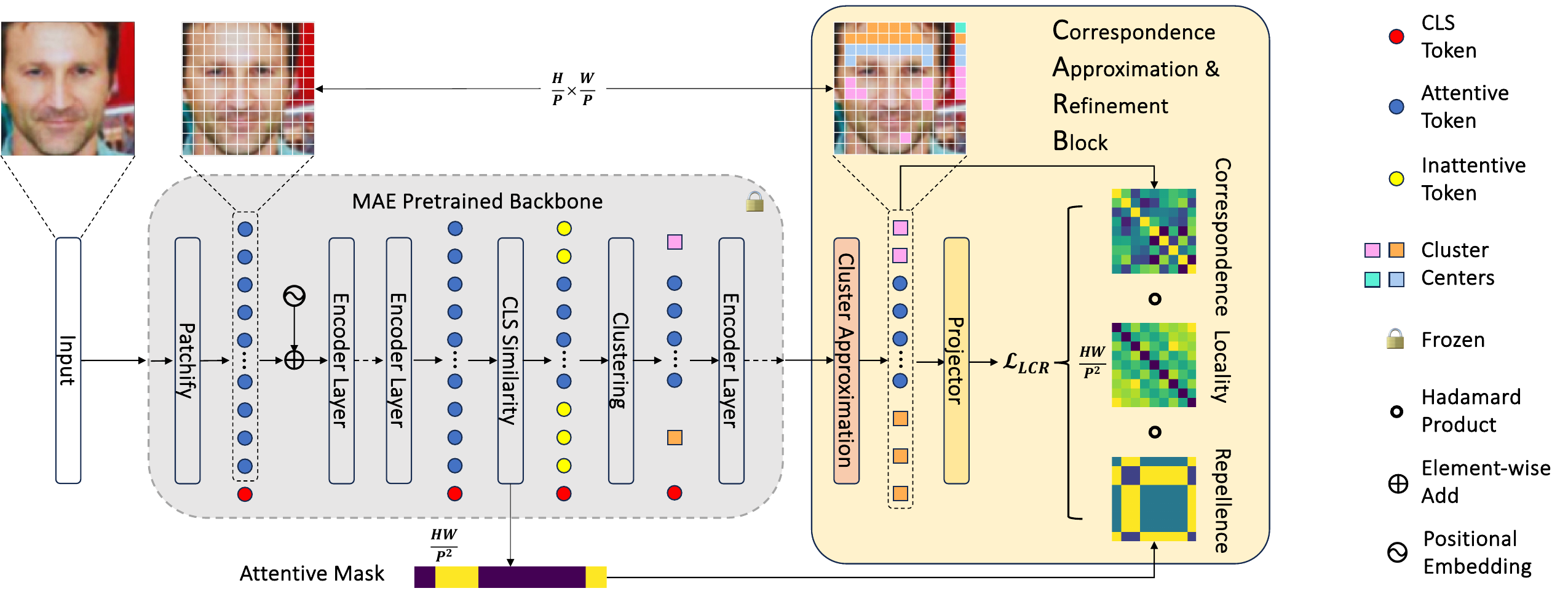}
\caption{An overview of the second stage of our proposed SCE-MAE. We first split the MAE patch tokens into attentive (blue) and inattentive (yellow) tokens based on CLS token similarity. The inattentive tokens are clustered into $K$ cluster centers. In the Correspondence Approximation and Refinement Block (CARB), we first substitute the inattentive tokens using the cluster centers (square symbols) and then refine the local features using our novel Locality-Constrained Repellence (LCR) Loss. The LCR loss weakens existing erroneous correspondences in a weighted manner by considering the token-pair proximity (locality) and correspondence type (repellence) constraints.}
\label{figure2}
\end{figure*}

We depict our proposed Selective Correspondence Enhancement with MAE (SCE-MAE) framework in Fig.\ \ref{figure2} and detail each component in the following subsections. In Sec.\ \ref{ssec:mim} we revisit Masked Image Modeling to introduce the MAE \cite{MAE} as a more suitable and potent first stage protocol. In Sec.\ \ref{ssec:sel_corr_setup}, we elaborate on the setup to execute selective correspondence through the process of reducing the effective number of final correspondence pairs. In Sec.\ \ref{ssec:carb}, we introduce our Correspondence Approximation and Refinement Block, wherein we explain the components of our novel Locality-Constrained Repellence Loss and how it directly hones only the selected correspondences.  

\subsection{A Revisit of Masked Image Modeling} \label{ssec:mim}
Masked Image Modeling (MIM) \cite{MAE, SimMIM, BEiT, iBOT} is an SSL paradigm that involves the reconstruction of the original image from the unmasked patches. Taking MAE \cite{MAE} as an example, given an input image $x$, the encoder first divides the image into non-overlapping patches $x^p$ with positional embedding added to them. A class token is appended to the patch tokens but will not be affected by the following masking procedure. A binary mask $M$ is randomly sampled to determine the masked out regions. The unmasked patches are denoted by $\hat{x^p} = x^p \circ M$, where $\circ$ symbolizes the Hadamard product, and are processed by the encoder to output the patch embeddings $\hat{f^p}$. Finally, MAE uses a special embedding $[\mathbb{MASK}]$ to fill in the masked positions, $f^p = \hat{f^p} + [\mathbb{MASK}] \circ (1 - M)$, and reconstruct $x$ from $f^p$ by minimizing the pixel-level mean squared error via a light-weight decoder. The reconstruction task requires the network to capitalize on the limited semantic context provided by the unmasked patches and the supplied positional information. This encourages the network to forge discriminative features that are optimal for differentiating and localizing the important landmark regions.

\subsection{Setup for Selective Correspondence} \label{ssec:sel_corr_setup}
\textbf{Attentive-Inattentive Separation.}
The second stage of the framework aims to establish local correspondences effectively to ensure that the representations reflect the extent of similarity and dissimilarity between the different facial regions. To achieve this, we propose to execute \emph{selective correspondence}, i.e., the elimination of the direct refinement of unimportant non-landmark correspondences, and focus on optimizing those that are critical for landmark disambiguation. The first step in this endeavor is to identify potential landmark and non-landmark regions. Due to the observable opposing nature of facial landmarks (sparse and distinct) and non-landmark regions (dense and uniform), we hypothesize that they are coarsely distinguishable using the first stage backbone features.

Following MAE \cite{MAE}, we adopt the ViT \cite{ViT} as our backbone architecture. The class (CLS) token represents the image and is obtained by aggregating information from the other patch tokens over several layers. Since landmarks are sparse and have more distinct texture, we expect the corresponding tokens to have a large influence on the CLS token representation. After pretraining, we compute a similarity vector between the CLS token and all patch tokens as:
\begin{align}
    \label{eq_attn}
    Sim_{cls} = \text {Softmax}(\frac{K \cdot q_{cls}}{\sqrt{d}}) \in \mathbb{R}^N,
\end{align}
where $q_{cls}$, $K$, $d$, and $N$ denote the CLS token query vector, the patch token key matrix, latent dimension, and number of patch tokens respectively. Here, $q_{cls}\in \mathbb{R}^d$ and $K\in \mathbb{R}^{N\times d}$. We then split the $N$ patch tokens into two groups: (1) attentive group, consisting of the $\eta\cdot N$ tokens that have the highest similarity score with the CLS token, and (2) inattentive group, consisting of the remaining $(1-\eta)\cdot N$ tokens. Here, $\eta$ is a hyperparameter between 0 and 1. We observe that the inattentive tokens mostly cover non-landmark face regions (See Fig.\ \ref{figure2}), such as cheeks and forehead, as well as background. Henceforth, we presume that attentive tokens cover the landmark and important facial regions, while inattentive tokens correspond to unimportant non-landmark regions. 

\noindent \textbf{Inattentive Token Clustering.}
Since several inattentive tokens often correspond to the same facial region (e.g., cheek, forehead, etc.), the downstream correspondence objectives associated with them would likely be redundant. By applying a clustering algorithm on the inattentive tokens, we can represent numerous non-landmark regions with only a handful of cluster centers. Selective correspondence can then be set up by discarding all non-cluster center tokens, ensuring that no correspondence is established with them. Specifically, we adopt a simple density peak clustering algorithm \cite{long2023beyond}, wherein two variables $\rho$ and $\delta$ are defined for each inattentive token. Here, $\rho_i$ measures the density of the $i$-th token and $\delta_i$ computes the minimum distance from the $i$-th token to any other inattentive token which has a higher density. Mathematically, they are defined as:
\begin{equation}
    \label{eq_rho}
    \rho_i = \text{exp} \left(\sum_{t_j \in \text{T$_{inatt}$}} \| t_i - t_j \|^2_2\right) ,
\end{equation}
\begin{equation}
    \label{eq_delta}
    \delta_i = \left\{\begin{array}{rl}
        \text{min}_{j:\rho_j > \rho_i} \| t_i - t_j \|_2, & \text{if} \;\exists\; j\; \text{s.t.}\; \rho_j > \rho_i \\
        \text{max}_{j} \| t_i - t_j \|_2, & \text{otherwise}
    \end{array}\right. ,
\end{equation}
where $t_i, t_j \in \text{T$_{inatt}$}$, and $\text{T$_{inatt}$}$ denotes all inattentive tokens. Since the cluster center should have higher density than neighbouring tokens and should also be distant to other cluster centers, the cluster center score of the $i$-th token is computed by $\rho_i \cdot \delta_i$. We select the top-$K_{c}$ scoring tokens as cluster centers, where $K_{c}$ is a hyperparameter. The remaining inattentive tokens are discarded and the cluster center tokens subsequently act as representative proxies for them.

\subsection{Selective Correspondence using CARB} \label{ssec:carb}
In our Correspondence Approximation and Refinement Block (CARB), we first substitute the discarded inattentive tokens with their corresponding cluster centers and aggregate the relevant visual features to obtain a complete 2D feature map as illustrated in Fig.\ \ref{figure2}. With the backbone frozen, the feature map is passed through a light-weight projector, which is supervised by our novel Locality-Constrained Repellence (LCR) Loss. As the LCR loss operates on the features of attentive tokens and inattentive cluster centers, we directly refine only the most important correspondences, thereby achieving \emph{selective correspondence}.

\noindent\textbf{Locality-Constrained Repellence (LCR) Loss.} We design and operate the LCR loss to yield high-fidelity fine-grained features by optimally refining local correspondences. Henceforth, we use T$_{att}$ and T$_{inatt'}$ to denote the attentive and the approximated inattentive tokens (cluster centers) respectively, and define T$ = \text{T$_{att}$} \cup \text{T$_{inatt'}$}$, as the set of all considered tokens.

We begin by formally defining correspondence, i.e., the probability that a patch token $t_j$ corresponds to a patch token $t_i$ in the image $x$, which is expressed as:
\begin{equation}
    \label{eq_corr}
    p(t_j | t_i; \Phi, x) = \frac{\text{exp}(\langle\Phi_{t_i}(x),\Phi_{t_j}(x)/\tau\rangle)}{\sum_{t_k \in \text{T}}\text{exp}(\langle\Phi_{t_i}(x),\Phi_{t_k}(x)/\tau\rangle)},
\end{equation}
where $\Phi_{t_i}(x)$ is the final projected feature representation of patch $t_i$, and $\tau$ is the temperature parameter.

We observe that image patches that are spatially distant from each other often correspond to different facial regions. Hence, it should follow that strong correspondences between distant patches are likely to be erroneous and should be discouraged. We compute a locality constraint to formalize this idea using the following function:
\begin{equation}
    \label{eq_f_loc}
    f_{loc}(t_i, t_j) = \text{log} (\|t_i - t_j\| + 1),
\end{equation}
where $t_i, t_j \in \text{T}$, and $\|\cdot\|$ computes the spatial distance. The $\rm log$ function saturates the coefficient in order to discourage the network from excessively focusing on separating very distant correspondences. Although a similar constraint was introduced in \cite{Dense3D}, the primary motive was to avoid collapse during equivalence learning.

Considering the attentive (T$_{att}$) and the approximated inattentive (T$_{inatt'}$) token sets, there are three types of correspondences: attentive-attentive ($att-att$), attentive-inattentive ($att-inatt$), and inattentive-inattentive ($inatt-inatt$). We introduce a repellence coefficient to quantify the importance of each correspondence type:
\begin{equation}
    \label{eq_f_rep}
    \lambda_{rep}(t_i, t_j) = \left\{\begin{array}{rl}
         r_{att-att}, & \text{if}\; t_i,t_j \in \text{T$_{att}$} \\
         r_{inatt-inatt}, & \text{if}\; t_i,t_j \in \text{T$_{inatt'}$} \\
         r_{att-inatt}, & \text{otherwise}
    \end{array}\right. ,
\end{equation}
where each coefficient $r$ is a hyperparameter. In practice, we set $r_{att-att}$ and $r_{att-inatt}$ to be higher than $r_{inatt-inatt}$ since we aim to prioritize facial landmark differentiation and landmark vs non-landmark disambiguation over non-landmark differentiation respectively.

Combining all of the above defined components, we mathematically express the LCR loss as:
\begin{equation}
    \label{eq_l_lcr}
    \mathcal{L}_{LCR} = \sum_{t_i \in \text{T}}\sum_{t_j \in \text{T}} f_{loc}(t_i, t_j)\cdot \lambda_{rep}(t_i, t_j) \cdot p(t_j | t_i; \Phi, x),
\end{equation}

The LCR loss aims to forge effective local features by systematically weakening erroneous, spatially distant correspondences among and between the important and unimportant patch tokens. 

\noindent\textbf{Inference.}
After training, we obtain optimized representations for all image regions. Hence, during inference, we bypass the clustering and inattentive token approximation procedures and only utilize the original features for the downstream tasks. Additionally, we require to spatially expand the size of the feature input to the projector for fair comparison against prior works. Na\"{\i}vely decreasing the patch size to expand the output not only quadratically increases the computation and memory costs but also may lead to the formation of inferior feature representations \cite{SimMIM, understandingMAE}. Instead, we adopt the cover-and-stride technique \cite{xie2023maester} to produce more fine-grained and rich expanded representations.

\begin{figure}[!t]
\begin{tabular}{cccc}
      \includegraphics[width=0.2\linewidth, height=0.2\linewidth]{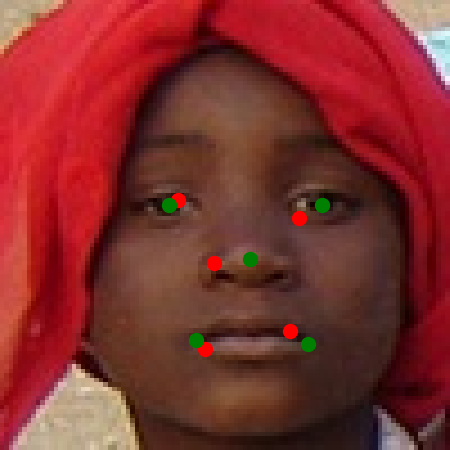}
    & \includegraphics[width=0.2\linewidth, height=0.2\linewidth]{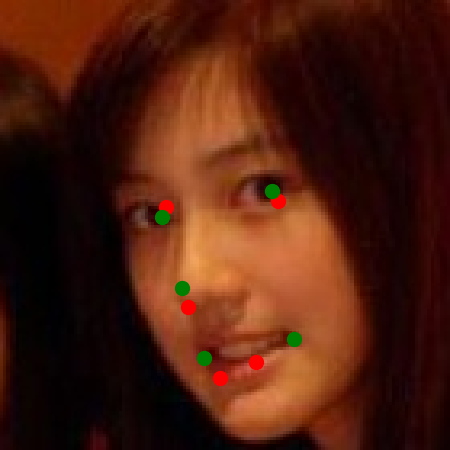}
    & \includegraphics[width=0.2\linewidth, height=0.2\linewidth]{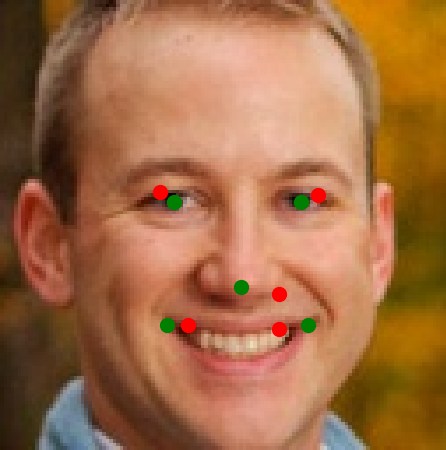}
    & \includegraphics[width=0.2\linewidth, height=0.2\linewidth]{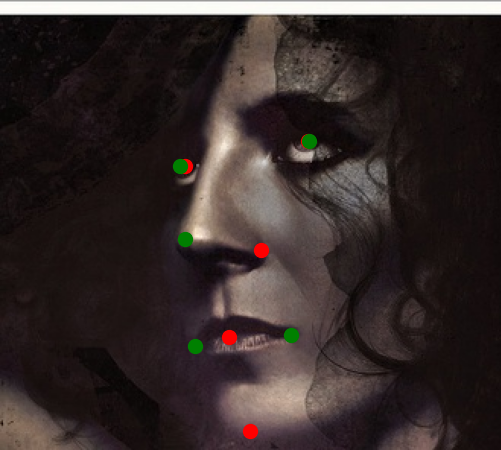} \\
      \includegraphics[width=0.2\linewidth, height=0.2\linewidth]{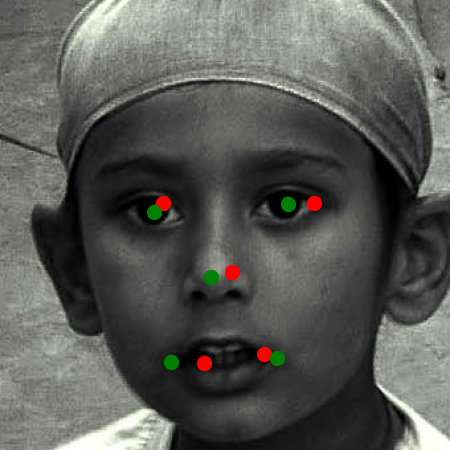}
    & \includegraphics[width=0.2\linewidth, height=0.2\linewidth]{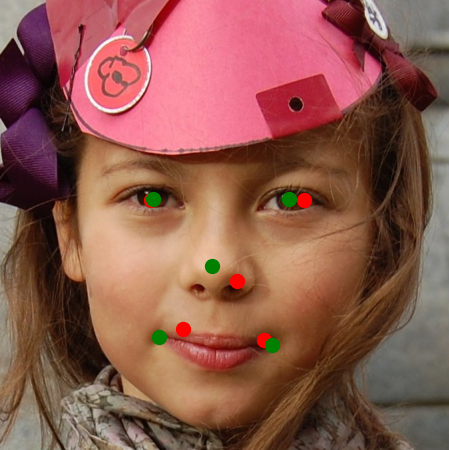}
    & \includegraphics[width=0.2\linewidth, height=0.2\linewidth]{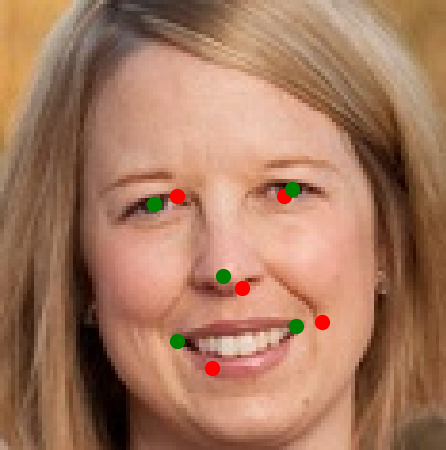}
    & \includegraphics[width=0.2\linewidth, height=0.2\linewidth]{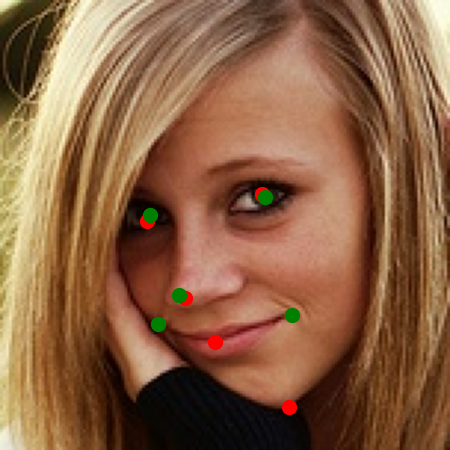} \\
    
\end{tabular}
\centering
\caption{Comparison between the original (red) and re-annotated (green) landmarks in AFLW$_R$ test set. We denote the original and corrected test sets as AFLW$_{RO}$ and AFLW$_{RC}$ respectively.}
\label{Figure AFLW re-annotation}
\end{figure}
\section{Experiments}
\label{sec:experiments}
\renewcommand{\arraystretch}{1.02}

\begin{figure*}[!t]
\begin{tabular}{p{0.33in}|ccc|ccc|ccc}
    Ref
    & \includegraphics[align=c, width=0.08\linewidth, height=0.08\linewidth]{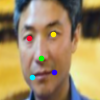}
    & \includegraphics[align=c, width=0.08\linewidth, height=0.08\linewidth]{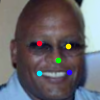}
    & \includegraphics[align=c, width=0.08\linewidth, height=0.08\linewidth]{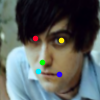}
    & \includegraphics[align=c, width=0.08\linewidth, height=0.08\linewidth]{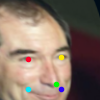}
    & \includegraphics[align=c, width=0.08\linewidth, height=0.08\linewidth]{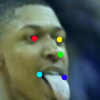}
    & \includegraphics[align=c, width=0.08\linewidth, height=0.08\linewidth]{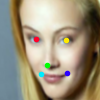}
    & \includegraphics[align=c, width=0.08\linewidth, height=0.08\linewidth]{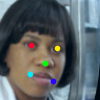}
    & \includegraphics[align=c, width=0.08\linewidth, height=0.08\linewidth]{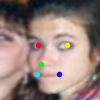}
    & \includegraphics[align=c, width=0.08\linewidth, height=0.08\linewidth]{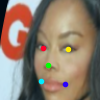} \rule[5.5ex]{0pt}{0pt}\rule[-4.4ex]{0pt}{0pt}\\
    \hline
    CL
    & \includegraphics[align=c, width=0.08\linewidth, height=0.08\linewidth]{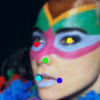}
    & \includegraphics[align=c, width=0.08\linewidth, height=0.08\linewidth]{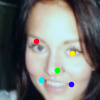}
    & \includegraphics[align=c, width=0.08\linewidth, height=0.08\linewidth]{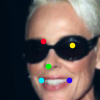}
    & \includegraphics[align=c, width=0.08\linewidth, height=0.08\linewidth]{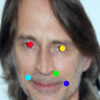}
    & \includegraphics[align=c, width=0.08\linewidth, height=0.08\linewidth]{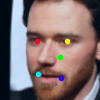}
    & \includegraphics[align=c, width=0.08\linewidth, height=0.08\linewidth]{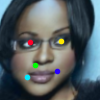}
    & \includegraphics[align=c, width=0.08\linewidth, height=0.08\linewidth]{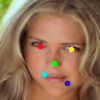}
    & \includegraphics[align=c, width=0.08\linewidth, height=0.08\linewidth]{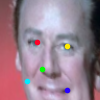}
    & \includegraphics[align=c, width=0.08\linewidth, height=0.08\linewidth]{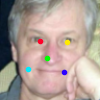} \rule[5.5ex]{0pt}{0pt}\rule[-4.4ex]{0pt}{0pt}\\
    LEAD
    & \includegraphics[align=c, width=0.08\linewidth, height=0.08\linewidth]{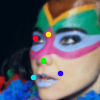}
    & \includegraphics[align=c, width=0.08\linewidth, height=0.08\linewidth]{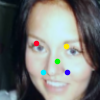}
    & \includegraphics[align=c, width=0.08\linewidth, height=0.08\linewidth]{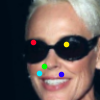}
    & \includegraphics[align=c, width=0.08\linewidth, height=0.08\linewidth]{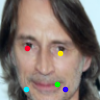}
    & \includegraphics[align=c, width=0.08\linewidth, height=0.08\linewidth]{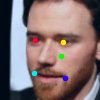}
    & \includegraphics[align=c, width=0.08\linewidth, height=0.08\linewidth]{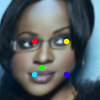}
    & \includegraphics[align=c, width=0.08\linewidth, height=0.08\linewidth]{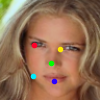}
    & \includegraphics[align=c, width=0.08\linewidth, height=0.08\linewidth]{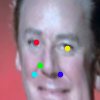}
    & \includegraphics[align=c, width=0.08\linewidth, height=0.08\linewidth]{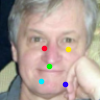} \rule[5.5ex]{0pt}{0pt}\rule[-4.4ex]{0pt}{0pt}\\
    Ours \smallf{(DeiT-S)}
    & \includegraphics[align=c, width=0.08\linewidth, height=0.08\linewidth]{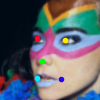}
    & \includegraphics[align=c, width=0.08\linewidth, height=0.08\linewidth]{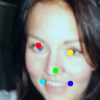}
    & \includegraphics[align=c, width=0.08\linewidth, height=0.08\linewidth]{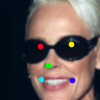}
    & \includegraphics[align=c, width=0.08\linewidth, height=0.08\linewidth]{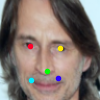}
    & \includegraphics[align=c, width=0.08\linewidth, height=0.08\linewidth]{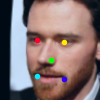}
    & \includegraphics[align=c, width=0.08\linewidth, height=0.08\linewidth]{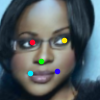}
    & \includegraphics[align=c, width=0.08\linewidth, height=0.08\linewidth]{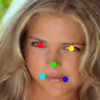}
    & \includegraphics[align=c, width=0.08\linewidth, height=0.08\linewidth]{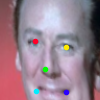}
    & \includegraphics[align=c, width=0.08\linewidth, height=0.08\linewidth]{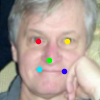} \rule[5.5ex]{0pt}{0pt}\rule[-4.4ex]{0pt}{0pt}\\
    \hline
    GT
    & \includegraphics[align=c, width=0.08\linewidth, height=0.08\linewidth]{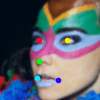}
    & \includegraphics[align=c, width=0.08\linewidth, height=0.08\linewidth]{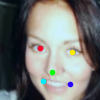}
    & \includegraphics[align=c, width=0.08\linewidth, height=0.08\linewidth]{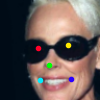}
    & \includegraphics[align=c, width=0.08\linewidth, height=0.08\linewidth]{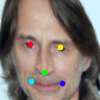}
    & \includegraphics[align=c, width=0.08\linewidth, height=0.08\linewidth]{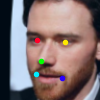}
    & \includegraphics[align=c, width=0.08\linewidth, height=0.08\linewidth]{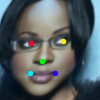}
    & \includegraphics[align=c, width=0.08\linewidth, height=0.08\linewidth]{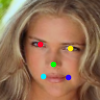}
    & \includegraphics[align=c, width=0.08\linewidth, height=0.08\linewidth]{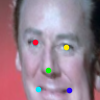}
    & \includegraphics[align=c, width=0.08\linewidth, height=0.08\linewidth]{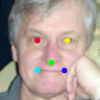} \rule[5.5ex]{0pt}{0pt}\rule[-4.4ex]{0pt}{0pt}\\
\end{tabular}
\centering
\caption{\textbf{Qualitative results on landmark matching.} The reference/ground-truth are shown in the top/bottom row. The middle rows show the matching results of our method and prior works, grouped column-wise by errors occurring with the eyes, nose and lip corner landmarks respectively. Our method outputs consistently more accurate matching resulting from leveraging higher fidelity projected features.}
\label{Figure Matching qualitative}
\end{figure*}

\textbf{Datasets.} Following prior works \cite{DVE, ContrastLandmark, LEAD}, we pretrain our backbone on the CelebA \cite{CelebA} dataset, which contains 162,770 images. Face landmark detection is evaluated on four datasets: MAFL \cite{MAFL}, 300W \cite{300W} and two variants of AFLW \cite{AFLW}. MAFL consists of 19,000 training images and 1,000 test images. 300W has 3148 training images and 689 test images. AFLW$_M$ contains 10,122 training images and 2995 testing images, which are crops from MTFL\cite{MTFL}. AFLW$_R$ contains tighter crops of face images where the training and test set has 10,122 and 2,991 images respectively. Note that 300W provides 68 annotations per image while the other three datasets only provides 5 annotations.

\noindent\textbf{Re-annotation.} Although the AFLW$_R$ has been used in prior works, the fidelity of the annotations are questionable. In Fig.\ \ref{Figure AFLW re-annotation}, we visualize several annotation errors in the AFLW$_R$ test set using red dots. These include errors arising due to semantic mismatches, translations, and random shifts. For a more consistent and reliable evaluation, we re-annotate the AFLW$_R$ test set and illustrate a few of these corrections using green dots in Fig.\ \ref{Figure AFLW re-annotation}. In the following sections, we use AFLW$_{RO}$ and AFLW$_{RC}$ to denote the original and corrected dataset respectively.

\noindent\textbf{Implementation Details.} 
We pretrain our models on the CelebA dataset using MAE with three backbones: DeiT-T, DeiT-S and DeiT-B. All models were trained for 400 epochs with a batch size of 512, a learning rate of 3e-4 and patch size of 8. Following DVE \cite{DVE}, we resize the image to 136$\times$136 and crop the center 96$\times$96 as input for both landmark matching and regression. We set the attentive rate $\eta$ to 0.25 for DeiT-B and 0.1 for DeiT-T and DeiT-S. We apply clustering after the third encoder layer and set the number of clusters $K_{c}$ to 4. For the LCR loss, the three repellence hyperparameters are set to $r_{attn-attn}=5, r_{attn-inattn}=5, r_{inattn-inattn}=2$. Ablation studies on the various hyperparameters are included in the \emph{Supplementary Material}.

\begin{table}[!t]
\caption{\textbf{Quantitative evaluations on landmark matching.} We report the mean pixel error between the prediction and ground-truth on 1000 image pairs sampled from MAFL. The best and second best results are shown in \textbf{bold} and \underl{underline} respectively. We group the results by the projected feature dimension. Our method outperforms all prior works by large margins within each group for both the same and different identity settings.}
\centering
\label{table matching}
\begin{tabular}{l|cc|cc}
\hline
\multirow{2}{*}{\textbf{Method}} & \textbf{\#Param.} & \textbf{Feat.}    & \textbf{Same}         & \textbf{Diff.}      \rule[2.5ex]{0pt}{0pt}\\
                            & \textbf{Millions}     & \textbf{Dim.}     & \multicolumn{2}{c}{\textbf{Mean Pixel Error$\downarrow$}} \rule[-1ex]{0pt}{0pt}\\
\hline
DVE\cite{DVE}               & 12.4                  & 64                & 0.92                  & \underl{2.38}          \\
CL\cite{ContrastLandmark}   & 23.8                  & 64                & 0.92                  & 2.62          \\
LEAD\cite{LEAD}             & 23.8                  & 64                & \underl{0.51}         & 2.60          \\
Ours \smallf[8]{DeiT-T}     & 5.4                   & 64                & \textbf{0.47}         & \textbf{1.99}          \\
\hline
CL\cite{ContrastLandmark}   & 23.8                  & 128               & \underl{0.82}         & \underl{2.19}          \\
Ours \smallf[8]{DeiT-S}     & 21.4                  & 128               & \textbf{0.31}         & \textbf{1.69} \\
\hline
CL\cite{ContrastLandmark}   & 23.8                  & 256               & 0.71                  & 2.06          \\
LEAD\cite{LEAD}             & 23.8                  & 256               & 0.48                  & 2.50          \\
Ours \smallf[8]{DeiT-S}     & 21.4                  & 256               & \underl{0.33}         & \underl{1.72} \\
Ours \smallf[8]{DeiT-B}     & 85.3                  & 256               & \textbf{0.27}         & \textbf{1.61} \\
\hline
\end{tabular}
\end{table}

\begin{table*}[!t]
\caption{\textbf{Quantitative evaluations on landmark detection with all annotated samples.} We compare our method with existing SOTA and report the error as the percentage of inter-ocular distance on four human face datasets: MAFL, AFLW$_M$, AFLW$_R$ and 300W. For AFLW$_R$, we report the results on both the original (AFLW$_{RO}$) and corrected (AFLW$_{RC}$) datasets. Our method, despite using significantly smaller features by avoiding expensive hypercolumns, outperforms prior works on all four datasets, even with our smallest backbone, DeiT-T.}
\centering
\label{table detection}
\begin{tabular}{l|ccc|ccccc}
\hline
\multirow{2}{*}{\textbf{Method}} & \textbf{\#Params.} & \textbf{Feature} & \textbf{Hypercol.}& \textbf{MAFL} & \textbf{AFLW$_M$} & \textbf{AFLW$_{RO}$} & \textbf{AFLW$_{RC}$}   & \textbf{300W} \rule[2.5ex]{0pt}{0pt}\\
                            & \textbf{Millions}  & \textbf{Dim.}    & \textbf{Used} & \multicolumn{5}{c}{\textbf{Inter-ocular Distance (\%)$\downarrow$}} \rule[-1ex]{0pt}{0pt}\\
\hline
DVE\cite{DVE}               & 12.6      & 64        & \xmark & 2.76          & 6.96          & 6.33          & 5.58          & 4.58    \\
CL\cite{ContrastLandmark}   & 23.8      & 3840      & \cmark & 2.76          & 6.17          & 5.69          & 5.06          & 4.84    \\
LEAD\cite{LEAD}             & 23.8      & 3840      & \cmark & 2.44          & 6.05          & 5.71          & 5.11          & 4.87    \\
\hline
Ours DeiT-T                 & 5.4       & 256       & \xmark & 2.20          & 5.89          & 5.54          & 4.86          & 4.22    \\
Ours DeiT-S                 & 21.4      & 512       & \xmark & \underl{2.08} & \underl{5.33} & \underl{5.40} & \underl{4.69} & \textbf{3.94}    \\
Ours DeiT-B                 & 85.3      & 1024      & \xmark & \textbf{2.07} & \textbf{5.23} & \textbf{5.33} & \textbf{4.60} & \underl{3.95}    \\
\hline
\end{tabular}
\end{table*}

\subsection{Landmark Matching} \label{exp: landmark matching}

\textbf{Evaluation Protocol.} Following \cite{DVE}, 1000 reference-and-test image pairs are generated from MAFL test set for evaluation. The first 500 pairs serve as the benchmark for landmark matching between same identities, which contains the original image and its thin-plate-spline (TPS) deformed counterpart. The other 500 pairs are of different identities. During evaluation, all feature maps are bi-linearly up-sampled to the image resolution. Landmark representations of the reference image are used to query the test image. The location with the highest cosine similarity is considered as the matched prediction. Finally, we compute the Mean Pixel Error between the prediction and ground-truth.

\noindent\textbf{Quantitative Results.} We compare our method with existing SOTA methods in Table \ref{table matching} by grouping the results based on the final feature size. We use three different backbones to control the number of parameters for a fair comparison. In the first group, our model with DeiT-T, being \emph{a fraction of the size of prior works}, already outperforms the SOTA. In the second and third group, our method visibly outperforms prior works by large margins of \emph{$\sim$20\% and $\sim$44\%} for the same and different identities respectively. We attribute this to the highly potent initial features from the MAE pretraining, which, when strategically refined through selective correspondence using CARB, yields distinctive final features that were vital for successful landmark matching.

\noindent\textbf{Qualitative Results.} We visualize our landmark matching results between different identities and compare our results with existing SOTA methods in Fig.\ \ref{Figure Matching qualitative}. The mismatches on different landmarks when using CL \cite{ContrastLandmark} and LEAD \cite{LEAD} are shown in different columnar groups, e.g., the first group of three columns contain eye-related mismatches. Our method clearly achieves a more accurate matching performance across all landmarks even on difficult examples such as those wearing eye-glasses. Admittedly, our method experiences some failures when the poses of reference and test samples are vastly dissimilar or when landmark regions are severely occluded. Some of the failure cases are shown in the \emph{Supplementary Material}.

\begin{table*}[!t]
\caption{\textbf{Quantitative evaluations on landmark detection with limited annotated samples.} We compare our method with existing SOTA under different annotation settings on the AFLW$_{M}$ dataset and report the error as the percentage of inter-ocular distance. Our method proves to be more effective by offering notably lower errors and more robust by yielding a lower std of errors than prior works.}
\centering
\small
\label{table detection limitted annotation}
\begin{tabular}{ll|cccccc}
\hline
\multirow{2}{*}{\textbf{Method}} & \textbf{Feat.} & \multicolumn{6}{c}{\textbf{Number of Annotated Samples}} \rule[2.5ex]{0pt}{0pt}\rule[-1ex]{0pt}{0pt}\\
\cline{3-8}
                    & \textbf{Dim.} & \textbf{1}                    & \textbf{5}                    & \textbf{10} 
                                    & \textbf{20}                   & \textbf{50}                   & \textbf{100} \rule[2.5ex]{0pt}{0pt}\rule[-1ex]{0pt}{0pt}\\
\hline
DVE\cite{DVE}               & 64    & \textbf{14.23 $\pm$ 1.45}     & 12.04 $\pm$ 2.03              & 12.25 $\pm$ 2.42 
                                    & 11.46 $\pm$ 0.83              & 12.76 $\pm$ 0.53              & 11.88 $\pm$ 0.16   \\
                                    
CL\cite{ContrastLandmark}   & 64    & 24.87 $\pm$ 2.67              & 15.15 $\pm$ 0.53              & 13.52 $\pm$ 1.08 
                                    & 11.77 $\pm$ 0.68              & 11.57 $\pm$ 0.03              & 10.06 $\pm$ 0.45   \\
                                    
LEAD\cite{LEAD}             & 64    & 21.80 $\pm$ 2.54              & 13.34 $\pm$ 0.43              & 11.50 $\pm$ 0.34 
                                    & 10.13 $\pm$ 0.45              &  9.29 $\pm$ 0.41              &  9.11 $\pm$ 0.25   \\
                                    
SCE-MAE \smallf{(Ours)}     & 64    & 18.41 $\pm$ 1.21              & \textbf{11.79 $\pm$ 0.44}     & \textbf{10.57 $\pm$ 0.24} 
                                    & \textbf{ 9.65 $\pm$ 0.14}     & \textbf{ 8.60 $\pm$ 0.17}     & \textbf{ 8.31 $\pm$ 0.06}   \\
\hline
CL\cite{ContrastLandmark}   & 128   & 27.31 $\pm$ 1.39              & 18.66 $\pm$ 4.59              & 13.39 $\pm$ 0.30 
                                    & 11.77 $\pm$ 0.85              & 10.25 $\pm$ 0.22              &  9.46 $\pm$ 0.05   \\
                                    
LEAD\cite{LEAD}             & 128   & 21.20 $\pm$ 1.67              & 13.22 $\pm$ 1.43              & 10.83 $\pm$ 0.65 
                                    &  9.69 $\pm$ 0.41              &  8.89 $\pm$ 0.20              &  8.83 $\pm$ 0.33   \\

SCE-MAE \smallf{(Ours)}     & 128   & \textbf{20.14 $\pm$ 1.76}     & \textbf{11.99 $\pm$ 0.71}     & \textbf{10.40 $\pm$ 0.22} 
                                    & \textbf{ 9.25 $\pm$ 0.14}     & \textbf{ 8.49 $\pm$ 0.19}     & \textbf{ 7.96 $\pm$ 0.21}   \\
\hline
CL\cite{ContrastLandmark}   & 256   & 28.00 $\pm$ 1.39              & 15.85 $\pm$ 0.86              & 12.98 $\pm$ 0.16 
                                    & 11.18 $\pm$ 0.19              &  9.56 $\pm$ 0.44              &  9.30 $\pm$ 0.20   \\
                                    
LEAD\cite{LEAD}             & 256   & 21.39 $\pm$ 0.74              & 12.38 $\pm$ 1.28              & 11.01 $\pm$ 0.48 
                                    & 10.06 $\pm$ 0.59              &  8.51 $\pm$ 0.09              &  8.56 $\pm$ 0.21   \\

SCE-MAE \smallf{(Ours)}     & 256   & \textbf{17.08 $\pm$ 1.35}     & \textbf{11.28 $\pm$ 0.54}     & \textbf{10.30 $\pm$ 0.09} 
                                    & \textbf{ 8.95 $\pm$ 0.08}     & \textbf{ 8.20 $\pm$ 0.20}     & \textbf{ 7.58 $\pm$ 0.09}   \\
\hline
\end{tabular}
\end{table*}

\subsection{Landmark Detection}

\textbf{Evaluation Protocol.} Following prior works \cite{DVE, ContrastLandmark, LEAD}, we freeze the pretrained backbone and projector, and only train a light-weight regressor. Our regressor consists of a convolution block (instead of a linear layer) and a linear layer when training on all annotated samples. The convolution block utilizes the spatial context to produce $I$ intermediate heatmaps for each landmark which are converted to $I$ pairs of 2D coordinates by a soft-argmax operation and fed to a linear layer that outputs the final landmark prediction. We set $I=50$ for all experiments \cite{DVE, ContrastLandmark, LEAD}. We leverage the first and second stage concatenated features as a more robust input to the regressor as we expect the backbone to provide rich task-agnostic representations during the first stage and the projector to supplement task-specific cues critical for landmark detection during the second stage. For a fair comparison, the reported results are produced with their official implementation and checkpoints, if available.

\noindent\textbf{All Annotated Samples.} We compare our method with prior works on landmark detection benchmarks in Table \ref{table detection}. Once again, even with our smallest backbone DeiT-T, being \emph{a fraction of the size of prior works}, outperforms existing SOTA. Considering our best results, our method achieves \emph{$\sim$9\%-15\%}  performance gain across different benchmarks. Such a compelling performance is an attestation to the excellent discriminative ability of our features, which provide intricate disambiguation cues to the regressor for locating landmarks. We also highlight that \emph{all methods} achieve lower error with our re-annotated AFLW$_{RC}$ test set, hence confirming the higher annotation quality.

\noindent\textbf{Limited Annotated Samples.}
We compare our method with prior works on landmark detection under limited annotation in Table \ref{table detection limitted annotation}. Our method outperforms all existing SSL-based methods with significant performance gain under all annotation and feature dimension settings. Specifically, we achieved a relative gain of \emph{8.6\% on average} and \emph{as high as 20.1\%} compared to the existing SOTA. Furthermore, we observe a smaller standard deviation with repeated experiments, indicating that our method produces optimal features more consistently, hence attesting to its robustness.

\begin{table}[!t]
\caption{\textbf{Component-wise ablation on landmark matching.} The first three rows compare the results using backbone features only. Our raw MAE features (Baseline) are compared against the hypercolumns used in CL and LEAD. The last two rows indicate the inclusion of the clustering and the LCR loss in our framework.}
\centering
\label{table ablation each component}
\begin{tabular}{l|cc|cc}
\hline
\multirow{2}{*}{\textbf{Method}}  & \multirow{2}{*}{\textbf{Cluster}}   & \multirow{2}{*}{$\bm{\mathcal{L_{LCR}}}$} & \textbf{Same}         & \textbf{Diff.}      \rule[2.5ex]{0pt}{0pt}\rule[-1ex]{0pt}{0pt}\\
                             &                       &                   & \multicolumn{2}{c}{\small\textbf{Mean Pixel Error$\downarrow$}} \\
\hline
CL \cite{ContrastLandmark}   & -                   & -            & 0.69                  & 5.37          \\
LEAD \cite{LEAD}             & -                   & -            & 2.35                  & 6.22          \\
Baseline                    & \xmark                & \xmark            & 0.55                  & \underl{3.51}         \\
\hline
\multirow{2}{*}{SCE-MAE \tiny(Ours)}                    & \cmark                & \xmark            & \underl{0.30}         & 4.04          \\
                      & \cmark                & \cmark            & \textbf{0.27}         & \textbf{1.61}          \\

\hline
\end{tabular}
\vspace{-1ex}
\end{table}

\subsection{Ablation Studies}

\textbf{Importance of Each Component.} To better understand our proposed SCE-MAE framework, we report the component-wise ablation analysis on the landmark matching task in Table \ref{table ablation each component}. The first three rows indicate the usage of only the first-stage backbone features, while the last two rows, respectively, indicate the inclusion of clustering and LCR loss in our framework. CL \cite{ContrastLandmark} and LEAD \cite{LEAD} utilize hypercolumns whereas we leverage the vanilla last layer features of the pretrained MAE, which we indicate as Baseline. Using the backbone alone, our baseline outperforms CL and LEAD, validating our motivation that MIM is a more suitable pretext task for landmark representation learning. We observe that the clustering assists the matching between the same identity while the LCR loss boosts the matching performance between different identities. Overall, these trends align with our expectations: initially, the region-level first-stage MAE features capture local intricacies but are too raw to generalize the landmarks across different identities; the clustering disambiguates the landmarks from the unimportant regions, which improves the same identity matching performance; and finally, the LCR loss forges critical local correspondences between important facial regions, resulting in the best performance for both settings.

\begin{figure}[!t]
\begin{tabular}{ccc}
    CL \cite{ContrastLandmark} & LEAD$^\dag$ \cite{LEAD} & SCE-MAE (Ours) \\
    \footnotesize{SC: 0.602 $\pm$ 0.013}  & \footnotesize{SC: 0.613 $\pm$ 0.014}   & \footnotesize{SC: 0.679 $\pm$ 0.013} \\
    \includegraphics[width=0.25\linewidth, height=0.25\linewidth, trim=4 4 4 4,clip]{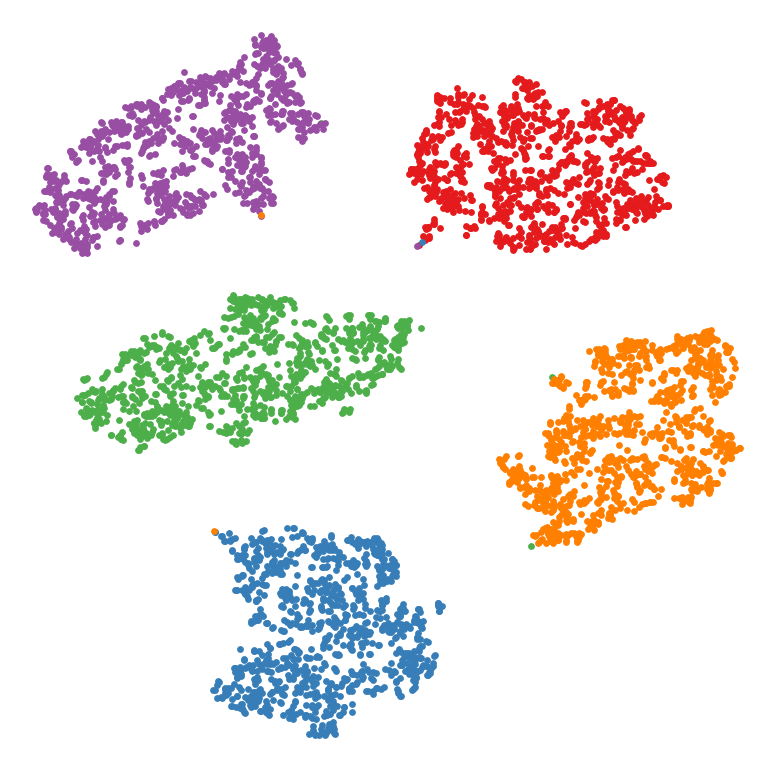} 
    & \includegraphics[width=0.25\linewidth, height=0.25\linewidth, trim=4 4 4 4,clip]{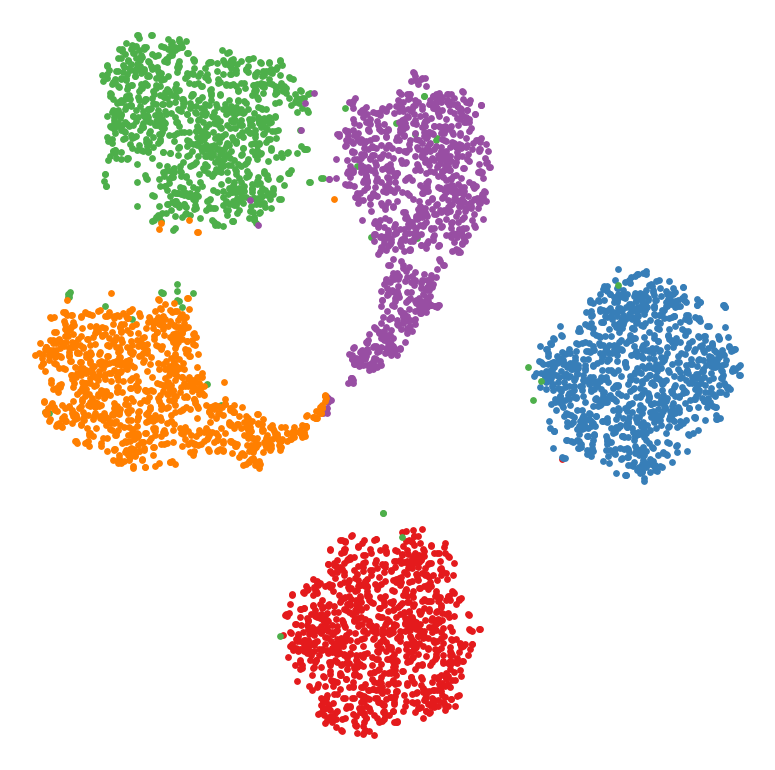}
    & \includegraphics[width=0.25\linewidth, height=0.25\linewidth, trim=4 4 4 4,clip]{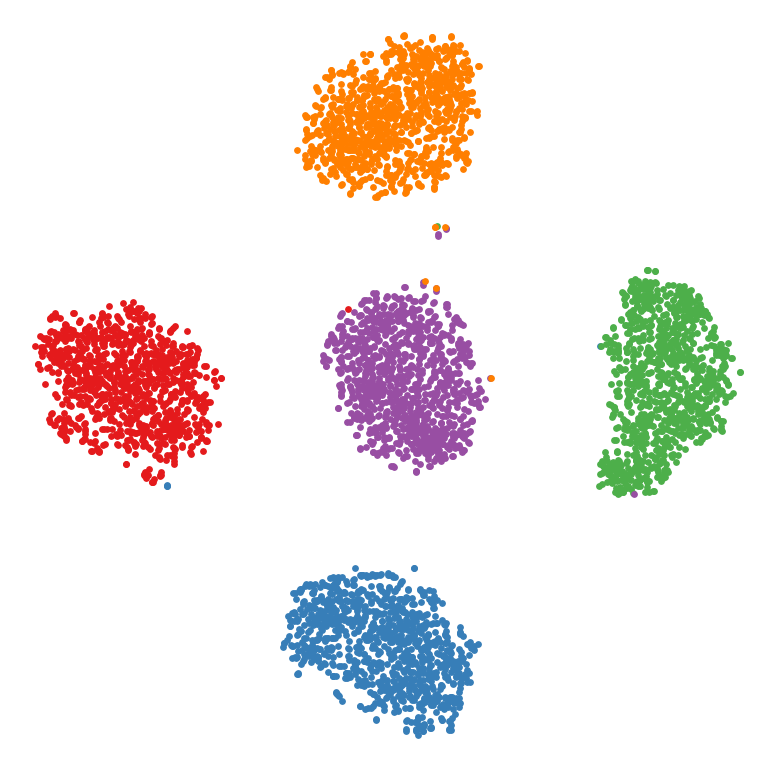} \\
    \multicolumn{3}{c}{\includegraphics[scale=0.2]{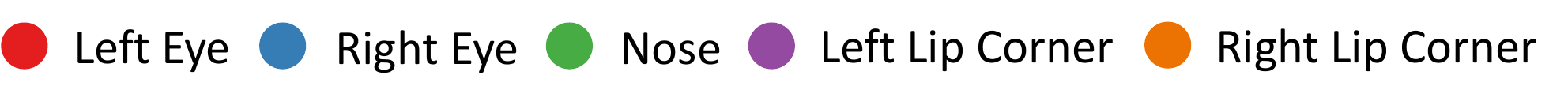}}    
\end{tabular}
\centering
\caption{\textbf{t-SNE plot of the landmark representations.} $\dag$ denotes the usage of the stage 1 hypercolumn representations. SC denotes the Silhouette Coefficient \cite{SilhouetteCoefficient}, a score (higher is better) which measures the quality of the clustering. Our method results in both a clear separation between the landmarks and the densest landmark clusters, resulting in the highest Silhouette Coefficient.}
\label{fig_t-sne}
\end{figure}

\noindent\textbf{Visualization of Landmark Representations.} We visualize the t-SNE plots of the landmark representations corresponding to 1000 test images in Fig.\ \ref{fig_t-sne}. Since LEAD \cite{LEAD} only performs knowledge distillation in its second stage, we use the first-stage hypercolumn representations as it can be considered as the upper bound of the second-stage objective. For each method, we execute t-SNE 100 times and report the mean and standard deviation of the Silhouette Coefficient \cite{SilhouetteCoefficient}, a metric (higher is better) indicating the quality of the clustering as a function of the mean intra-cluster (lower is better) and inter-cluster (higher is better) distance. For CL \cite{ContrastLandmark}, the samples within the cluster are more scattered, resulting in a higher intra-cluster distance. For LEAD, though the clusters are more dense, the clusters of the left/right lip corner and nose are not clearly separated, resulting in a lower inter-cluster distance. Our method results in both well-separated and dense clusters, which is reflected by a high Silhoutte Coefficient, thereby corroborating the superior quality of our landmark representations.

\section{Conclusion}
\label{sec:conclusion}

In this work, we present SCE-MAE, a two-stage framework to tackle the self-supervised face landmark estimation tasks. To learn effective and locally distinct representations, we target structured improvements on both stages. For the first stage, we leverage the region-level MAE instead of instance-level SSL methods to derive more potent initial representations. For the second stage, we demonstrate that it is beneficial to identify important facial regions and directly hone only the salient correspondences. Our approach yields discriminative and high-quality landmark representations that result in superior performance over prior SOTA works on both the landmark matching and detection tasks. Due to the nature of facial data, we believe that further research on the sparsification of the correspondence computation through the systematic elimination of insignificant correspondences could allow future self-supervised landmark estimation methods to better exploit inter-landmark dependencies and form higher-caliber landmark representations.


\clearpage
\clearpage
{
    \small
    \bibliographystyle{ieeenat_fullname}
    \bibliography{main}

\begin{thebibliography}{59}
\providecommand{\natexlab}[1]{#1}
\providecommand{\url}[1]{\texttt{#1}}
\expandafter\ifx\csname urlstyle\endcsname\relax
  \providecommand{\doi}[1]{doi: #1}\else
  \providecommand{\doi}{doi: \begingroup \urlstyle{rm}\Url}\fi

\bibitem[3dF(2021)]{3dFaceReconstruction_2}
3d face reconstruction and dense alignment with a new generated dataset.
\newblock \emph{Displays}, page 102094, 2021.

\bibitem[Balestriero and LeCun(2022)]{NEURIPS2022_aa56c745}
Randall Balestriero and Yann LeCun.
\newblock Contrastive and non-contrastive self-supervised learning recover global and local spectral embedding methods.
\newblock In \emph{NeurIPS}, 2022.

\bibitem[Bao et~al.(2021)Bao, Dong, Piao, and Wei]{BEiT}
Hangbo Bao, Li Dong, Songhao Piao, and Furu Wei.
\newblock {BEiT}: {BERT} pre-training of image transformers.
\newblock In \emph{ICLR}, 2021.

\bibitem[Bougourzi et~al.(2022)Bougourzi, Dornaika, and Taleb-Ahmed]{FBP_1}
F. Bougourzi, F. Dornaika, and A. Taleb-Ahmed.
\newblock Deep learning based face beauty prediction via dynamic robust losses and ensemble regression.
\newblock \emph{Knowledge-Based Systems}, 2022.

\bibitem[Caron et~al.(2021)Caron, Touvron, Misra, J{\'e}gou, Mairal, Bojanowski, and Joulin]{DINO}
Mathilde Caron, Hugo Touvron, Ishan Misra, Herv{\'e} J{\'e}gou, Julien Mairal, Piotr Bojanowski, and Armand Joulin.
\newblock Emerging properties in self-supervised vision transformers.
\newblock In \emph{ICCV}, 2021.

\bibitem[Chen et~al.(2020{\natexlab{a}})Chen, Kornblith, Norouzi, and Hinton]{SimCLR}
Ting Chen, Simon Kornblith, Mohammad Norouzi, and Geoffrey Hinton.
\newblock A simple framework for contrastive learning of visual representations.
\newblock In \emph{ICML}, 2020{\natexlab{a}}.

\bibitem[Chen et~al.(2020{\natexlab{b}})Chen, Fan, Girshick, and He]{MOCOv2}
Xinlei Chen, Haoqi Fan, Ross Girshick, and Kaiming He.
\newblock Improved baselines with momentum contrastive learning.
\newblock \emph{arXiv preprint arXiv:2003.04297}, 2020{\natexlab{b}}.

\bibitem[Chen et~al.(2021)Chen, Xie, and He]{MOCOv3}
Xinlei Chen, Saining Xie, and Kaiming He.
\newblock An empirical study of training self-supervised vision transformers.
\newblock In \emph{ICCV}, 2021.

\bibitem[Cheng et~al.(2021)Cheng, Su, and Maji]{ContrastLandmark}
Zezhou Cheng, Jong-Chyi Su, and Subhransu Maji.
\newblock On equivariant and invariant learning of object landmark representations.
\newblock In \emph{ICCV}, 2021.

\bibitem[Dosovitskiy et~al.(2021)Dosovitskiy, Beyer, Kolesnikov, Weissenborn, Zhai, Unterthiner, Dehghani, Minderer, Heigold, Gelly, Uszkoreit, and Houlsby]{ViT}
Alexey Dosovitskiy, Lucas Beyer, Alexander Kolesnikov, Dirk Weissenborn, Xiaohua Zhai, Thomas Unterthiner, Mostafa Dehghani, Matthias Minderer, Georg Heigold, Sylvain Gelly, Jakob Uszkoreit, and Neil Houlsby.
\newblock An image is worth 16x16 words: Transformers for image recognition at scale.
\newblock In \emph{ICLR}, 2021.

\bibitem[Gidaris et~al.(2018)Gidaris, Singh, and Komodakis]{RotNet}
Spyros Gidaris, Praveer Singh, and Nikos Komodakis.
\newblock Unsupervised representation learning by predicting image rotations.
\newblock In \emph{ICLR}, 2018.

\bibitem[Grill et~al.(2020)Grill, Strub, Altch{\'e}, Tallec, Richemond, Buchatskaya, Doersch, Avila~Pires, Guo, Gheshlaghi~Azar, et~al.]{BYOL}
Jean-Bastien Grill, Florian Strub, Florent Altch{\'e}, Corentin Tallec, Pierre Richemond, Elena Buchatskaya, Carl Doersch, Bernardo Avila~Pires, Zhaohan Guo, Mohammad Gheshlaghi~Azar, et~al.
\newblock Bootstrap your own latent-a new approach to self-supervised learning.
\newblock In \emph{NeurIPS}, 2020.

\bibitem[He et~al.(2020)He, Fan, Wu, Xie, and Girshick]{MOCO}
Kaiming He, Haoqi Fan, Yuxin Wu, Saining Xie, and Ross Girshick.
\newblock Momentum contrast for unsupervised visual representation learning.
\newblock In \emph{CVPR}, 2020.

\bibitem[He et~al.(2022)He, Chen, Xie, Li, Doll{\'a}r, and Girshick]{MAE}
Kaiming He, Xinlei Chen, Saining Xie, Yanghao Li, Piotr Doll{\'a}r, and Ross Girshick.
\newblock Masked autoencoders are scalable vision learners.
\newblock In \emph{CVPR}, 2022.

\bibitem[Huang et~al.(2021)Huang, Yang, Li, Kim, and Wei]{adnet}
Yangyu Huang, Hao Yang, Chong Li, Jongyoo Kim, and Fangyun Wei.
\newblock Adnet: Leveraging error-bias towards normal direction in face alignment.
\newblock In \emph{ICCV}, 2021.

\bibitem[J.~Iyer et~al.(2021)J.~Iyer, K., Nersisson, Zhuang, Joseph~Raj, and Refayee]{FBP_2}
Tharun J.~Iyer, Rahul K., Ruban Nersisson, Zhemin Zhuang, Alex~Noel Joseph~Raj, and Imthiaz Refayee.
\newblock Machine learning-based facial beauty prediction and analysis of frontal facial images using facial landmarks and traditional image descriptors.
\newblock \emph{Computational Intelligence and Neuroscience}, 2021.

\bibitem[Jakab et~al.(2018)Jakab, Gupta, Bilen, and Vedaldi]{ImGen}
Tomas Jakab, Ankush Gupta, Hakan Bilen, and Andrea Vedaldi.
\newblock Unsupervised learning of object landmarks through conditional image generation.
\newblock In \emph{NeurIPS}, 2018.

\bibitem[Juhong and Pintavirooj(2017)]{FaceRecog_1}
Aniwat Juhong and C. Pintavirooj.
\newblock Face recognition based on facial landmark detection.
\newblock In \emph{2017 10th Biomedical Engineering International Conference (BMEiCON)}, 2017.

\bibitem[Karmali et~al.(2022)Karmali, Atrishi, Harsha, Agrawal, Jampani, and Babu]{LEAD}
Tejan Karmali, Abhinav Atrishi, Sai~Sree Harsha, Susmit Agrawal, Varun Jampani, and R~Venkatesh Babu.
\newblock {LEAD}: Self-supervised landmark estimation by aligning distributions of feature similarity.
\newblock In \emph{WACV}, 2022.

\bibitem[Kips et~al.(2021)Kips, Jiang, Ba, Phung, Aarabi, Gori, Perrot, and Bloch]{Makeup_Tryon_2}
Robin Kips, Ruowei Jiang, Sileye Ba, Edmund Phung, Parham Aarabi, Pietro Gori, Matthieu Perrot, and Isabelle Bloch.
\newblock Deep graphics encoder for real-time video makeup synthesis from example.
\newblock In \emph{CVPRW}, 2021.

\bibitem[Koestinger et~al.(2011)Koestinger, Wohlhart, Roth, and Bischof]{AFLW}
Martin Koestinger, Paul Wohlhart, Peter~M Roth, and Horst Bischof.
\newblock Annotated facial landmarks in the wild: A large-scale, real-world database for facial landmark localization.
\newblock In \emph{ICCVW}, 2011.

\bibitem[Kong et~al.(2023)Kong, Ma, Chen, Xing, Chi, Morency, and Zhang]{understandingMAE}
Lingjing Kong, Martin~Q Ma, Guangyi Chen, Eric~P Xing, Yuejie Chi, Louis-Philippe Morency, and Kun Zhang.
\newblock Understanding masked autoencoders via hierarchical latent variable models.
\newblock In \emph{CVPR}, 2023.

\bibitem[Kumar et~al.(2020)Kumar, Marks, Mou, Wang, Jones, Cherian, Koike-Akino, Liu, and Feng]{Luvli}
Abhinav Kumar, Tim~K. Marks, Wenxuan Mou, Ye Wang, Michael Jones, Anoop Cherian, Toshiaki Koike-Akino, Xiaoming Liu, and Chen Feng.
\newblock Luvli face alignment: Estimating landmarks' location, uncertainty, and visibility likelihood.
\newblock In \emph{CVPR}, 2020.

\bibitem[Li et~al.(2019)Li, Yu, Phung, Duke, Kezele, and Aarabi]{Makeup_Tryon_3}
TianXing Li, Zhi Yu, Edmund Phung, Brendan Duke, Irina Kezele, and Parham Aarabi.
\newblock Lightweight real-time makeup try-on in mobile browsers with tiny cnn models for facial tracking.
\newblock In \emph{CVPRW}, 2019.

\bibitem[Li et~al.(2023)Li, Xie, and Loy]{li2023correlational}
Wei Li, Jiahao Xie, and Chen~Change Loy.
\newblock Correlational image modeling for self-supervised visual pre-training.
\newblock In \emph{CVPR}, 2023.

\bibitem[Liu et~al.(2023)Liu, Huang, Zheng, Liu, and Li]{MixMAE}
Jihao Liu, Xin Huang, Jinliang Zheng, Yu Liu, and Hongsheng Li.
\newblock {MixMAE}: Mixed and masked autoencoder for efficient pretraining of hierarchical vision transformers.
\newblock In \emph{CVPR}, 2023.

\bibitem[Liu et~al.(2015)Liu, Luo, Wang, and Tang]{CelebA}
Ziwei Liu, Ping Luo, Xiaogang Wang, and Xiaoou Tang.
\newblock Deep learning face attributes in the wild.
\newblock In \emph{ICCV}, 2015.

\bibitem[Long et~al.(2023)Long, Zhao, Pi, Wang, and Wang]{long2023beyond}
Sifan Long, Zhen Zhao, Jimin Pi, Shengsheng Wang, and Jingdong Wang.
\newblock Beyond attentive tokens: Incorporating token importance and diversity for efficient vision transformers.
\newblock In \emph{CVPR}, 2023.

\bibitem[Lorenz et~al.(2019)Lorenz, Bereska, Milbich, and Ommer]{lorenz2019unsupervised}
Dominik Lorenz, Leonard Bereska, Timo Milbich, and Bj{\"o}rn Ommer.
\newblock Unsupervised part-based disentangling of object shape and appearance.
\newblock In \emph{CVPR}, 2019.

\bibitem[Mallis et~al.(2020)Mallis, Sanchez, Bell, and Tzimiropoulos]{mallis2020unsupervised}
Dimitrios Mallis, Enrique Sanchez, Matthew Bell, and Georgios Tzimiropoulos.
\newblock Unsupervised learning of object landmarks via self-training correspondence.
\newblock In \emph{NeurIPS}, 2020.

\bibitem[Marelli et~al.(2022)Marelli, Bianco, and Ciocca]{Makeup_Tryon_1}
Davide Marelli, Simone Bianco, and Gianluigi Ciocca.
\newblock Designing an {AI-based} virtual try-on web application.
\newblock \emph{Sensors (Basel)}, 2022.

\bibitem[Munasinghe(2018)]{FER_1}
M.~I. N.~P. Munasinghe.
\newblock Facial expression recognition using facial landmarks and random forest classifier.
\newblock In \emph{2018 IEEE/ACIS 17th International Conference on Computer and Information Science (ICIS)}, 2018.

\bibitem[Ngoc et~al.(2020)Ngoc, Lee, and Song]{FER_2}
Quang~Tran Ngoc, Seunghyun Lee, and Byung~Cheol Song.
\newblock Facial landmark-based emotion recognition via directed graph neural network.
\newblock \emph{Electronics}, 2020.

\bibitem[Noroozi and Favaro(2016)]{Jigsaw}
Mehdi Noroozi and Paolo Favaro.
\newblock Unsupervised learning of visual representations by solving jigsaw puzzles.
\newblock In \emph{ECCV}, 2016.

\bibitem[Noroozi et~al.(2018)Noroozi, Vinjimoor, Favaro, and Pirsiavash]{Jigsaw++}
Mehdi Noroozi, Ananth Vinjimoor, Paolo Favaro, and Hamed Pirsiavash.
\newblock Boosting self-supervised learning via knowledge transfer.
\newblock In \emph{CVPR}, 2018.

\bibitem[Oord et~al.(2018)Oord, Li, and Vinyals]{InfoNCE}
Aaron van~den Oord, Yazhe Li, and Oriol Vinyals.
\newblock Representation learning with contrastive predictive coding.
\newblock \emph{arXiv preprint arXiv:1807.03748}, 2018.

\bibitem[Oquab et~al.(2023)Oquab, Darcet, Moutakanni, Vo, Szafraniec, Khalidov, Fernandez, Haziza, Massa, El-Nouby, et~al.]{DINOv2}
Maxime Oquab, Timoth{\'e}e Darcet, Th{\'e}o Moutakanni, Huy Vo, Marc Szafraniec, Vasil Khalidov, Pierre Fernandez, Daniel Haziza, Francisco Massa, Alaaeldin El-Nouby, et~al.
\newblock {DINOv2}: Learning robust visual features without supervision.
\newblock \emph{arXiv preprint arXiv:2304.07193}, 2023.

\bibitem[Roh et~al.(2021)Roh, Shin, Kim, and Kim]{SCRL}
Byungseok Roh, Wuhyun Shin, Ildoo Kim, and Sungwoong Kim.
\newblock Spatially consistent representation learning.
\newblock In \emph{CVPR}, 2021.

\bibitem[Rousseeuw(1987)]{SilhouetteCoefficient}
Peter~J. Rousseeuw.
\newblock Silhouettes: A graphical aid to the interpretation and validation of cluster analysis.
\newblock \emph{Journal of Computational and Applied Mathematics}, 1987.

\bibitem[Sagonas et~al.(2013)Sagonas, Tzimiropoulos, Zafeiriou, and Pantic]{300W}
Christos Sagonas, Georgios Tzimiropoulos, Stefanos Zafeiriou, and Maja Pantic.
\newblock 300 faces in-the-wild challenge: The first facial landmark localization challenge.
\newblock In \emph{ICCVW}, 2013.

\bibitem[Sarsenov and Latuta(2017)]{FaceRecog_2}
Adil Sarsenov and Konstantin Latuta.
\newblock Face recognition based on facial landmarks.
\newblock In \emph{2017 IEEE 11th International Conference on Application of Information and Communication Technologies (AICT)}, 2017.

\bibitem[Shu et~al.(2018)Shu, Sahasrabudhe, Guler, Samaras, Paragios, and Kokkinos]{DeformAE}
Zhixin Shu, Mihir Sahasrabudhe, Riza~Alp Guler, Dimitris Samaras, Nikos Paragios, and Iasonas Kokkinos.
\newblock Deforming autoencoders: Unsupervised disentangling of shape and appearance.
\newblock In \emph{ECCV}, 2018.

\bibitem[Sohn(2016)]{NT-Xent}
Kihyuk Sohn.
\newblock Improved deep metric learning with multi-class n-pair loss objective.
\newblock In \emph{NeurIPS}, 2016.

\bibitem[Thewlis et~al.(2017{\natexlab{a}})Thewlis, Bilen, and Vedaldi]{Dense3D}
James Thewlis, Hakan Bilen, and Andrea Vedaldi.
\newblock Unsupervised learning of object frames by dense equivariant image labelling.
\newblock In \emph{NeurIPS}, 2017{\natexlab{a}}.

\bibitem[Thewlis et~al.(2017{\natexlab{b}})Thewlis, Bilen, and Vedaldi]{Sparse}
James Thewlis, Hakan Bilen, and Andrea Vedaldi.
\newblock Unsupervised learning of object landmarks by factorized spatial embeddings.
\newblock In \emph{ICCV}, 2017{\natexlab{b}}.

\bibitem[Thewlis et~al.(2019)Thewlis, Albanie, Bilen, and Vedaldi]{DVE}
James Thewlis, Samuel Albanie, Hakan Bilen, and Andrea Vedaldi.
\newblock Unsupervised learning of landmarks by descriptor vector exchange.
\newblock In \emph{ICCV}, 2019.

\bibitem[Touvron et~al.(2021)Touvron, Cord, Douze, Massa, Sablayrolles, and J{\'e}gou]{DeiT}
Hugo Touvron, Matthieu Cord, Matthijs Douze, Francisco Massa, Alexandre Sablayrolles, and Herv{\'e} J{\'e}gou.
\newblock Training data-efficient image transformers \& distillation through attention.
\newblock In \emph{ICML}, 2021.

\bibitem[Wang et~al.(2021)Wang, Zhang, Shen, Kong, and Li]{DenseCL}
Xinlong Wang, Rufeng Zhang, Chunhua Shen, Tao Kong, and Lei Li.
\newblock Dense contrastive learning for self-supervised visual pre-training.
\newblock In \emph{CVPR}, 2021.

\bibitem[Wood et~al.(2022)Wood, Baltru{\v{s}}aitis, Hewitt, Johnson, Shen, Milosavljevi{\'{c}}, Wilde, Garbin, Sharp, Stojiljkovi{\'{c}}, Cashman, and Valentin]{3dFaceReconstruction_1}
Erroll Wood, Tadas Baltru{\v{s}}aitis, Charlie Hewitt, Matthew Johnson, Jingjing Shen, Nikola Milosavljevi{\'{c}}, Daniel Wilde, Stephan Garbin, Toby Sharp, Ivan Stojiljkovi{\'{c}}, Tom Cashman, and Julien Valentin.
\newblock 3d face reconstruction with dense landmarks.
\newblock In \emph{ECCV}, 2022.

\bibitem[Xie et~al.(2023)Xie, Pang, Bader, and Wang]{xie2023maester}
Ronald Xie, Kuan Pang, Gary~D Bader, and Bo Wang.
\newblock {MAESTER}: Masked autoencoder guided segmentation at pixel resolution for accurate, self-supervised subcellular structure recognition.
\newblock In \emph{CVPR}, 2023.

\bibitem[Xie et~al.(2022)Xie, Zhang, Cao, Lin, Bao, Yao, Dai, and Hu]{SimMIM}
Zhenda Xie, Zheng Zhang, Yue Cao, Yutong Lin, Jianmin Bao, Zhuliang Yao, Qi Dai, and Han Hu.
\newblock {SimMIM}: a simple framework for masked image modeling.
\newblock In \emph{CVPR}, 2022.

\bibitem[Xu et~al.(2020)Xu, Yang, Liu, Dai, and Zhou]{xu2020unsupervised}
Yinghao Xu, Ceyuan Yang, Ziwei Liu, Bo Dai, and Bolei Zhou.
\newblock Unsupervised landmark learning from unpaired data.
\newblock \emph{arXiv preprint arXiv:2007.01053}, 2020.

\bibitem[Yeh et~al.(2022)Yeh, Hong, Hsu, Liu, Chen, and LeCun]{DCL}
Chun-Hsiao Yeh, Cheng-Yao Hong, Yen-Chi Hsu, Tyng-Luh Liu, Yubei Chen, and Yann LeCun.
\newblock Decoupled contrastive learning.
\newblock In \emph{ECCV}, 2022.

\bibitem[Zhang et~al.(2022)Zhang, Zhang, Pham, Niu, Qiao, Yoo, and Kweon]{SimMoCo}
Chaoning Zhang, Kang Zhang, Trung~X. Pham, Axi Niu, Zhinan Qiao, Chang~D. Yoo, and In~So Kweon.
\newblock Dual temperature helps contrastive learning without many negative samples: Towards understanding and simplifying moco.
\newblock In \emph{CVPR}, 2022.

\bibitem[Zhang et~al.(2018)Zhang, Guo, Jin, Luo, He, and Lee]{StructuralRepr}
Yuting Zhang, Yijie Guo, Yixin Jin, Yijun Luo, Zhiyuan He, and Honglak Lee.
\newblock Unsupervised discovery of object landmarks as structural representations.
\newblock In \emph{CVPR}, 2018.

\bibitem[Zhang et~al.(2014)Zhang, Luo, Loy, and Tang]{MTFL}
Zhanpeng Zhang, Ping Luo, Chen~Change Loy, and Xiaoou Tang.
\newblock Facial landmark detection by deep multi-task learning.
\newblock In \emph{ECCV}, 2014.

\bibitem[Zhang et~al.(2015)Zhang, Luo, Loy, and Tang]{MAFL}
Zhanpeng Zhang, Ping Luo, Chen~Change Loy, and Xiaoou Tang.
\newblock Learning deep representation for face alignment with auxiliary attributes.
\newblock \emph{PAMI}, 2015.

\bibitem[Zhou et~al.(2021)Zhou, Wei, Wang, Shen, Xie, Yuille, and Kong]{iBOT}
Jinghao Zhou, Chen Wei, Huiyu Wang, Wei Shen, Cihang Xie, Alan Yuille, and Tao Kong.
\newblock Image {BERT} pre-training with online tokenizer.
\newblock In \emph{ICLR}, 2021.

\bibitem[Zhou et~al.(2023)Zhou, Li, Liu, Wang, Yu, and Ji]{starloss}
Zhenglin Zhou, Huaxia Li, Hong Liu, Nanyang Wang, Gang Yu, and Rongrong Ji.
\newblock {STAR Loss}: Reducing semantic ambiguity in facial landmark detection.
\newblock In \emph{CVPR}, 2023.

\end{thebibliography}
}
\clearpage
\setcounter{page}{1}
\setcounter{table}{1}
\renewcommand{\thetable}{S\arabic{table}}
\renewcommand{\thefigure}{S\arabic{figure}}
\twocolumn[{%
\renewcommand\twocolumn[1][]{#1}%
\maketitlesupplementary

\vspace{-1.5em}
\begin{center}
\begin{tabular}{cccccccc}
      \includegraphics[align=c, width=0.1\linewidth, height=0.1\linewidth]{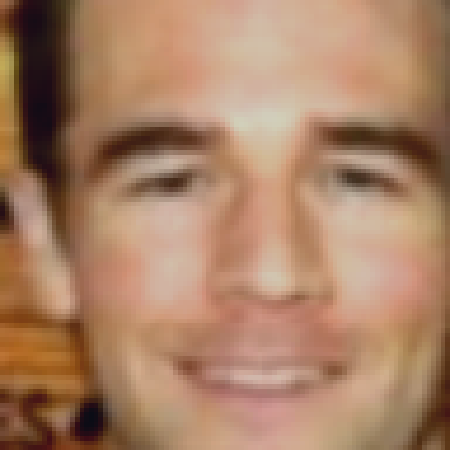}
    & \includegraphics[align=c, width=0.1\linewidth, height=0.1\linewidth]{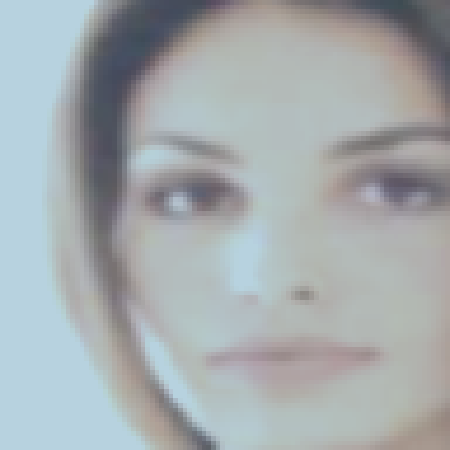}
    & \includegraphics[align=c, width=0.1\linewidth, height=0.1\linewidth]{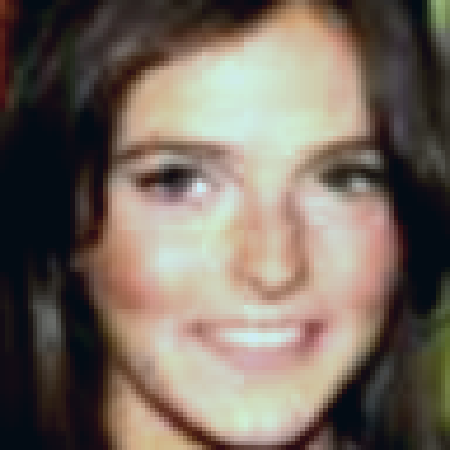}
    & \includegraphics[align=c, width=0.1\linewidth, height=0.1\linewidth]{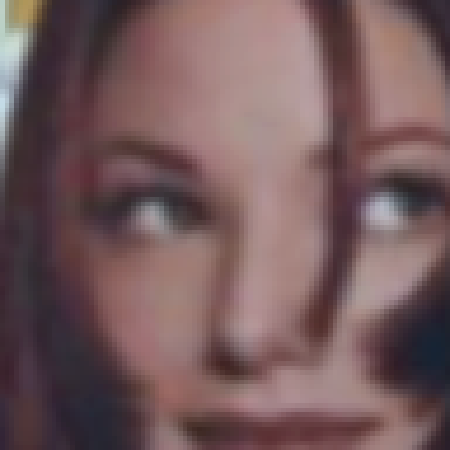}
    & \includegraphics[align=c, width=0.1\linewidth, height=0.1\linewidth]{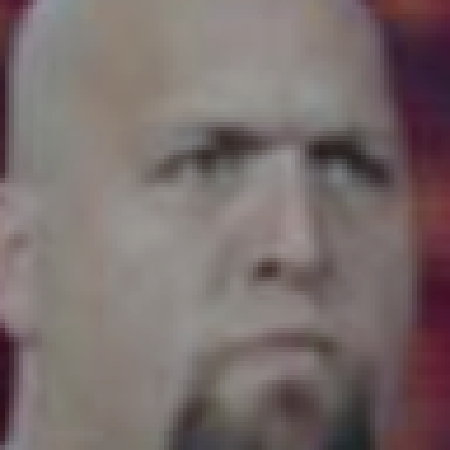}
    & \includegraphics[align=c, width=0.1\linewidth, height=0.1\linewidth]{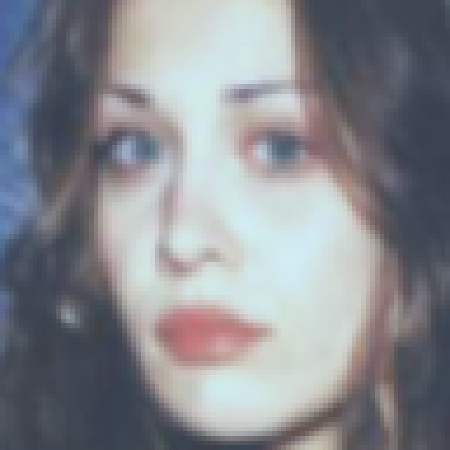}
    & \includegraphics[align=c, width=0.1\linewidth, height=0.1\linewidth]{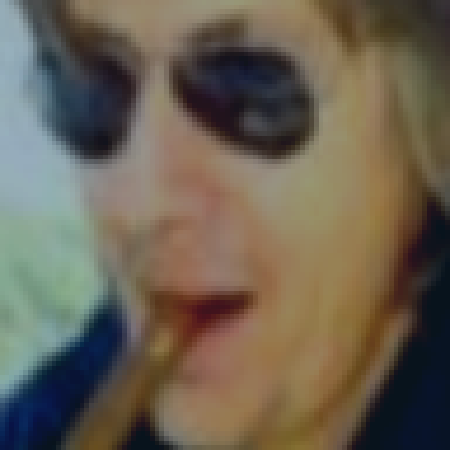}
    & \includegraphics[align=c, width=0.1\linewidth, height=0.1\linewidth]{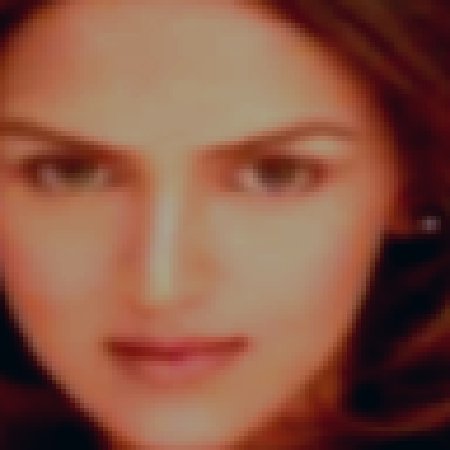} \rule[5ex]{0pt}{0pt}\rule[-4ex]{0pt}{0pt}\\
    \hline
      \includegraphics[align=c, width=0.1\linewidth, height=0.1\linewidth]{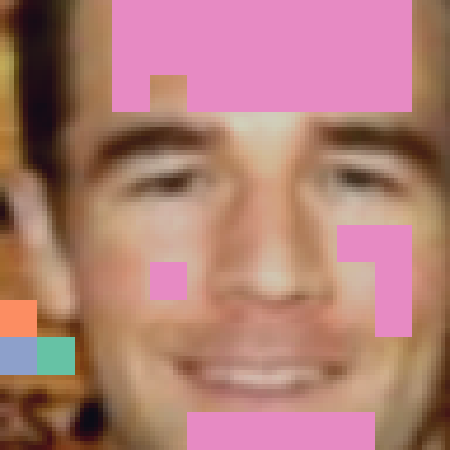}
    & \includegraphics[align=c, width=0.1\linewidth, height=0.1\linewidth]{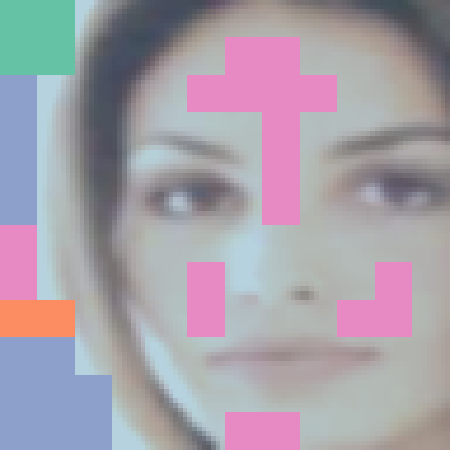}
    & \includegraphics[align=c, width=0.1\linewidth, height=0.1\linewidth]{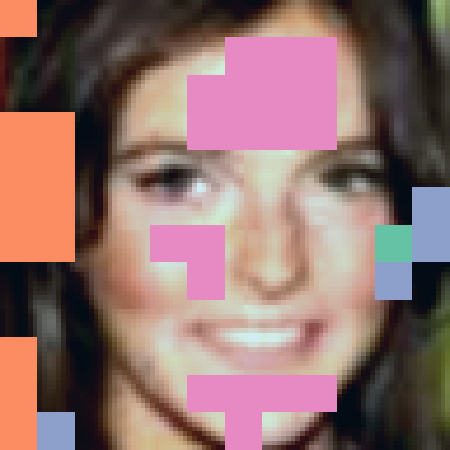}
    & \includegraphics[align=c, width=0.1\linewidth, height=0.1\linewidth]{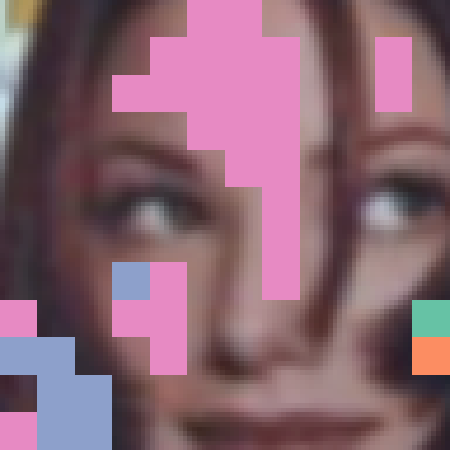}
    & \includegraphics[align=c, width=0.1\linewidth, height=0.1\linewidth]{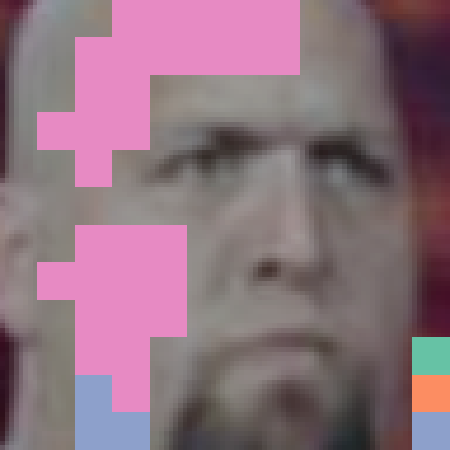}
    & \includegraphics[align=c, width=0.1\linewidth, height=0.1\linewidth]{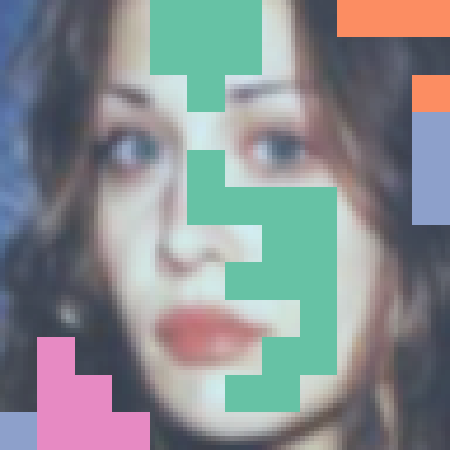}
    & \includegraphics[align=c, width=0.1\linewidth, height=0.1\linewidth]{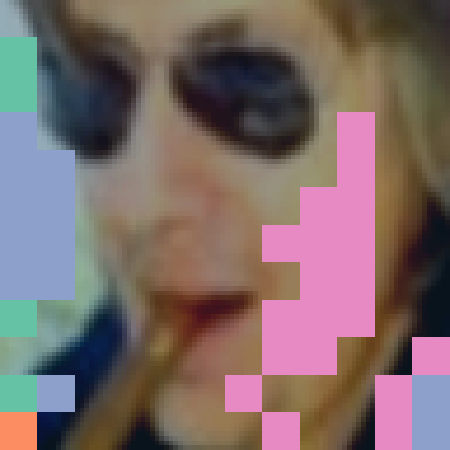}
    & \includegraphics[align=c, width=0.1\linewidth, height=0.1\linewidth]{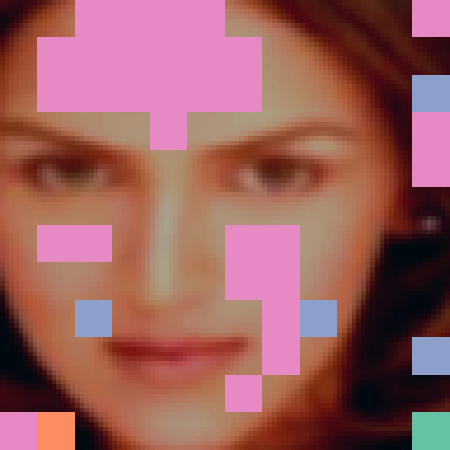} \rule[5ex]{0pt}{0pt}\rule[-4ex]{0pt}{0pt}\\
\end{tabular}
\centering
\captionsetup{type=figure}
\captionof{figure}{\textbf{Visualization of inattentive regions and clustering.} The original images are shown in the first row and the visualization results are shown in the second row. The inattentive regions are represented by colored patches and each color represents one cluster. We can see none of the landmark regions are classified as inattentive regions and semantically similar inattentive regions are grouped together.}
\label{fig visualization of clustering}
\end{center}%
}]

\section{Qualitative Studies on Inattentive Regions.} %
In Section 1 of the main paper, we highlighted the observation that the non-landmark regions (e.g., cheeks and foreheads) are larger and more uniform than the sparse and distinctive landmark regions. In Section 3.2, we explained that in order to setup selective correspondence, we first target the separation of the critical facial and the insignificant regions using the CLS token output of the MAE, and then run a simple clustering algorithm on the insignificant regions. To better understand this attentive-inattentive separation and how the clustering works, we visualize a few examples in Figure \ref{fig visualization of clustering}. The inattentive regions are denoted by colorful patches and the patches with the same color belong to the same cluster. We observe that semantically similar regions are clustered together and none of the landmark regions are classified as inattentive regions, hence corroborating our earlier hypothesis.

\section{Effects of Changing Backbone Architecture}
In this line of work, we are the first to adopt the Vision-Transformer \cite{ViT} as the backbone architecture. In this section, we evaluate whether prior works can benefit by simply changing the backbone to ViT architectures. We switch the backbone of CL \cite{ContrastLandmark} and LEAD \cite{LEAD} and report the quantitative results on landmark matching in Table \ref{table matching backbone}, and on landmark detection in Table \ref{table detection backbone}. Note that both CL and LEAD rely on the extraction of \emph{hypercolumns which require feature map of different spatial resolution.} However, the hypercolumns are not compatible with our backbones as DeiTs \cite{DeiT} are columnar (patch-based) architectures which can only output feature maps at the same spatial size. For this reason and for a fair comparison with our work, we evaluate the previous methods using the last layer feature from DeiTs. For landmark matching, the mean pixel error increases dramatically after changing the backbone for both the matching between same and different identities. We observe similar phenomenon on landmark detection where the performance drops on all evaluated datasets except for MAFL. Additionally, we find that the performance of CL and LEAD does not improve with a larger backbone (DeiT-S compared to DeiT-T). We attribute the performance drop to two main reasons: (1) the first stage SSL protocols of CL and LEAD, namely MoCo \cite{MOCO} and BYOL \cite{BYOL}, are not designed to accommodate the requirements of vision-transformer backbone. This also explains why the performance doesn't improve after applied a larger backbone. Although integration of ViT to the MoCo framework has been addressed in MoCov3 \cite{MOCOv3}, integrating MoCov3 to CL is beyond the scope of our work. (2) The use of hypercolumns is essential for CL and LEAD, however they are not available when using the DeiTs. In conclusion, na\"ively switching the backbone architecture does not necessarily yield better results. The performance gain of our SCE-MAE framework over existing SOTA originates due to the intrinsic compatibility of the first-stage MAE protocol (with ViT backbones) and the ability to leverage the ViT output during the second stage. 

\begin{table*}[!t]
\caption{\textbf{Quantitative evaluations on landmark matching using different backbone architectures.} We report the mean pixel error between the prediction and ground-truth on 1000 image pairs sampled from MAFL. The best and second best results are shown in \textbf{bold} and \underl{underline} respectively. We group the results by backbone architecture. The error of previous SOTA methods increase dramatically when switching the backbone from (ResNet-50 + Hypercolumn) to DeiTs. This demonstrates that na\"ively changing the backbone architecture does not yield better performance.}
\centering
\label{table matching backbone}
\begin{tabular}{l|cc|cc}
\hline
\multirow{2}{*}{\textbf{Method}} & \multirow{2}{*}{\textbf{Backbone}} & \textbf{\#Parameters}   & \textbf{Same}     & \textbf{Different}      \rule[2.5ex]{0pt}{0pt}\\
                            & & \textbf{Millions}        & \multicolumn{2}{c}{\textbf{Mean Pixel Error$\downarrow$}} \rule[-1ex]{0pt}{0pt}\\
\hline
CL\cite{ContrastLandmark}   & ResNet-50 + Hypercolumn     & 23.8       & 0.71                  & 2.50          \\
LEAD\cite{LEAD}             & ResNet-50 + Hypercolumn    & 23.8       & 0.48                  & 2.06          \\
\hline
CL\cite{ContrastLandmark}   & DeiT-T        & 5.4        & 1.31                  & 4.32          \\
LEAD\cite{LEAD}             & DeiT-T        & 5.4        & 0.93                  & 8.86          \\
Ours                        & DeiT-T        & 5.4        & \underl{0.47}         & \underl{1.99}          \\
\hline
CL\cite{ContrastLandmark}   & DeiT-S        & 21.4       & 3.31                  & 7.32          \\
LEAD\cite{LEAD}             & DeiT-S        & 21.4       & 0.91                  & 8.64          \\
Ours                        & DeiT-S        & 21.4       & \textbf{0.31}         & \textbf{1.69}          \\
\hline
\end{tabular}
\end{table*}

\begin{table*}[!t]
\caption{\textbf{Quantitative evaluations on landmark detection using different backbone architectures.} We report the error as the percentage of inter-ocular distance on four human face datasets: MAFL, AFLW$_M$, AFLW$_R$ and 300W. For AFLW$_R$, we report the results on both the original (AFLW$_{RO}$) and corrected (AFLW$_{RC}$) datasets. We group the results by backbone architecture. We can see the performance of CL and LEAD drops when using DeiTs on all datasets except for MAFL, which demonstrates that na\"ively changing the backbone architecture cannot necessarily yield better performance. }
\centering
\label{table detection backbone}
\begin{tabular}{l|cc|ccccc}
\hline
\multirow{2}{*}{\textbf{Method}} & \multirow{2}{*}{\textbf{Backbone}} & \textbf{\#Parameters} & \textbf{MAFL} & \textbf{AFLW$_M$} & \textbf{AFLW$_{RO}$} & \textbf{AFLW$_{RC}$}   & \textbf{300W} \rule[2.5ex]{0pt}{0pt}\\
                            & & \textbf{Millions}   & \multicolumn{5}{c}{\textbf{Inter-ocular Distance (\%)$\downarrow$}} \rule[-1ex]{0pt}{0pt}\\
\hline
CL\cite{ContrastLandmark}   & ResNet-50 + Hypercolumn & 23.8      & 2.76          & 6.17          & 5.69          & 5.06          & 4.84    \\
LEAD\cite{LEAD}             & ResNet-50 + Hypercolumn & 23.8      & 2.44          & 6.05          & 5.71          & 5.11          & 4.87    \\
\hline
CL\cite{ContrastLandmark}   & DeiT-T    & 5.4       & 2.51          & 6.72          & 5.98          & 5.43          & 4.92    \\
LEAD\cite{LEAD}             & DeiT-T    & 5.4       & 2.40          & 6.81          & 6.03          & 5.41          & 5.03    \\
Ours                        & DeiT-T    & 5.4       & \underl{2.20} & \underl{5.89} & \underl{5.54} & \underl{4.86} & \underl{4.22}  \\
\hline
CL\cite{ContrastLandmark}   & DeiT-S    & 21.4      & 2.43          & 6.73          & 5.88          & 5.29          & 4.90    \\
LEAD\cite{LEAD}             & DeiT-S    & 21.4      & 2.39          & 6.88          & 5.91          & 5.32          & 5.10    \\
Ours                        & DeiT-S    & 21.4      & \textbf{2.08} & \textbf{5.33} & \textbf{5.40} & \textbf{4.69} & \textbf{3.94}    \\
\hline
\end{tabular}
\end{table*}

\section{Choices of Hyperparameters}

\subsection{Attentive Rate} 
\begin{figure}[!t]
\begin{tabular}{c}
    \includegraphics[width=0.8\linewidth]{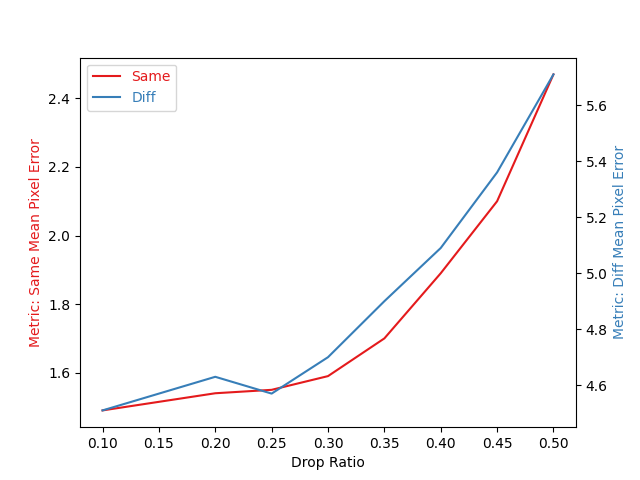}
\end{tabular}
\centering
\caption{Landmark matching results (Mean Pixel Error) at different drop rate.}
\label{fig drop rate}
\end{figure}

We use attentive rate $\eta$ to decide the portion of attentive and inattentive tokens. We study the best choice of this hyperparameter by directly dropping a certain portion of the patch tokens. The idea is that the inattentive tokens are not critical for downstream evaluation as they are mainly non-landmark regions, thus if we directly drop them, it will not affect the evaluation results much. We plot the landmark matching results at different drop rate in Figure \ref{fig drop rate}. We find the elbow point at 25\% to be the best choice.

\subsection{Clustering} 
As clustering is a critical step of our proposed method, we offer some quantitative ablations in this section. There are two hyperparameters for clustering --- the layer to apply clustering and the number of clusters $K_{c}$. As shown in Table \ref{table layers}, we first experiment with the first hyperparameter and find applying clustering after the third layer to be the best. Then we search for the best number of clusters and report the results in Table \ref{table number of clusters}. We find the best choice of the cluster number to be 4.

\subsection{Influence of the Correspondence Types} 
After attentive-inattentive separation, there are three possible correspondence types between the token pairs: attentive-attentive, attentive-inattentive and inattentive-inattentive. We study the importance of each type by setting the respective repellence hyperparameter to zero and evaluate how much the performance drops in Table \ref{table supple. each relation}. We find that the relationship between attentive-attentive tokens is the most important as the error increases the most when we don't enforce any repellence. This is expected as the attentive tokens covers most of the landmark regions and to distinguish between the different facial landmarks, the attentive-attentive relationship should be given more importance. We also find that the relation between attentive-inattentive to be more important than inattentive-inattentive. This is expected since the former may deliver intricate cues regarding the dependencies between the landmark and critical non-landmark regions such as landmark orientation (left vs.\ right) and landmark boundaries.

\begin{table}[!t]
\caption{\textbf{Landmark matching results of applying clustering after different layers.} We report the mean pixel error between the prediction and ground-truth on 1000 image pairs sampled from MAFL.}
\centering
\label{table layers}
\begin{tabular}{l|cc|l|cc}
\hline
\textbf{Layer}              & \textbf{Same}         & \textbf{Diff.}        &\textbf{Layer}              & \textbf{Same}         & \textbf{Diff.}  \rule[2.5ex]{0pt}{0pt}\rule[-1ex]{0pt}{0pt}\\
\hline
0                           & 0.30                  & 1.62                  & 6                          & 0.33                  & 1.61 \\
1                           & 0.30                  & 1.63                  & 7                          & 0.32                  & 1.64 \\
2                           & 0.27                  & 1.61                  & 8                          & 0.30                  & 1.62 \\
3                           & 0.30                  & 1.62                  & 9                          & 0.30                  & 1.63 \\
4                           & 0.31                  & 1.65                  & 10                         & 0.31                  & 1.66 \\
5                           & 0.30                  & 1.62                  & 11                         & 0.34                  & 1.66 \\
\hline
\end{tabular}
\end{table}

\begin{table}[!t]
\caption{\textbf{Landmark matching results of using different number of clusters.} We report the mean pixel error between the prediction and ground-truth on 1000 image pairs sampled from MAFL.}
\centering
\label{table number of clusters}
\renewcommand\arraystretch{1}
\begin{tabular}{l|cc}
\hline
\textbf{Number of clusters}              & \textbf{Same}         & \textbf{Diff.}    \rule[2.5ex]{0pt}{0pt}\rule[-1ex]{0pt}{0pt}\\
\hline
1                           & 0.33                  & 1.68                  \\
2                           & 0.32                  & 1.66                  \\
4                           & 0.27                  & 1.61                  \\
8                           & 0.30                  & 1.62                  \\
\hline
\end{tabular}
\end{table}

\begin{table}[t]
\caption{\textbf{Importance of each relationship between patch tokens.} We report the mean pixel error between the prediction and ground-truth on 1000 image pairs sampled from MAFL.}
\centering
\label{table supple. each relation}
\begin{tabular}{ccc|cc}
\hline
attn-attn   & attn-inattn       & inattn-inattn     & \textbf{Same}         & \textbf{Diff.}      \\
\hline
\xmark      & \cmark            & \cmark            & 0.33                  & 3.52          \\
\cmark      & \xmark            & \cmark            & 0.32                  & 1.64          \\
\cmark      & \cmark            & \xmark            & 0.29                  & 1.62          \\
\hline
\cmark      & \cmark            & \cmark            & 0.27                  & 1.61         \\
\hline
\end{tabular}
\end{table}

\section{More Visualizations}
\subsection{Visualization of Landmark Similarity Map}
\begin{figure*}[!t]
\begin{tabular}{c|ccccc|ccccc}
      \includegraphics[align=c, width=0.065\linewidth]{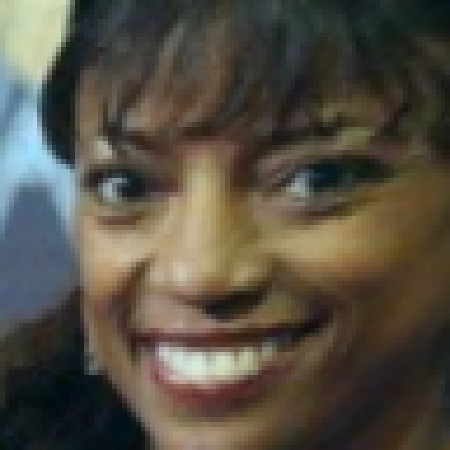}
    & \includegraphics[align=c, width=0.065\linewidth]{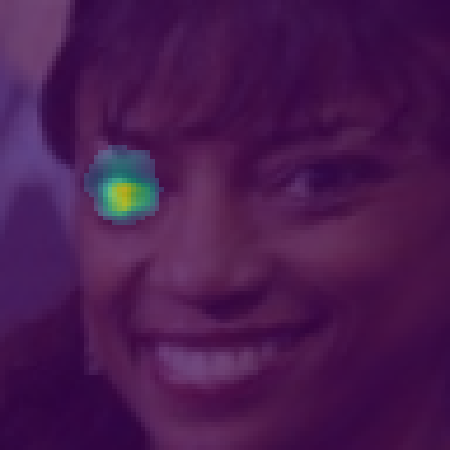}
    & \includegraphics[align=c, width=0.065\linewidth]{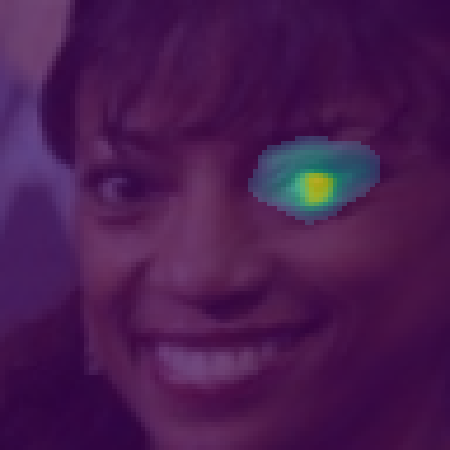}
    & \includegraphics[align=c, width=0.065\linewidth]{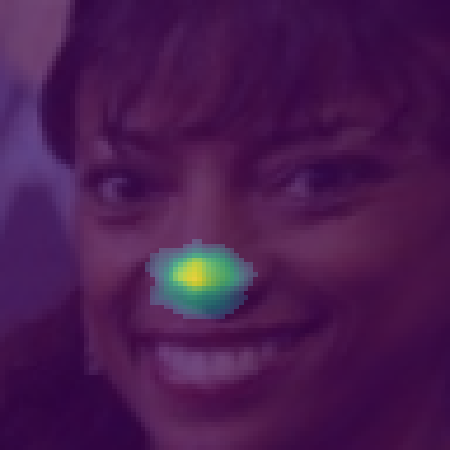}
    & \includegraphics[align=c, width=0.065\linewidth]{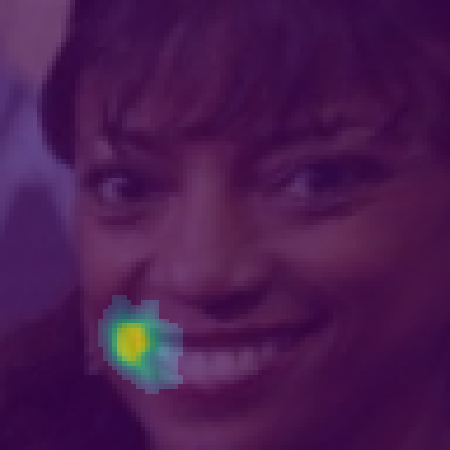}
    & \includegraphics[align=c, width=0.065\linewidth]{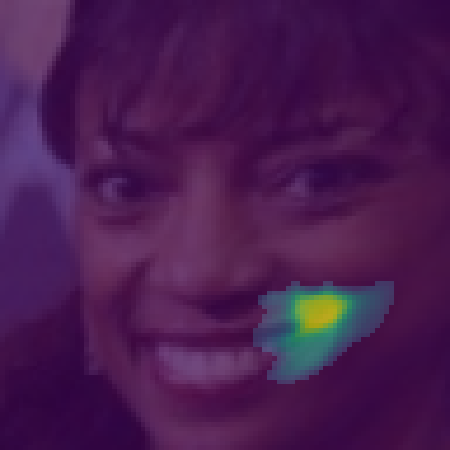}
    
    & \includegraphics[align=c, width=0.065\linewidth]{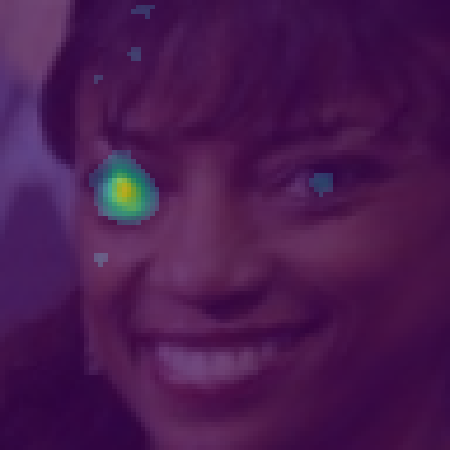}
    & \includegraphics[align=c, width=0.065\linewidth]{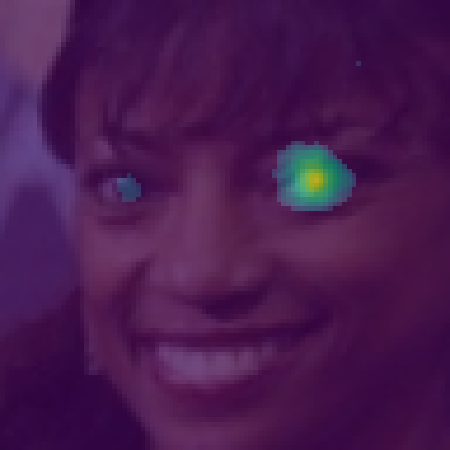}
    & \includegraphics[align=c, width=0.065\linewidth]{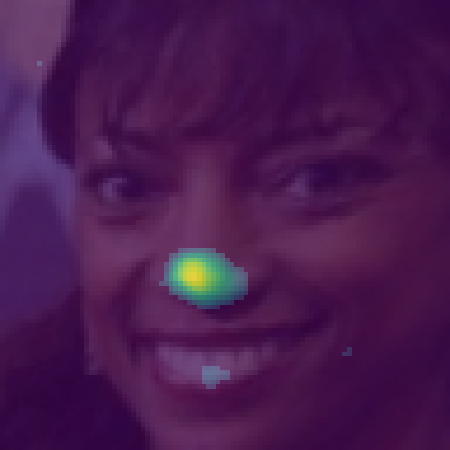}
    & \includegraphics[align=c, width=0.065\linewidth]{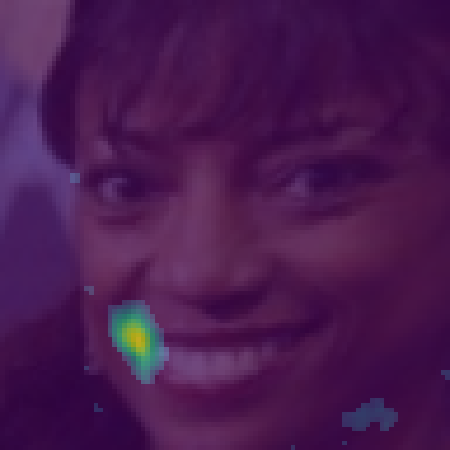}
    & \includegraphics[align=c, width=0.065\linewidth]{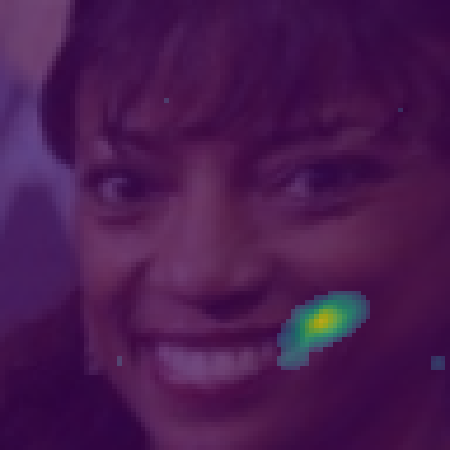} \rule[5ex]{0pt}{0pt}\rule[-4ex]{0pt}{0pt}\\
    
      \includegraphics[align=c, width=0.065\linewidth]{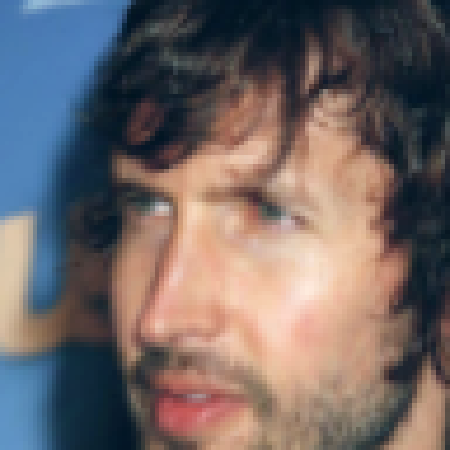}
    & \includegraphics[align=c, width=0.065\linewidth]{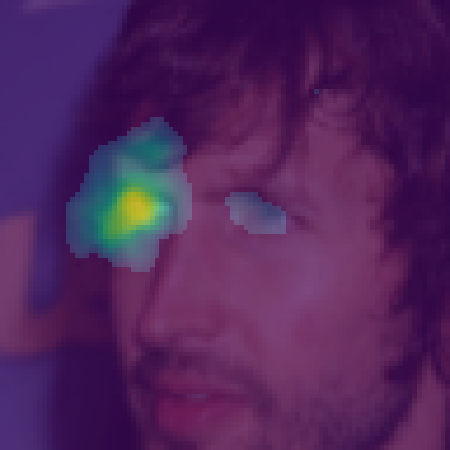}
    & \includegraphics[align=c, width=0.065\linewidth]{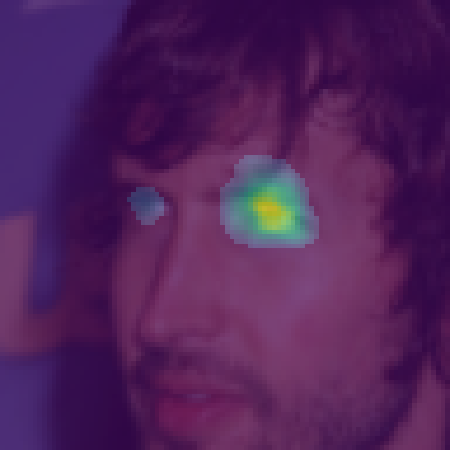}
    & \includegraphics[align=c, width=0.065\linewidth]{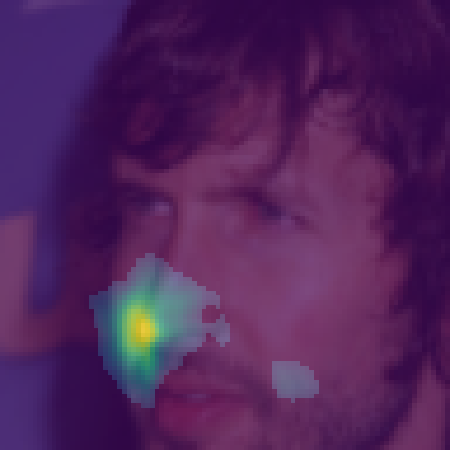}
    & \includegraphics[align=c, width=0.065\linewidth]{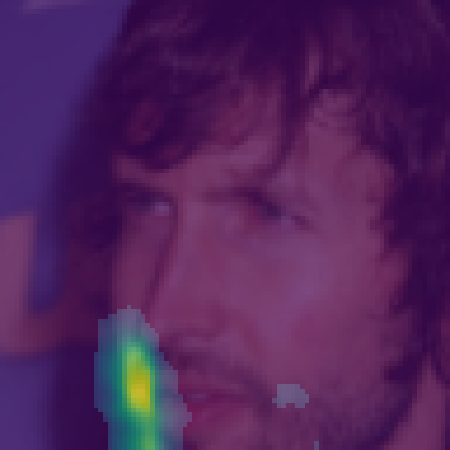}
    & \includegraphics[align=c, width=0.065\linewidth]{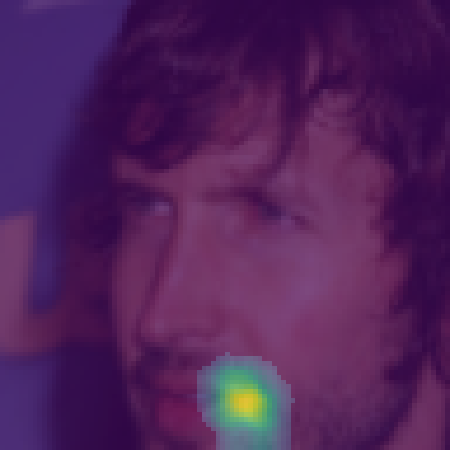}
    
    & \includegraphics[align=c, width=0.065\linewidth]{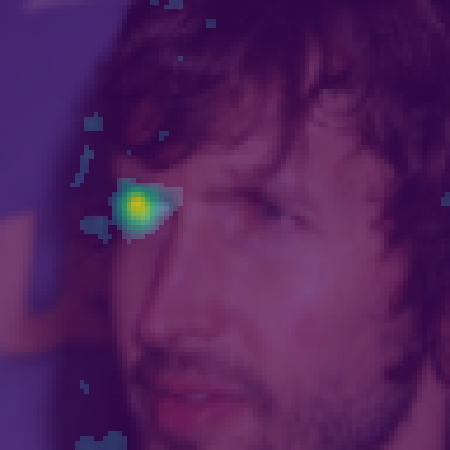}
    & \includegraphics[align=c, width=0.065\linewidth]{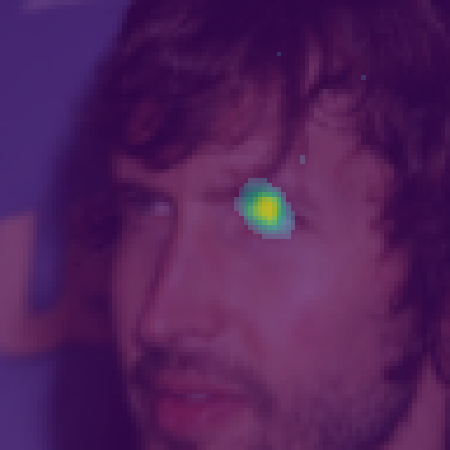}
    & \includegraphics[align=c, width=0.065\linewidth]{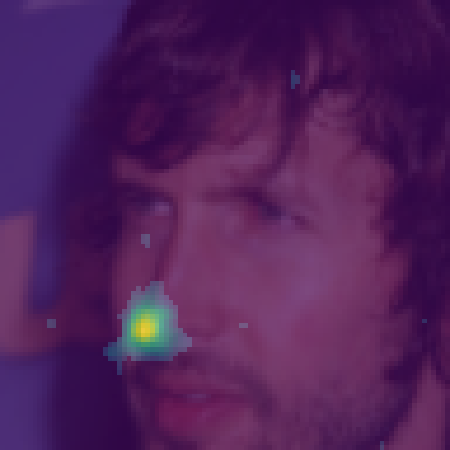}
    & \includegraphics[align=c, width=0.065\linewidth]{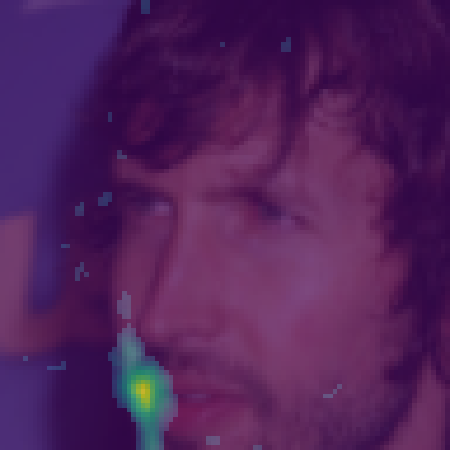}
    & \includegraphics[align=c, width=0.065\linewidth]{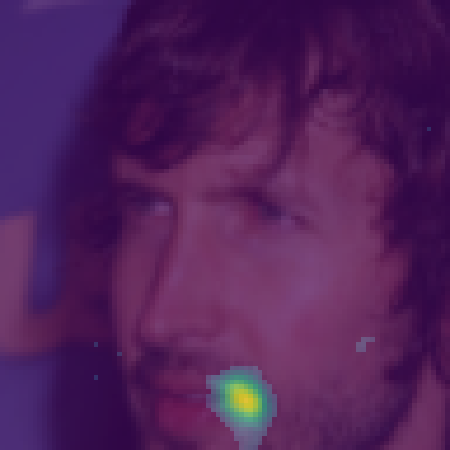} \rule[5ex]{0pt}{0pt}\rule[-4ex]{0pt}{0pt}\\
    
      \includegraphics[align=c, width=0.065\linewidth]{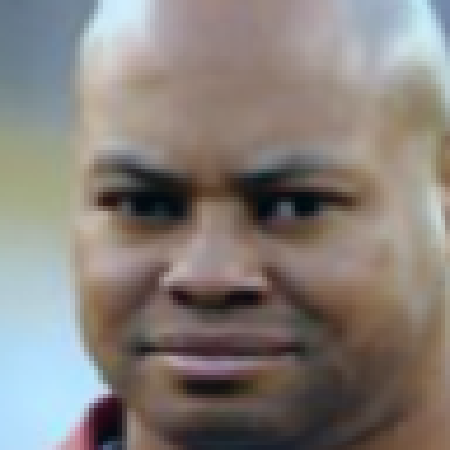}
    & \includegraphics[align=c, width=0.065\linewidth]{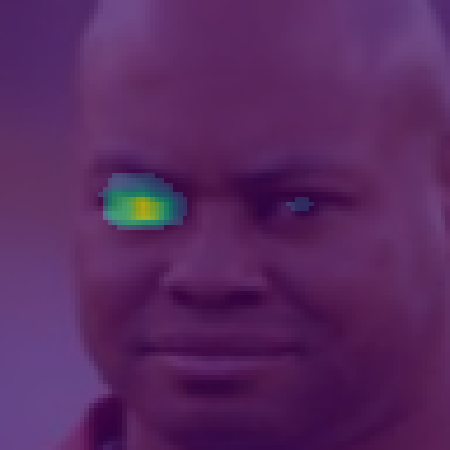}
    & \includegraphics[align=c, width=0.065\linewidth]{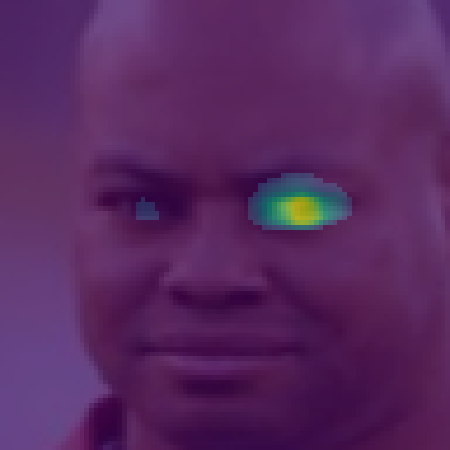}
    & \includegraphics[align=c, width=0.065\linewidth]{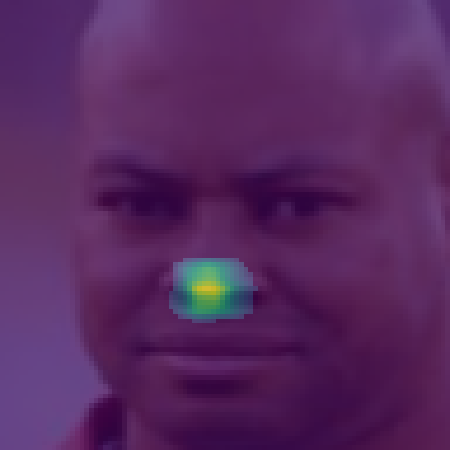}
    & \includegraphics[align=c, width=0.065\linewidth]{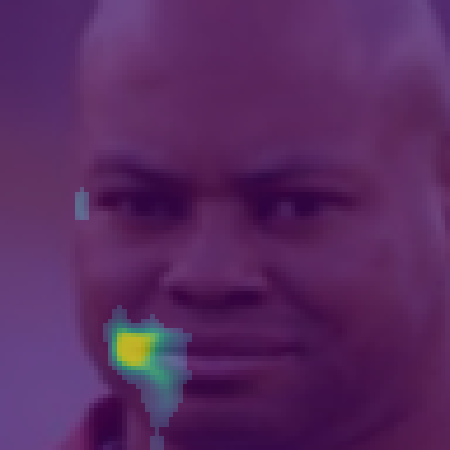}
    & \includegraphics[align=c, width=0.065\linewidth]{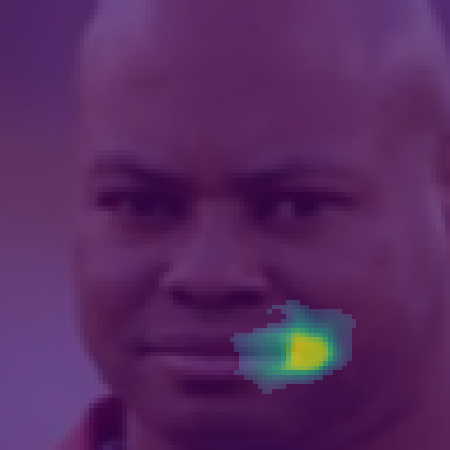}
    
    & \includegraphics[align=c, width=0.065\linewidth]{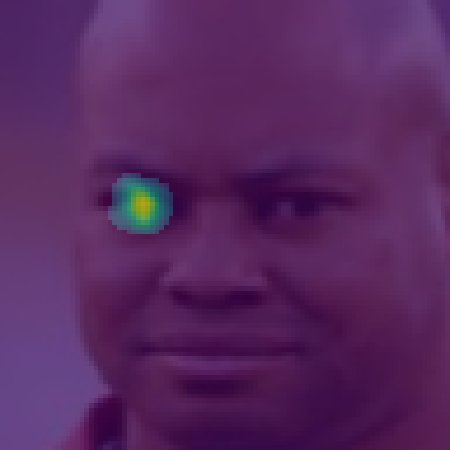}
    & \includegraphics[align=c, width=0.065\linewidth]{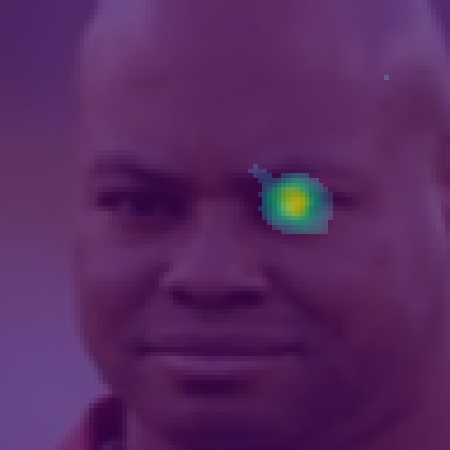}
    & \includegraphics[align=c, width=0.065\linewidth]{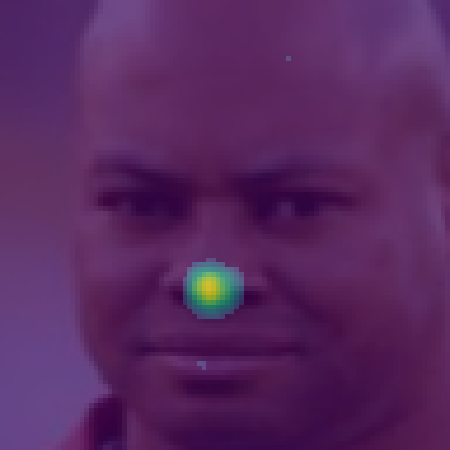}
    & \includegraphics[align=c, width=0.065\linewidth]{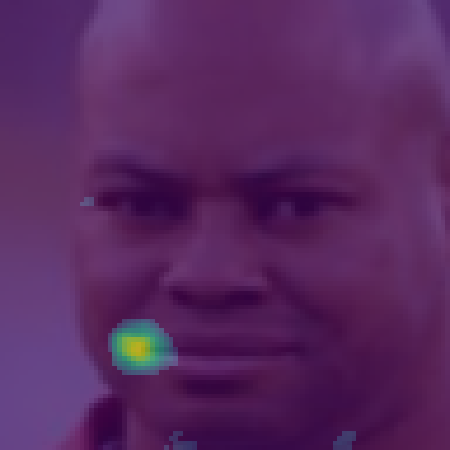}
    & \includegraphics[align=c, width=0.065\linewidth]{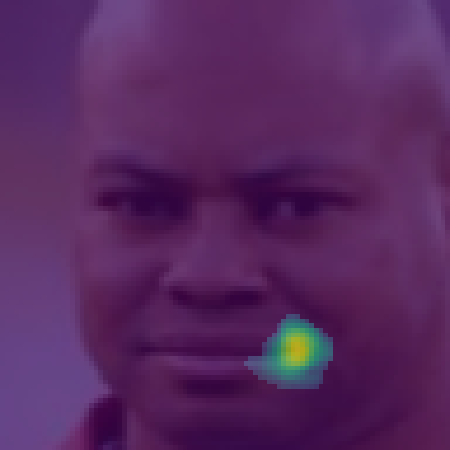} \rule[5ex]{0pt}{0pt}\rule[-4ex]{0pt}{0pt}\\

      \includegraphics[align=c, width=0.065\linewidth]{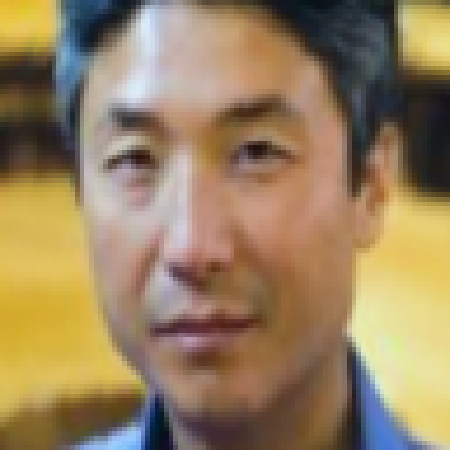}
    & \includegraphics[align=c, width=0.065\linewidth]{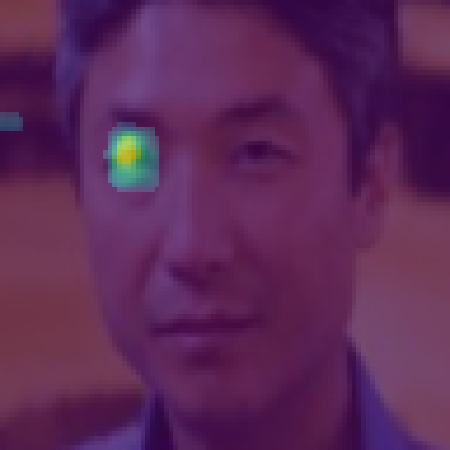}
    & \includegraphics[align=c, width=0.065\linewidth]{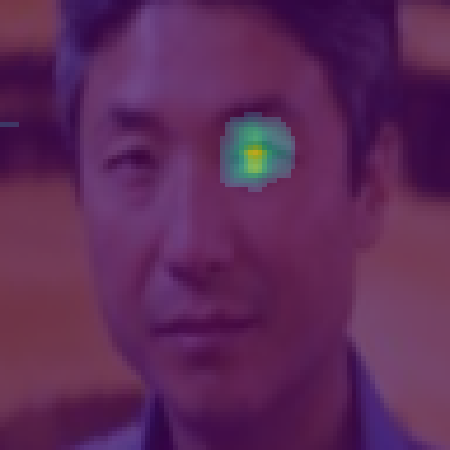}
    & \includegraphics[align=c, width=0.065\linewidth]{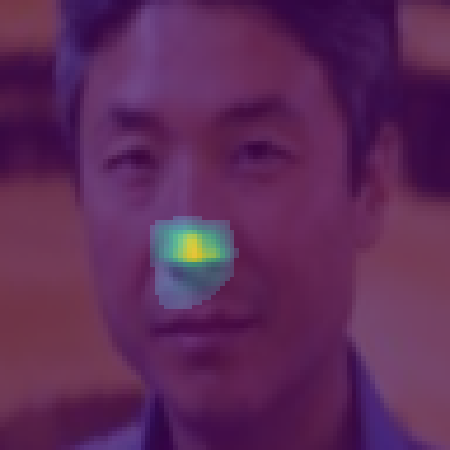}
    & \includegraphics[align=c, width=0.065\linewidth]{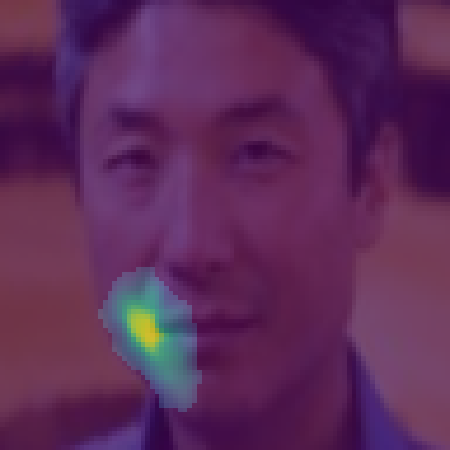}
    & \includegraphics[align=c, width=0.065\linewidth]{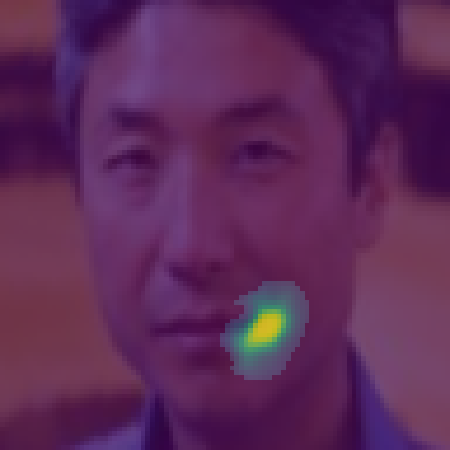}
    
    & \includegraphics[align=c, width=0.065\linewidth]{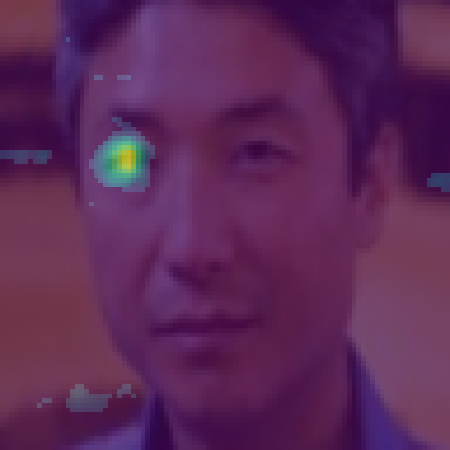}
    & \includegraphics[align=c, width=0.065\linewidth]{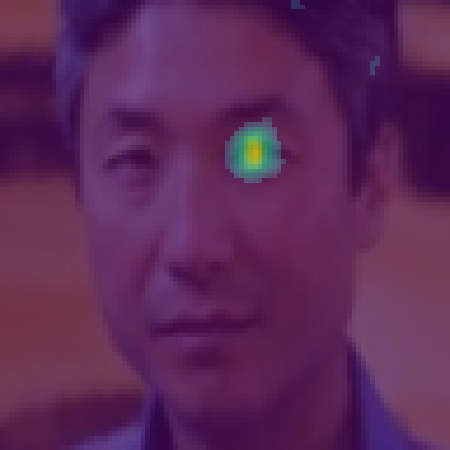}
    & \includegraphics[align=c, width=0.065\linewidth]{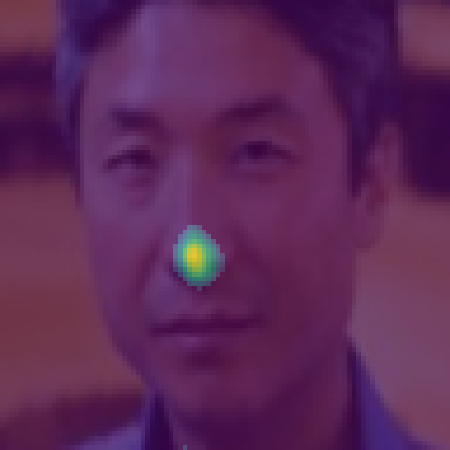}
    & \includegraphics[align=c, width=0.065\linewidth]{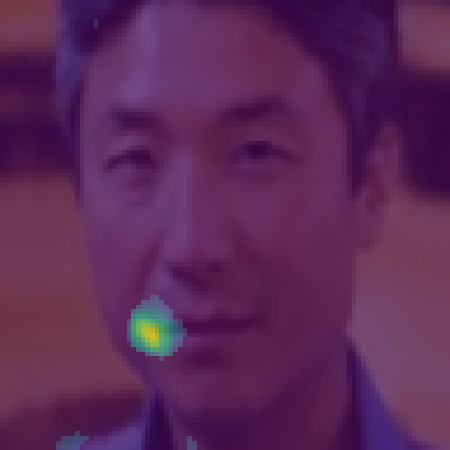}
    & \includegraphics[align=c, width=0.065\linewidth]{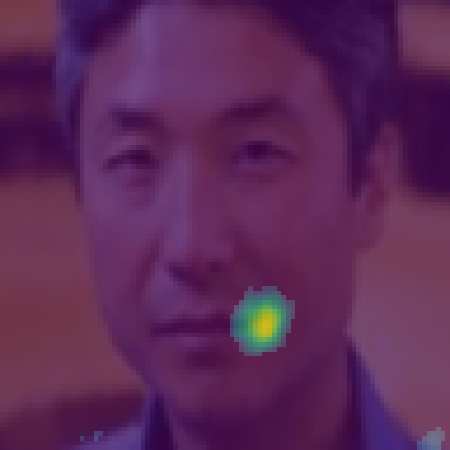} \rule[5ex]{0pt}{0pt}\rule[-4ex]{0pt}{0pt}\\

      \includegraphics[align=c, width=0.065\linewidth]{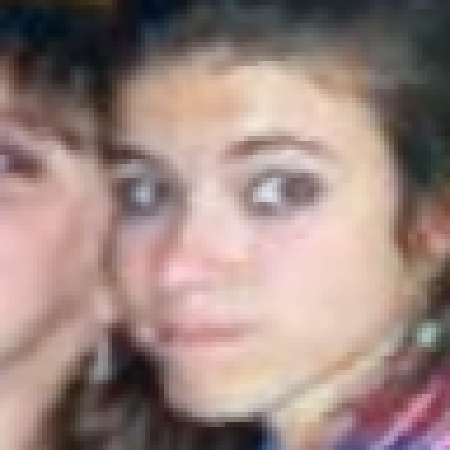}
    & \includegraphics[align=c, width=0.065\linewidth]{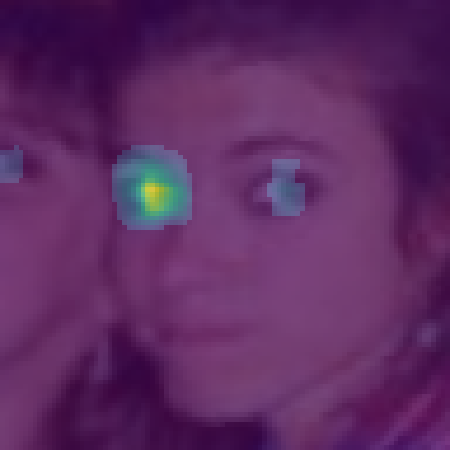}
    & \includegraphics[align=c, width=0.065\linewidth]{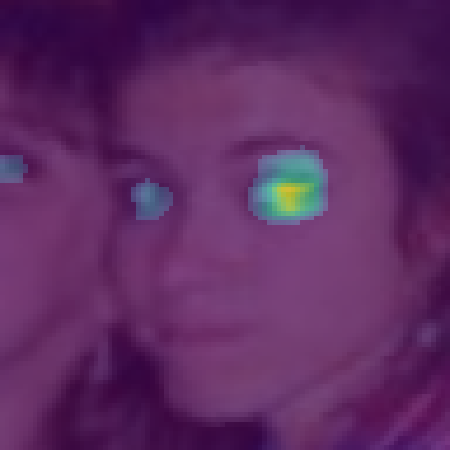}
    & \includegraphics[align=c, width=0.065\linewidth]{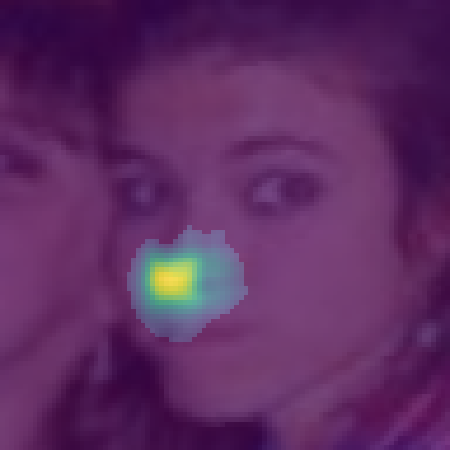}
    & \includegraphics[align=c, width=0.065\linewidth]{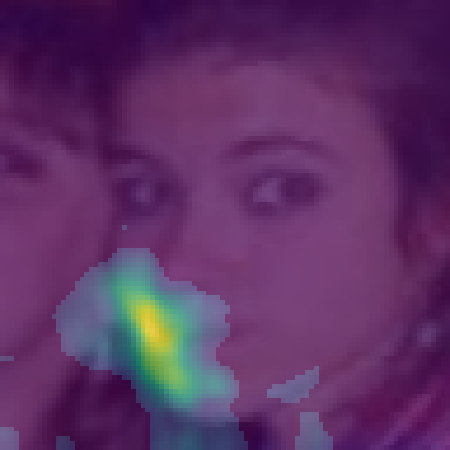}
    & \includegraphics[align=c, width=0.065\linewidth]{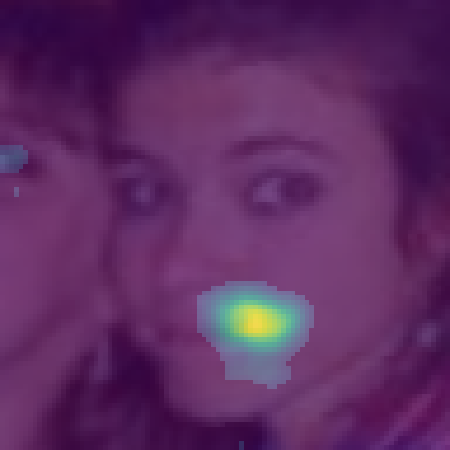}
    
    & \includegraphics[align=c, width=0.065\linewidth]{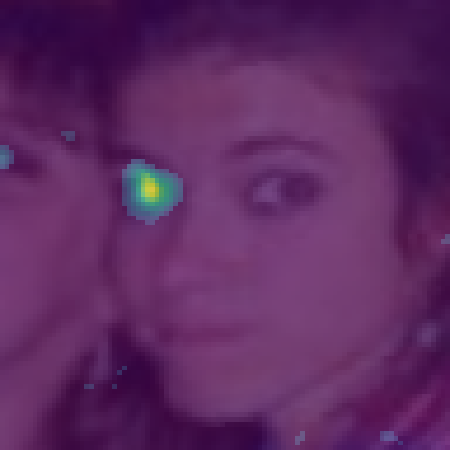}
    & \includegraphics[align=c, width=0.065\linewidth]{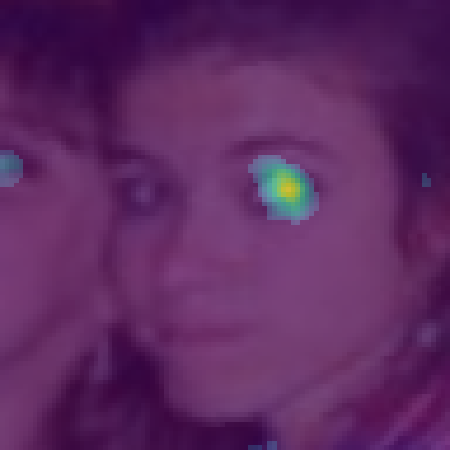}
    & \includegraphics[align=c, width=0.065\linewidth]{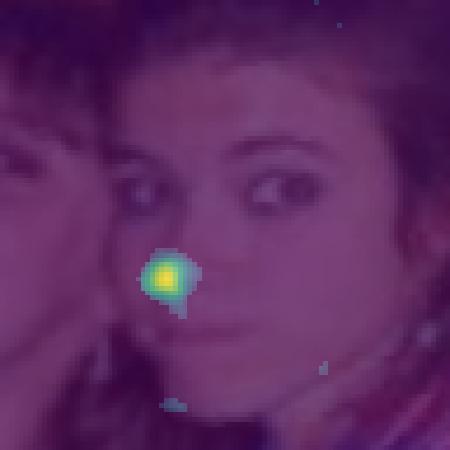}
    & \includegraphics[align=c, width=0.065\linewidth]{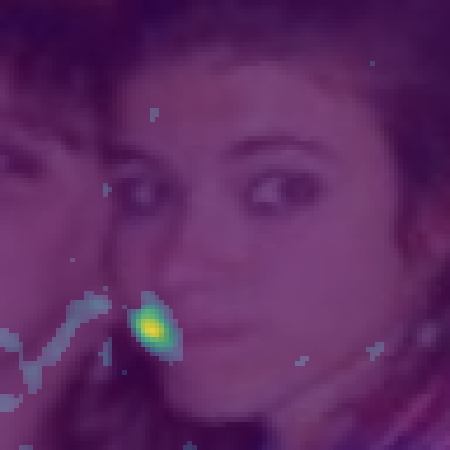}
    & \includegraphics[align=c, width=0.065\linewidth]{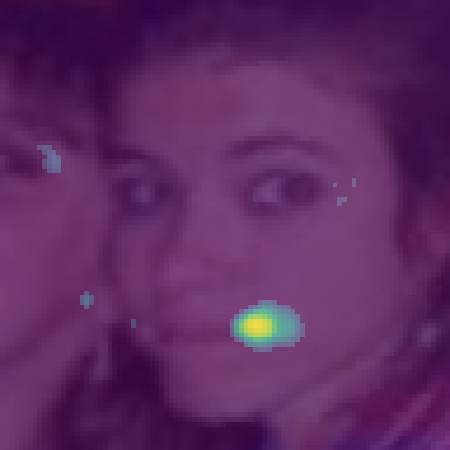} \rule[5ex]{0pt}{0pt}\rule[-4ex]{0pt}{0pt}\\
    \hline

      \includegraphics[align=c, width=0.065\linewidth]{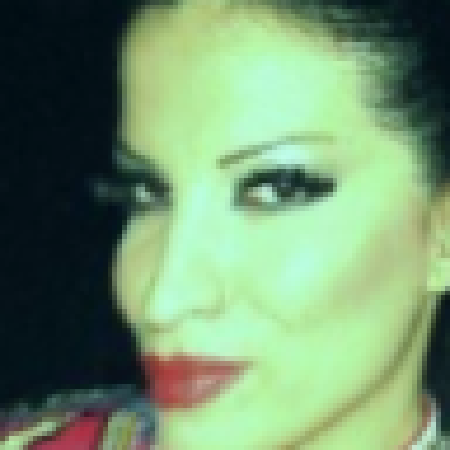}
    & \includegraphics[align=c, width=0.065\linewidth]{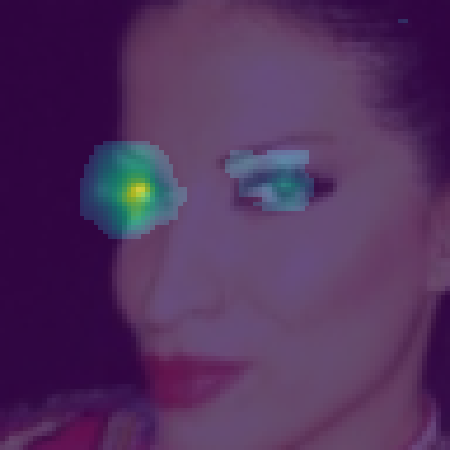}
    & \includegraphics[align=c, width=0.065\linewidth]{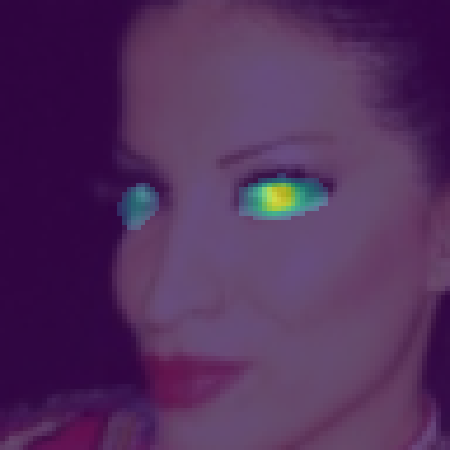}
    & \includegraphics[align=c, width=0.065\linewidth]{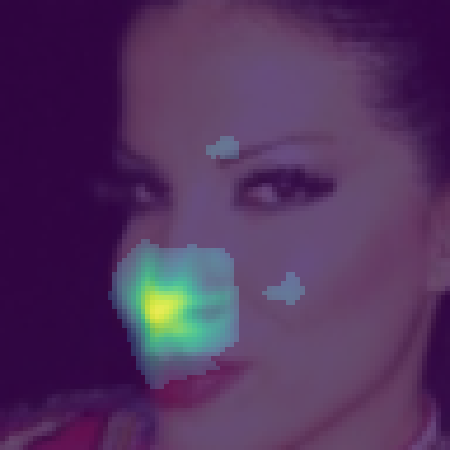}
    & \includegraphics[align=c, width=0.065\linewidth]{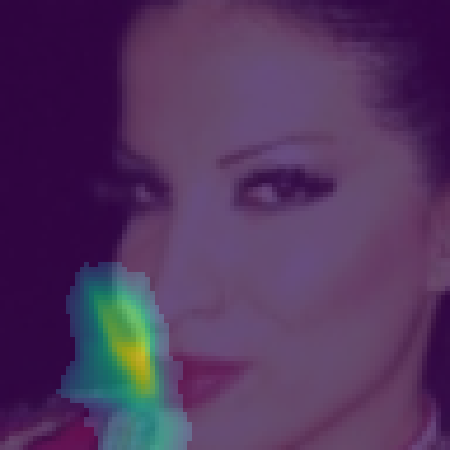}
    & \includegraphics[align=c, width=0.065\linewidth]{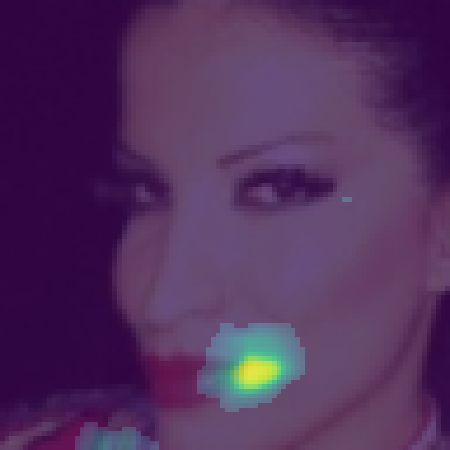}
    
    & \includegraphics[align=c, width=0.065\linewidth]{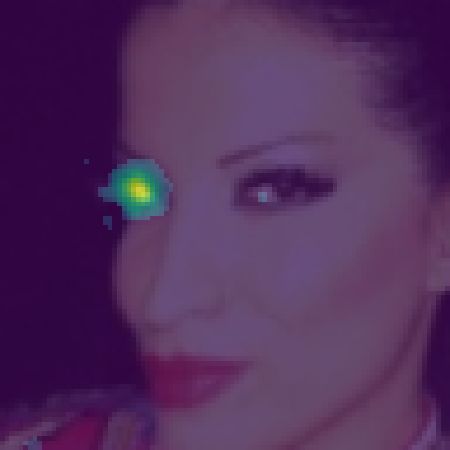}
    & \includegraphics[align=c, width=0.065\linewidth]{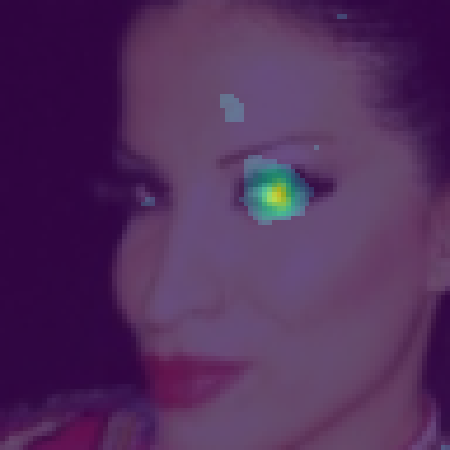}
    & \includegraphics[align=c, width=0.065\linewidth]{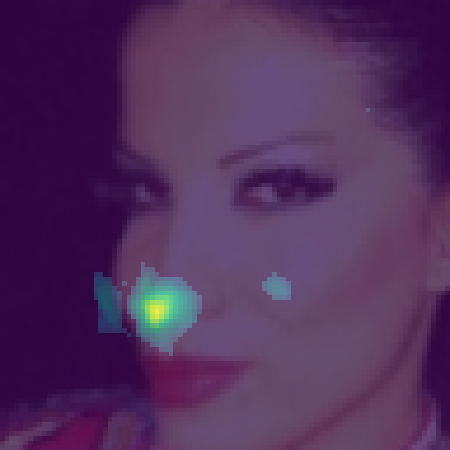}
    & \includegraphics[align=c, width=0.065\linewidth]{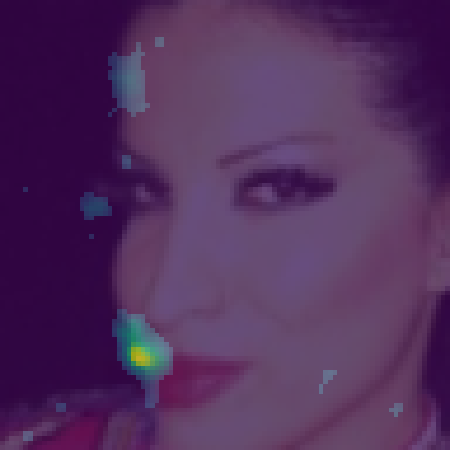}
    & \includegraphics[align=c, width=0.065\linewidth]{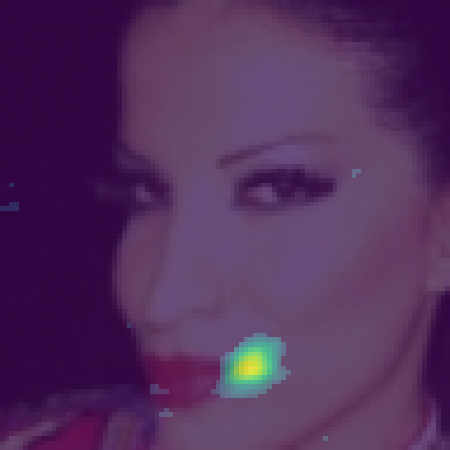} \rule[5ex]{0pt}{0pt}\rule[-4ex]{0pt}{0pt}\\

      \includegraphics[align=c, width=0.065\linewidth]{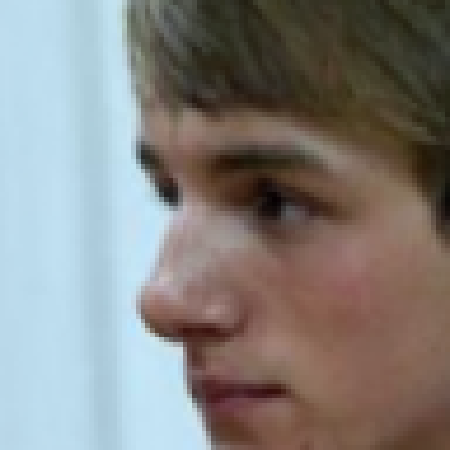}
    & \includegraphics[align=c, width=0.065\linewidth]{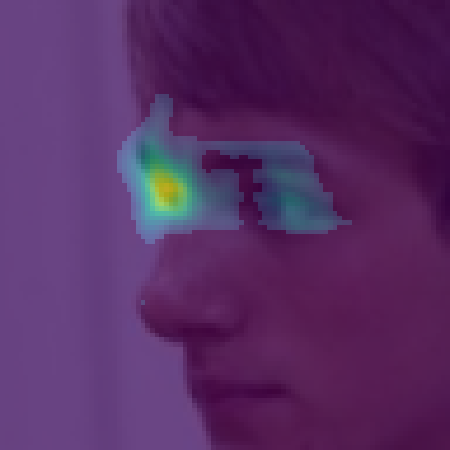}
    & \includegraphics[align=c, width=0.065\linewidth]{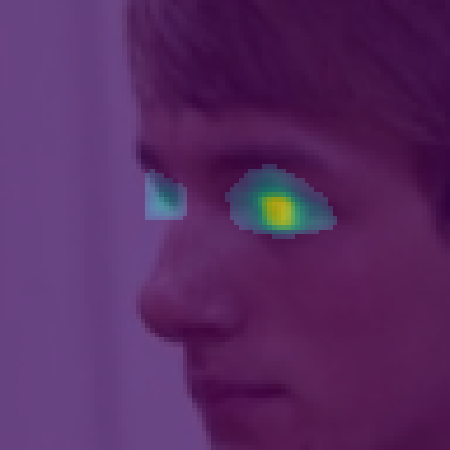}
    & \includegraphics[align=c, width=0.065\linewidth]{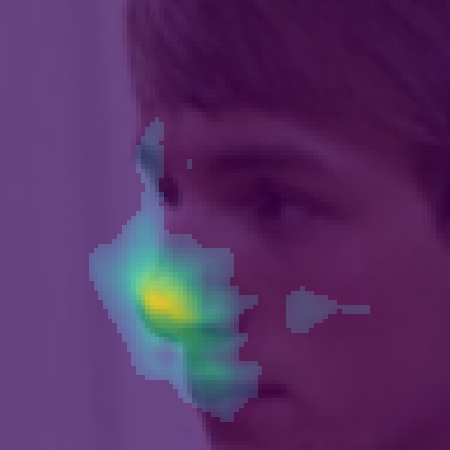}
    & \includegraphics[align=c, width=0.065\linewidth]{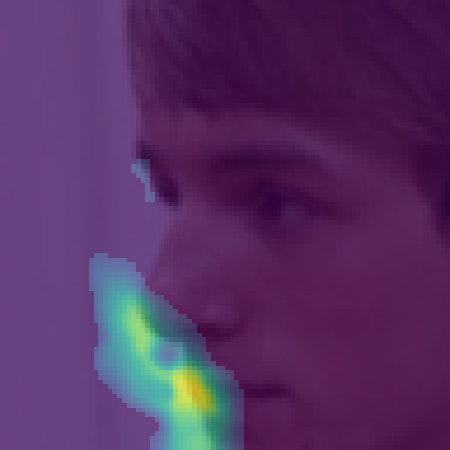}
    & \includegraphics[align=c, width=0.065\linewidth]{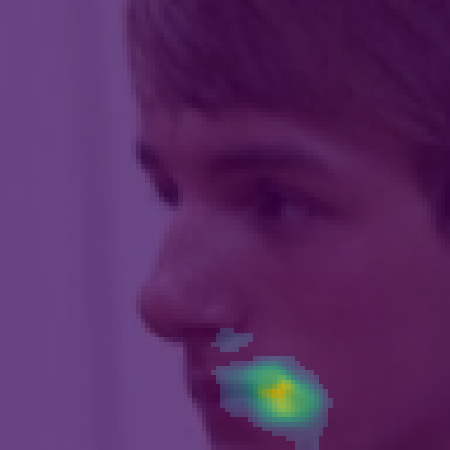}
    
    & \includegraphics[align=c, width=0.065\linewidth]{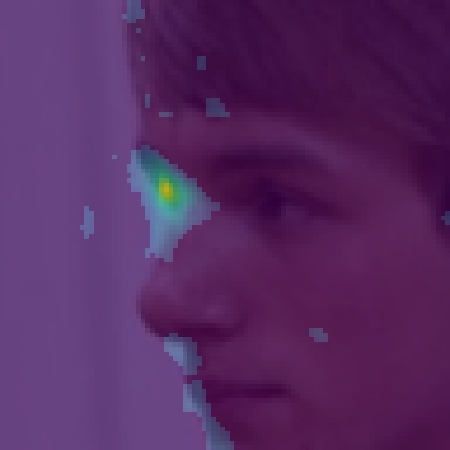}
    & \includegraphics[align=c, width=0.065\linewidth]{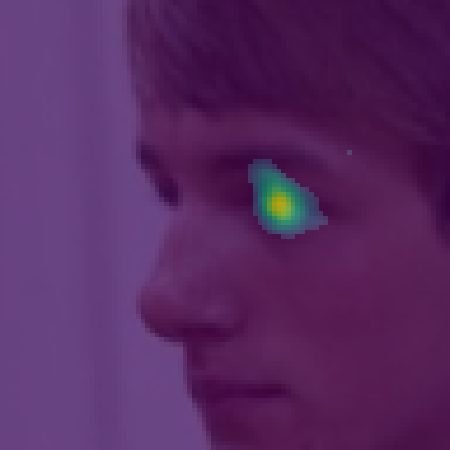}
    & \includegraphics[align=c, width=0.065\linewidth]{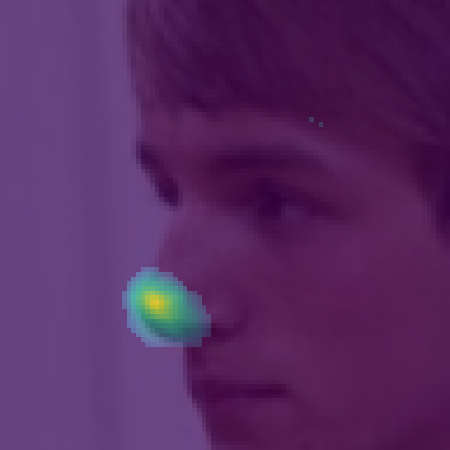}
    & \includegraphics[align=c, width=0.065\linewidth]{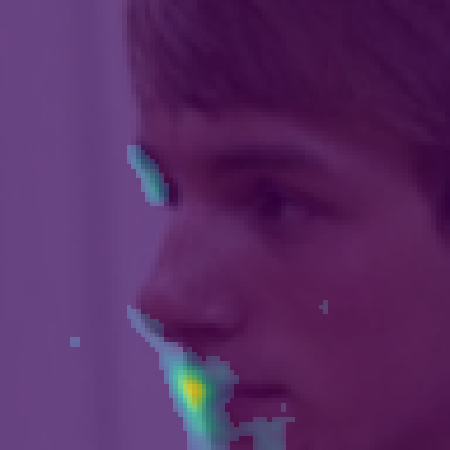}
    & \includegraphics[align=c, width=0.065\linewidth]{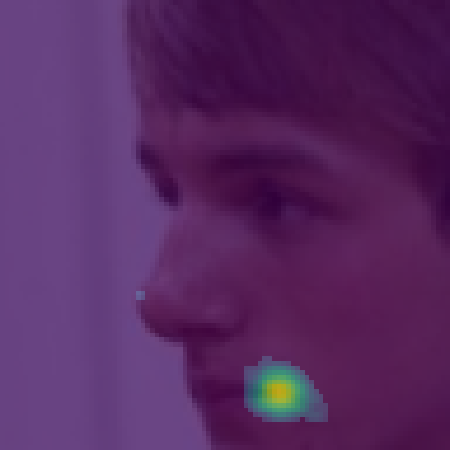} \rule[5ex]{0pt}{0pt}\rule[-4ex]{0pt}{0pt}\\

      \includegraphics[align=c, width=0.065\linewidth]{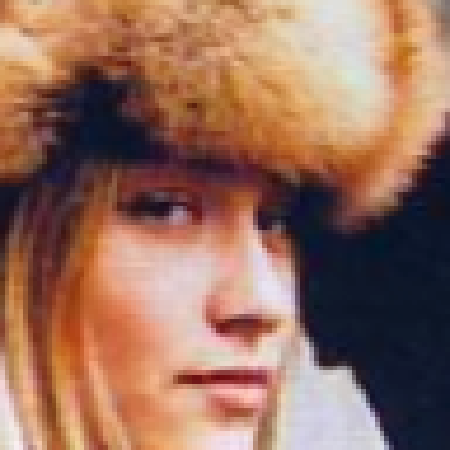}
    & \includegraphics[align=c, width=0.065\linewidth]{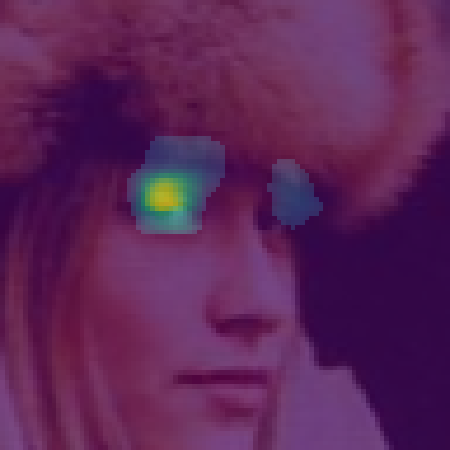}
    & \includegraphics[align=c, width=0.065\linewidth]{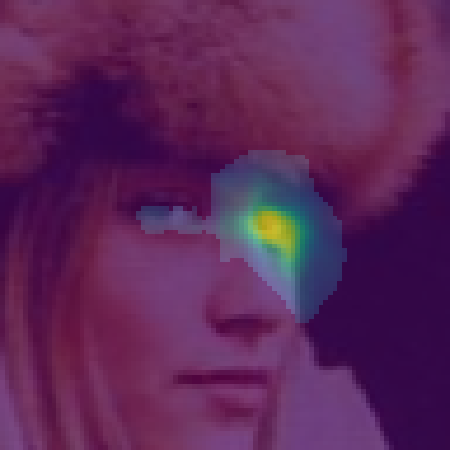}
    & \includegraphics[align=c, width=0.065\linewidth]{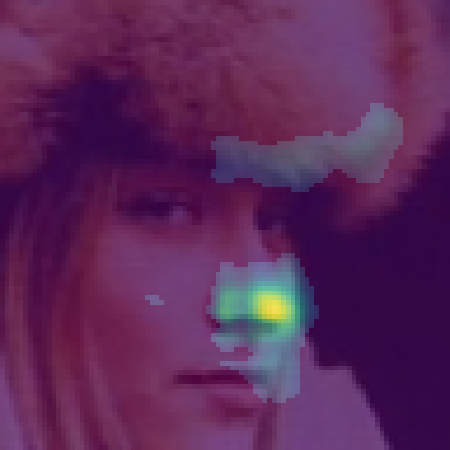}
    & \includegraphics[align=c, width=0.065\linewidth]{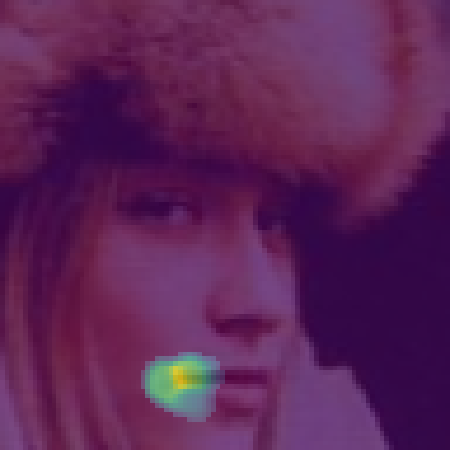}
    & \includegraphics[align=c, width=0.065\linewidth]{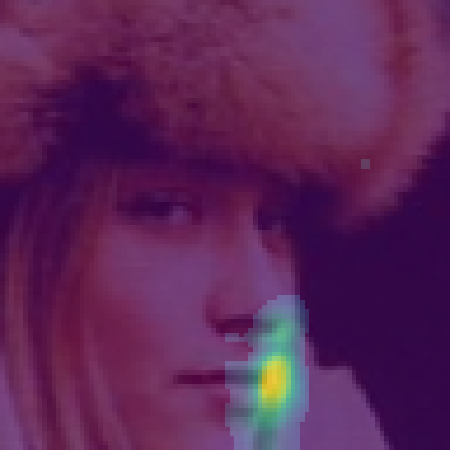}
    
    & \includegraphics[align=c, width=0.065\linewidth]{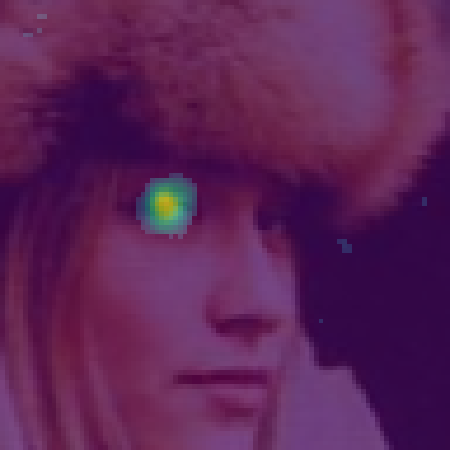}
    & \includegraphics[align=c, width=0.065\linewidth]{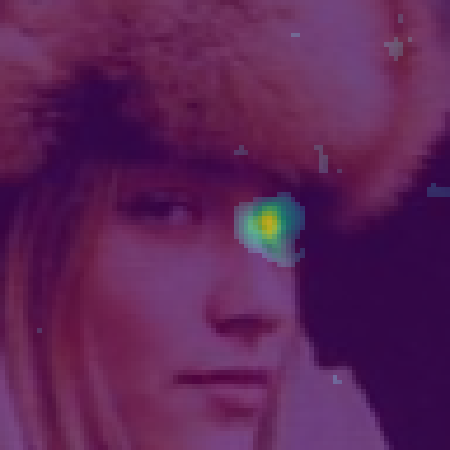}
    & \includegraphics[align=c, width=0.065\linewidth]{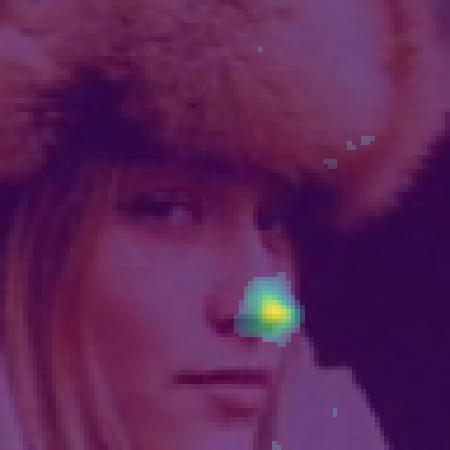}
    & \includegraphics[align=c, width=0.065\linewidth]{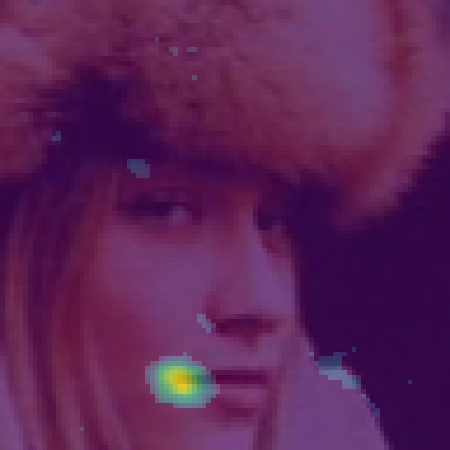}
    & \includegraphics[align=c, width=0.065\linewidth]{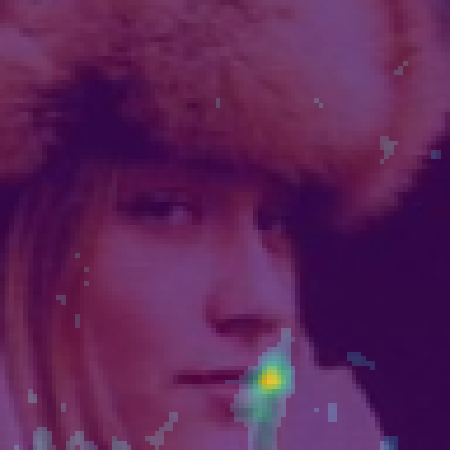} \rule[5ex]{0pt}{0pt}\rule[-4ex]{0pt}{0pt}\\
    \hline

      \includegraphics[align=c, width=0.065\linewidth]{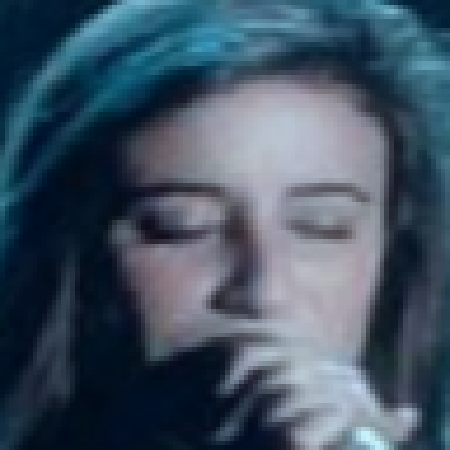}
    & \includegraphics[align=c, width=0.065\linewidth]{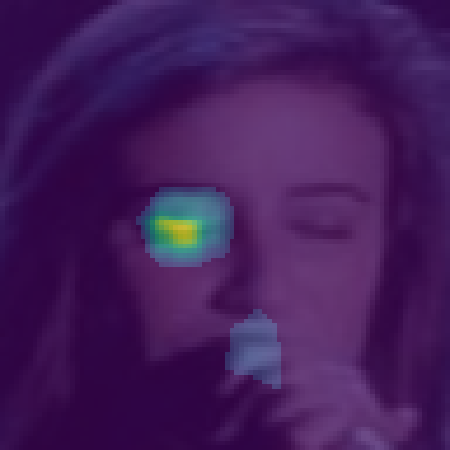}
    & \includegraphics[align=c, width=0.065\linewidth]{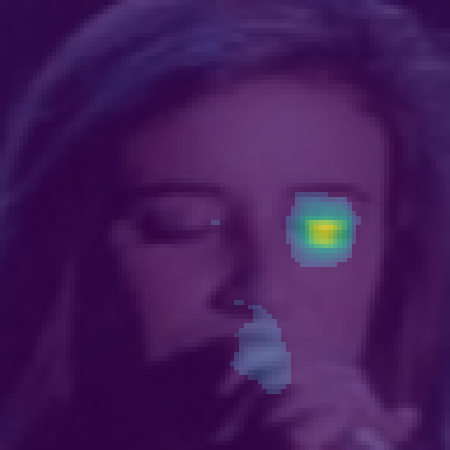}
    & \includegraphics[align=c, width=0.065\linewidth]{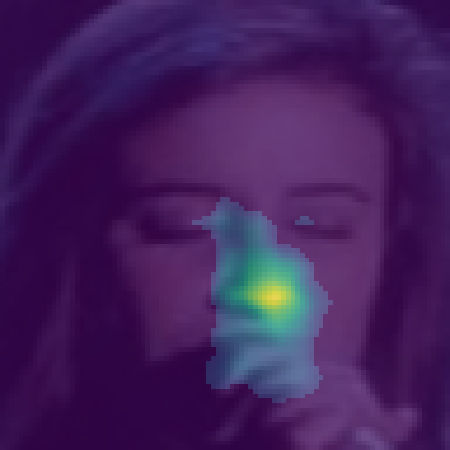}
    & \includegraphics[align=c, width=0.065\linewidth]{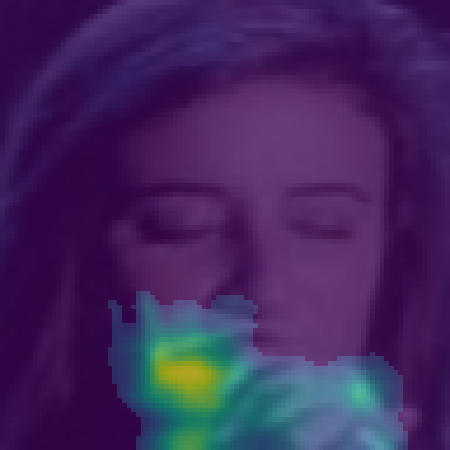}
    & \includegraphics[align=c, width=0.065\linewidth]{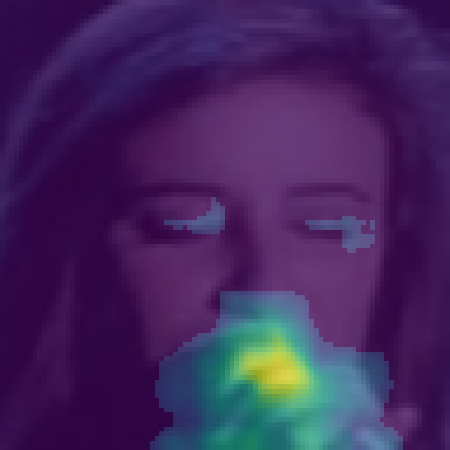}
    
    & \includegraphics[align=c, width=0.065\linewidth]{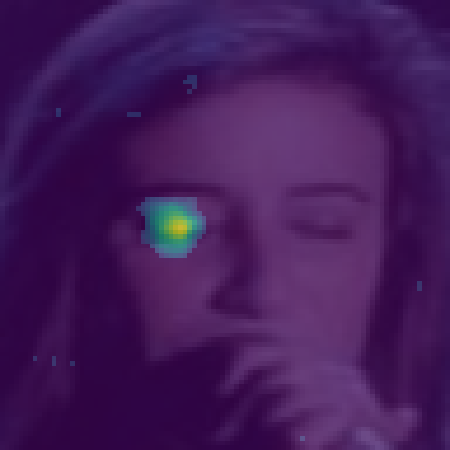}
    & \includegraphics[align=c, width=0.065\linewidth]{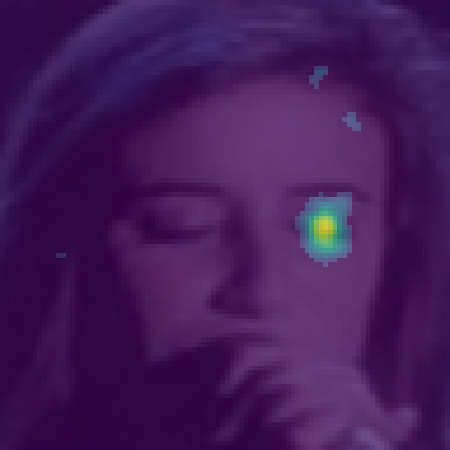}
    & \includegraphics[align=c, width=0.065\linewidth]{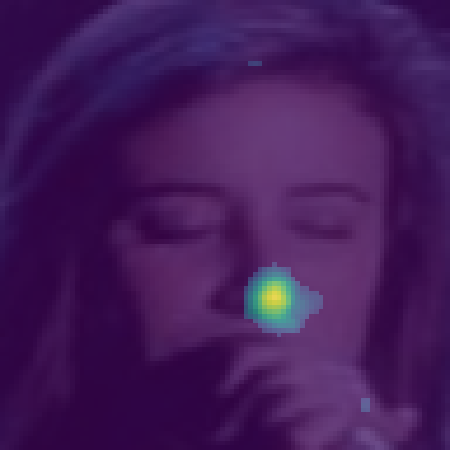}
    & \includegraphics[align=c, width=0.065\linewidth]{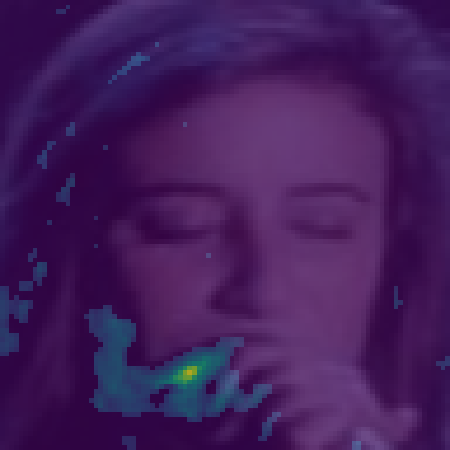}
    & \includegraphics[align=c, width=0.065\linewidth]{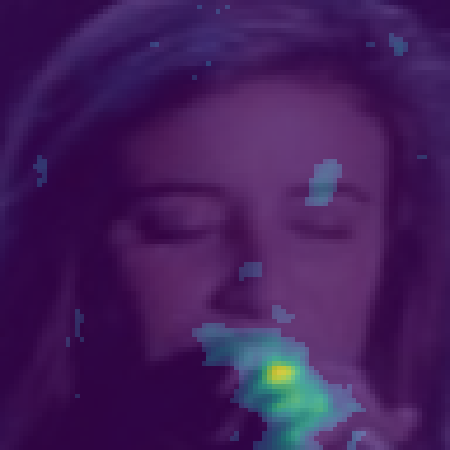} \rule[5ex]{0pt}{0pt}\rule[-4ex]{0pt}{0pt}\\

      \includegraphics[align=c, width=0.065\linewidth]{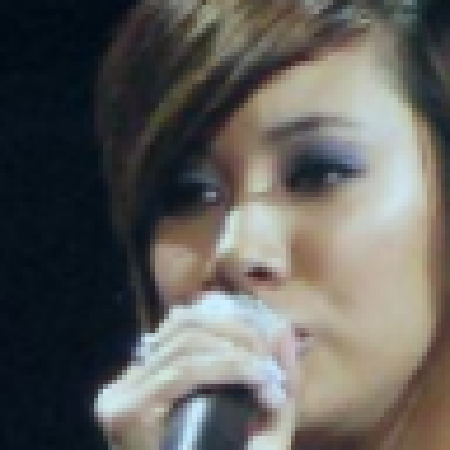}
    & \includegraphics[align=c, width=0.065\linewidth]{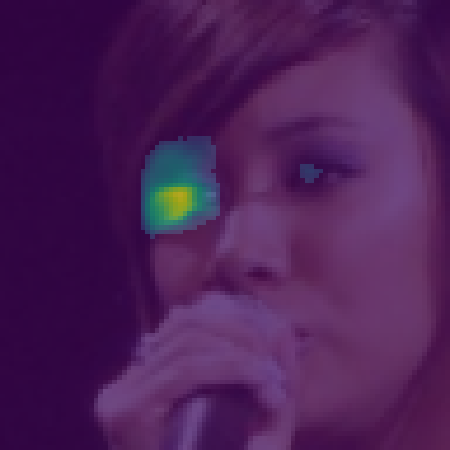}
    & \includegraphics[align=c, width=0.065\linewidth]{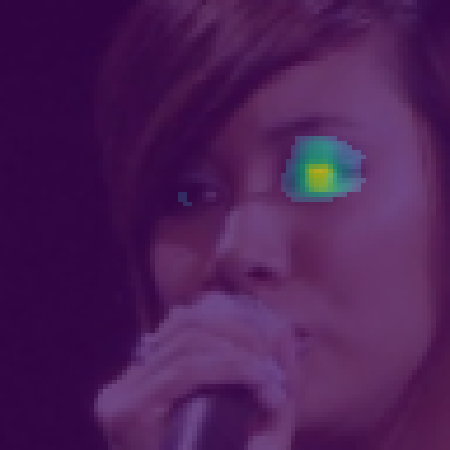}
    & \includegraphics[align=c, width=0.065\linewidth]{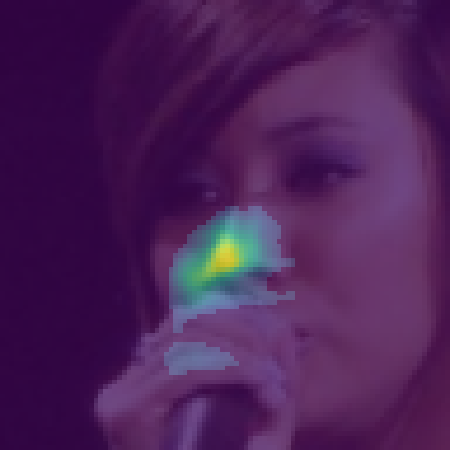}
    & \includegraphics[align=c, width=0.065\linewidth]{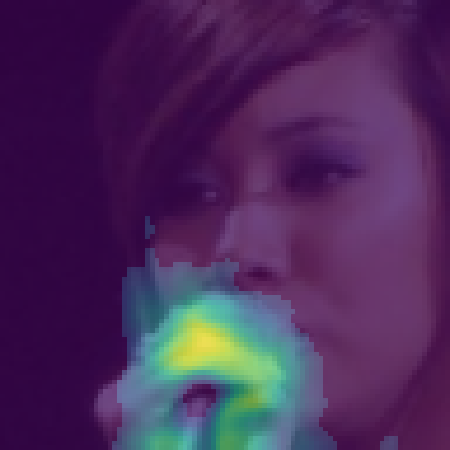}
    & \includegraphics[align=c, width=0.065\linewidth]{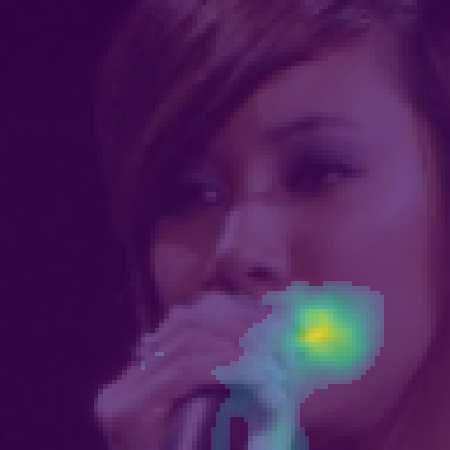}
    
    & \includegraphics[align=c, width=0.065\linewidth]{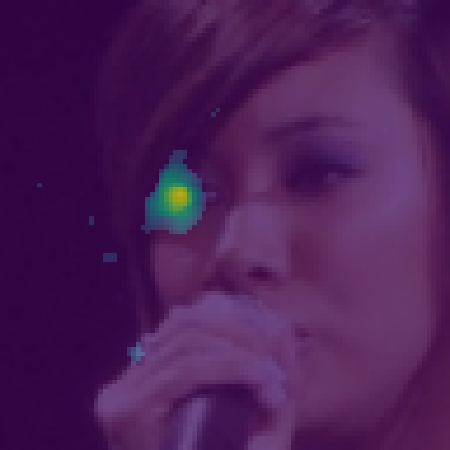}
    & \includegraphics[align=c, width=0.065\linewidth]{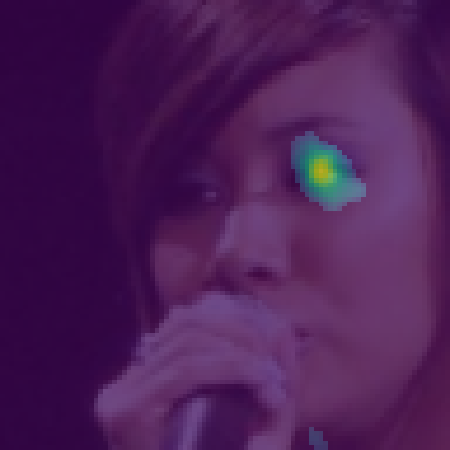}
    & \includegraphics[align=c, width=0.065\linewidth]{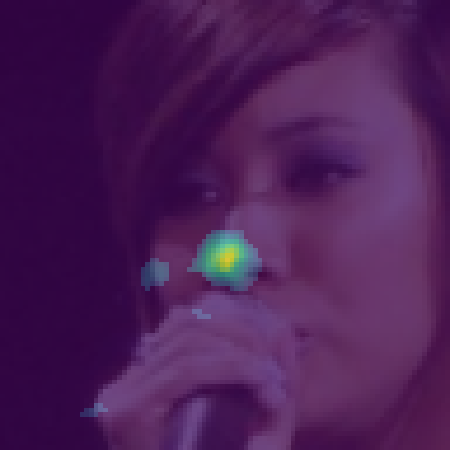}
    & \includegraphics[align=c, width=0.065\linewidth]{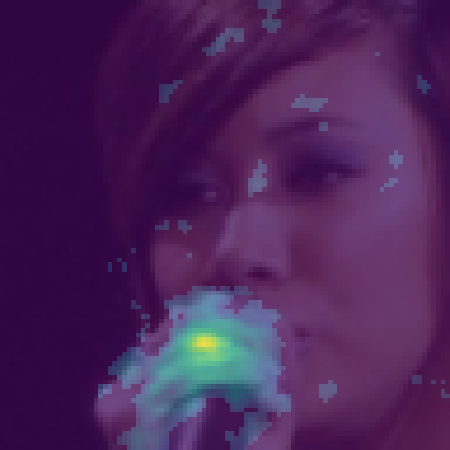}
    & \includegraphics[align=c, width=0.065\linewidth]{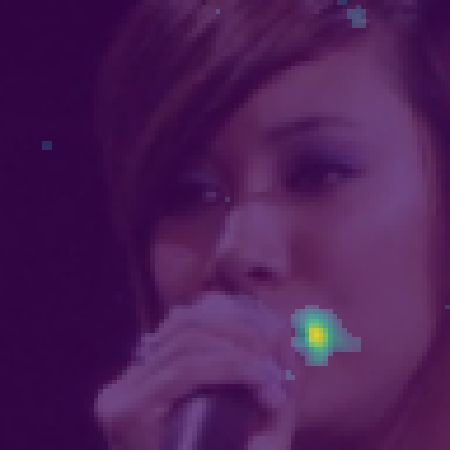} \rule[5ex]{0pt}{0pt}\rule[-4ex]{0pt}{0pt}\\
    \hline
    Original & \multicolumn{5}{c}{CL\cite{ContrastLandmark}} & \multicolumn{5}{c}{SCE-MAE (Ours)}
\end{tabular}
\centering
\caption{\textbf{Visualization of landmark similarity map.} We show the original images in the leftmost column, similarity map generated by CL in the middle and ours results on the right. We show the similarity map for each landmark and ours results are better localized at the corresponding landmark. }
\label{fig visualization of landmark similarity map}
\end{figure*}

We visualize some of the landmark similarity maps in Figure \ref{fig visualization of landmark similarity map}. We first obtain the dense feature map from each compared method, and then computes the cosine similarity between the landmark representation and the entire feature map. We also group the results based on the property of the original image --- front faces are shown in the upper rows, side faces are shown in the middle and the occluded faces are shown in the bottom. When there is occlusion or we can only see one side of the face, it is visibly difficult for the network to output discriminative representations for the occluded landmarks. As shown in Figure \ref{fig visualization of landmark similarity map}, our method generates sharper and more localized similarity map than prior arts.

\subsection{Qualitative Results on Landmark Detection}
\begin{figure*}[!t]
\begin{tabular}{l|ccccccc}
    MAFL
    & \includegraphics[align=c, width=0.1\linewidth, height=0.1\linewidth]{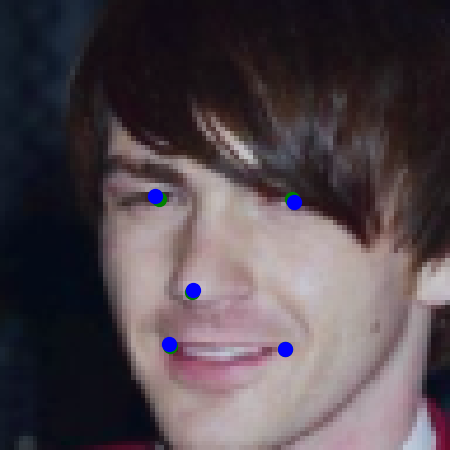}
    & \includegraphics[align=c, width=0.1\linewidth, height=0.1\linewidth]{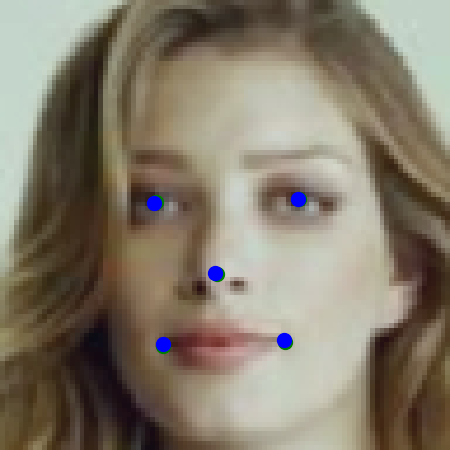}
    & \includegraphics[align=c, width=0.1\linewidth, height=0.1\linewidth]{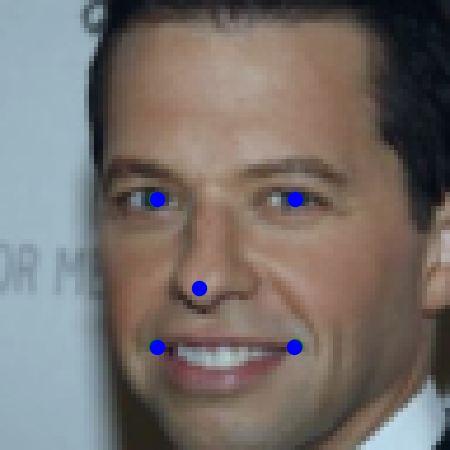}
    & \includegraphics[align=c, width=0.1\linewidth, height=0.1\linewidth]{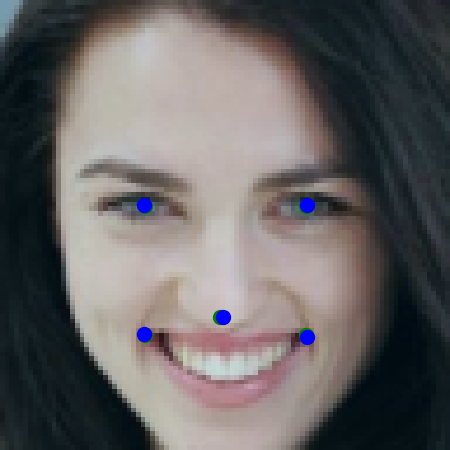}
    & \includegraphics[align=c, width=0.1\linewidth, height=0.1\linewidth]{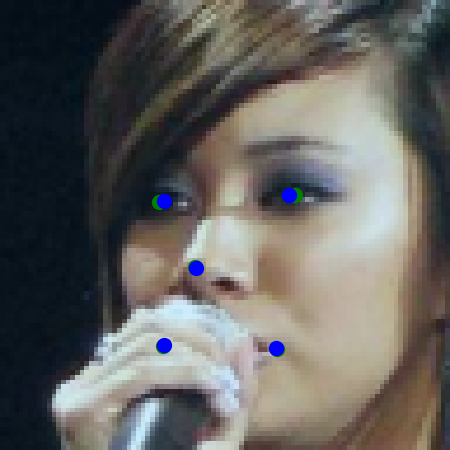}
    & \includegraphics[align=c, width=0.1\linewidth, height=0.1\linewidth]{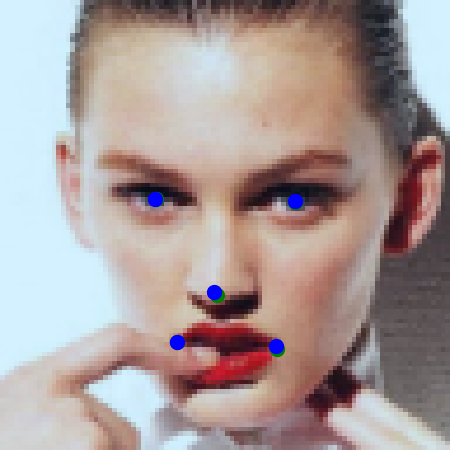}
    & \includegraphics[align=c, width=0.1\linewidth, height=0.1\linewidth]{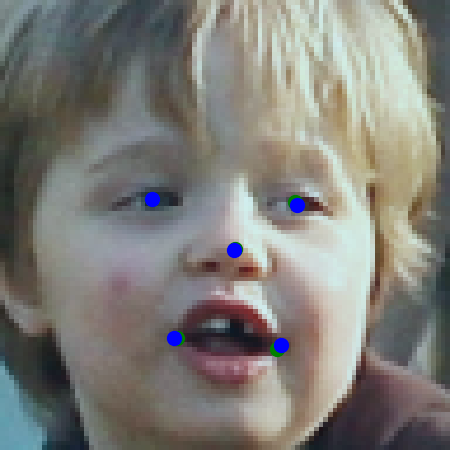} \rule[7ex]{0pt}{0pt}\rule[-6ex]{0pt}{0pt}\\
    \hline
    AFLW$_M$
    & \includegraphics[align=c, width=0.1\linewidth, height=0.1\linewidth]{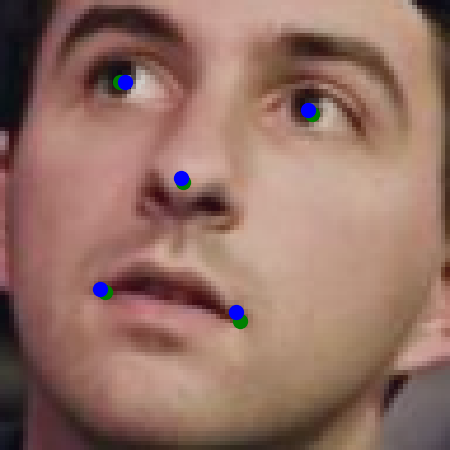}
    & \includegraphics[align=c, width=0.1\linewidth, height=0.1\linewidth]{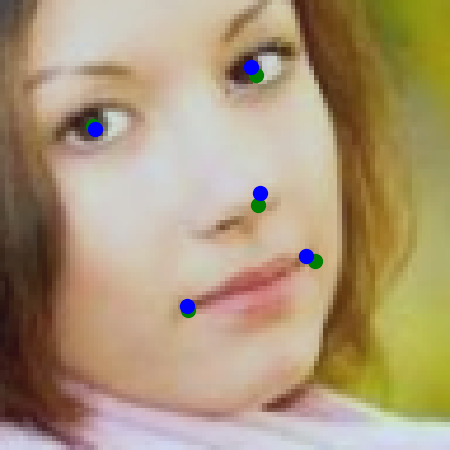}
    & \includegraphics[align=c, width=0.1\linewidth, height=0.1\linewidth]{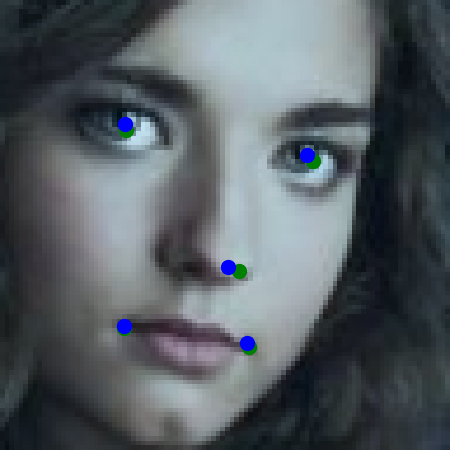}
    & \includegraphics[align=c, width=0.1\linewidth, height=0.1\linewidth]{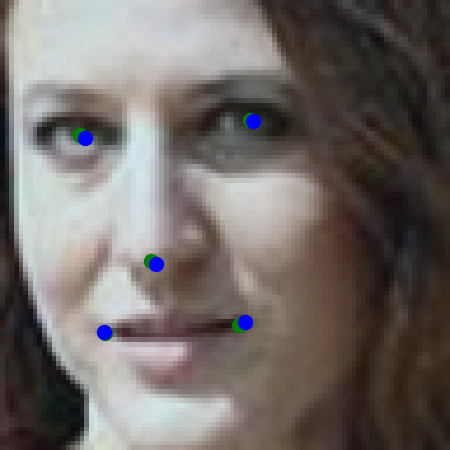}
    & \includegraphics[align=c, width=0.1\linewidth, height=0.1\linewidth]{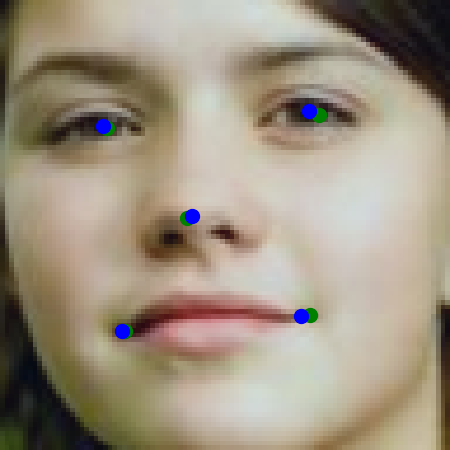}
    & \includegraphics[align=c, width=0.1\linewidth, height=0.1\linewidth]{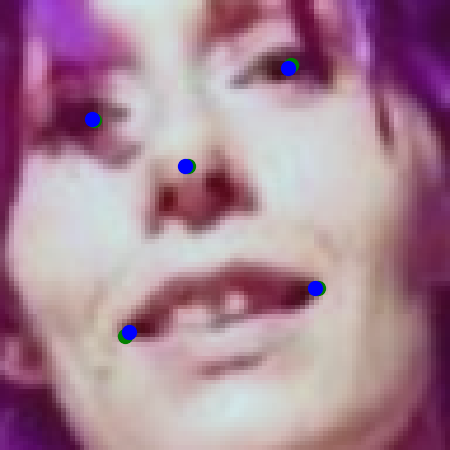}
    & \includegraphics[align=c, width=0.1\linewidth, height=0.1\linewidth]{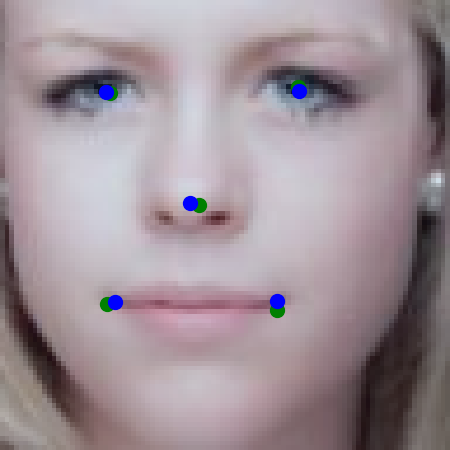} \rule[7ex]{0pt}{0pt}\rule[-6ex]{0pt}{0pt}\\
    \hline
    AFLW$_{RC}$
    & \includegraphics[align=c, width=0.1\linewidth, height=0.1\linewidth]{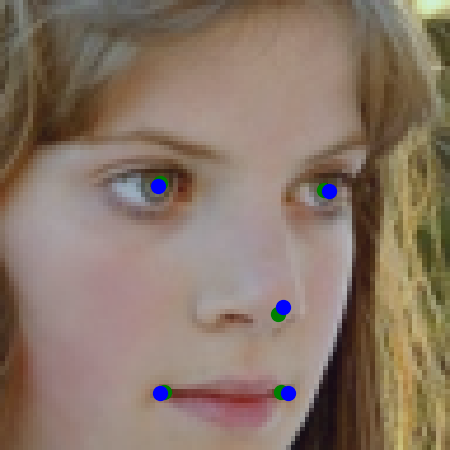}
    & \includegraphics[align=c, width=0.1\linewidth, height=0.1\linewidth]{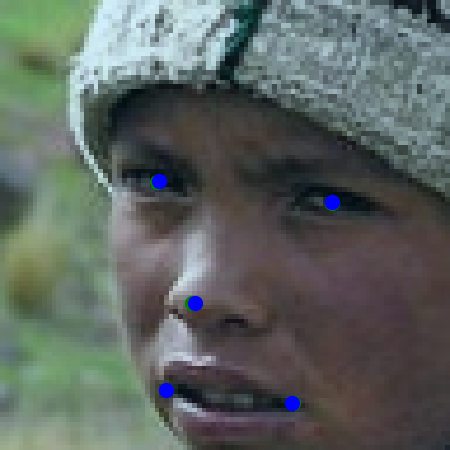}
    & \includegraphics[align=c, width=0.1\linewidth, height=0.1\linewidth]{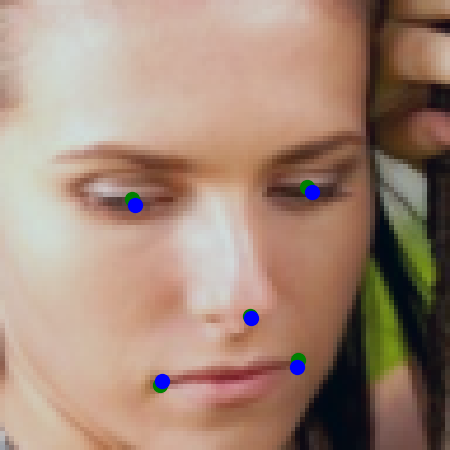}
    & \includegraphics[align=c, width=0.1\linewidth, height=0.1\linewidth]{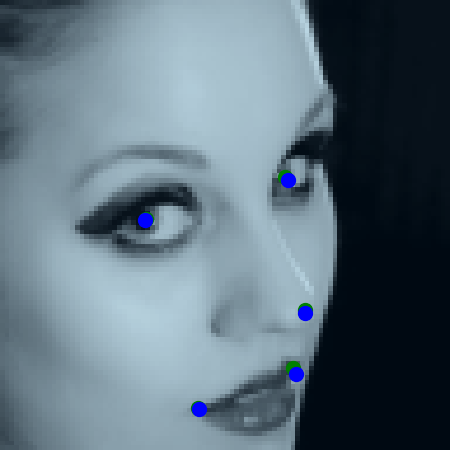}
    & \includegraphics[align=c, width=0.1\linewidth, height=0.1\linewidth]{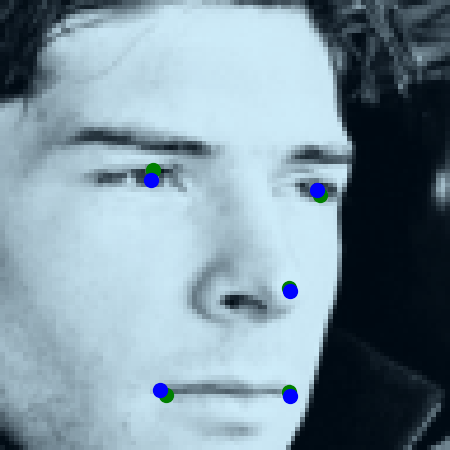}
    & \includegraphics[align=c, width=0.1\linewidth, height=0.1\linewidth]{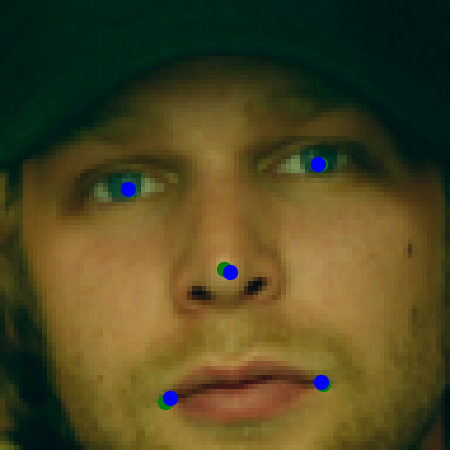}
    & \includegraphics[align=c, width=0.1\linewidth, height=0.1\linewidth]{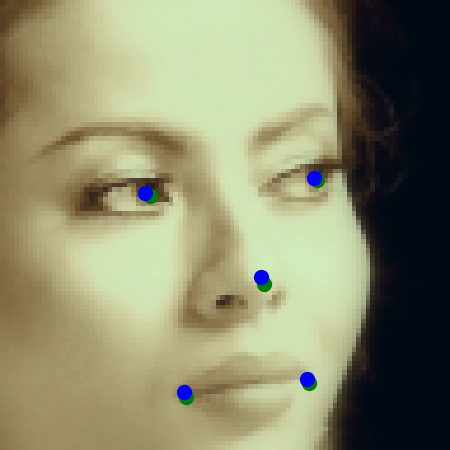} \rule[7ex]{0pt}{0pt}\rule[-6ex]{0pt}{0pt}\\
    \hline
    300W
    & \includegraphics[align=c, width=0.1\linewidth, height=0.1\linewidth]{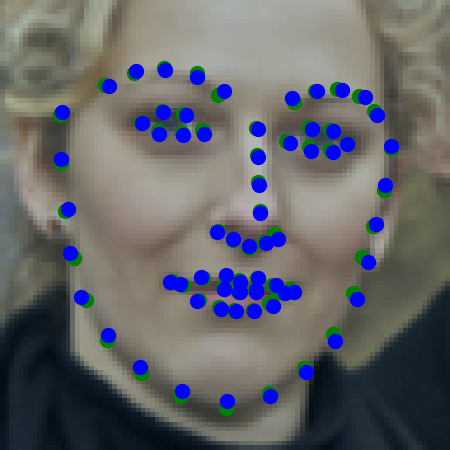}
    & \includegraphics[align=c, width=0.1\linewidth, height=0.1\linewidth]{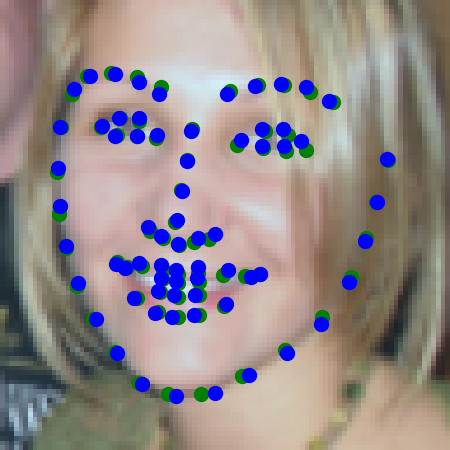}
    & \includegraphics[align=c, width=0.1\linewidth, height=0.1\linewidth]{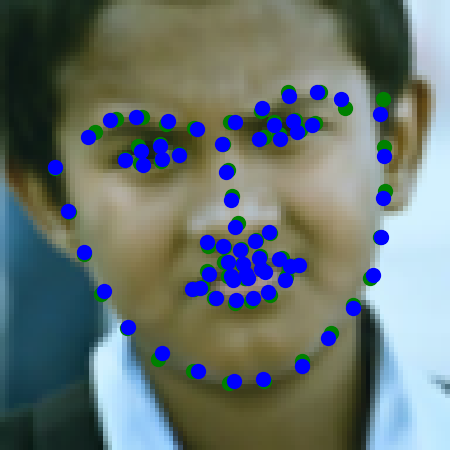}
    & \includegraphics[align=c, width=0.1\linewidth, height=0.1\linewidth]{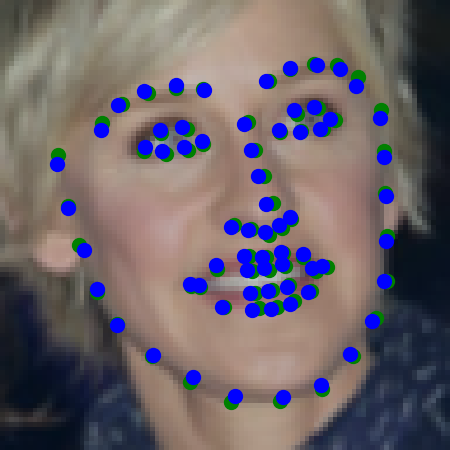}
    & \includegraphics[align=c, width=0.1\linewidth, height=0.1\linewidth]{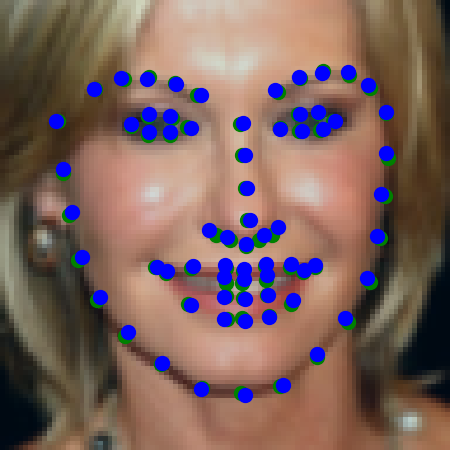}
    & \includegraphics[align=c, width=0.1\linewidth, height=0.1\linewidth]{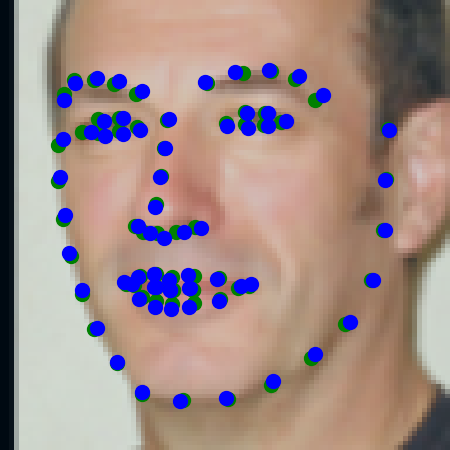}
    & \includegraphics[align=c, width=0.1\linewidth, height=0.1\linewidth]{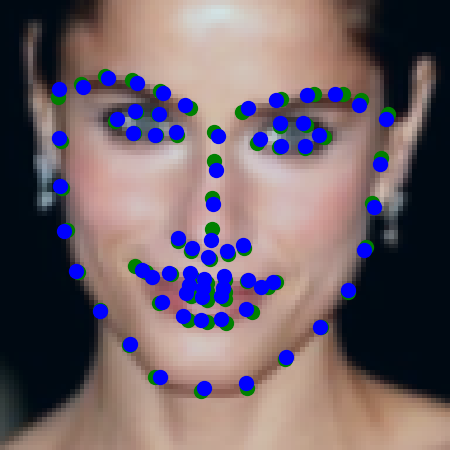} \rule[7ex]{0pt}{0pt}\rule[-6ex]{0pt}{0pt}\\
\end{tabular}
\centering
\caption{\textbf{Qualitative results on landmark detection.} The ground-truth and predictions are shown in green and blue dots respectively. In some cases we can only see blue dots because the prediction is almost/exact the same as ground-truth.}
\label{fig quantitative results on landmark detection}
\end{figure*}

Here, we show some qualitative results on landmark detection with our DeiT-B backbone in Figure \ref{fig quantitative results on landmark detection}. The model outputs accurate landmark prediction across four datasets. The model is also robust to different view angles and even some occlusions, e.g. the fifth image in MAFL.

\subsection{Failure Cases of Landmark Matching}
\begin{figure*}[!t]
\begin{tabular}{l|ccccc}
    Ref
    & \includegraphics[align=c, width=0.14\linewidth, height=0.14\linewidth]{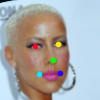}
    & \includegraphics[align=c, width=0.14\linewidth, height=0.14\linewidth]{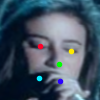}
    & \includegraphics[align=c, width=0.14\linewidth, height=0.14\linewidth]{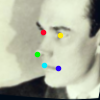}
    & \includegraphics[align=c, width=0.14\linewidth, height=0.14\linewidth]{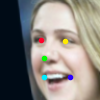}
    & \includegraphics[align=c, width=0.14\linewidth, height=0.14\linewidth]{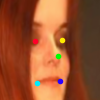} \rule[9.5ex]{0pt}{0pt}\rule[-8.5ex]{0pt}{0pt}\\
    \hline
    CL
    & \includegraphics[align=c, width=0.14\linewidth, height=0.14\linewidth]{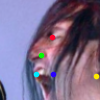}
    & \includegraphics[align=c, width=0.14\linewidth, height=0.14\linewidth]{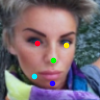}
    & \includegraphics[align=c, width=0.14\linewidth, height=0.14\linewidth]{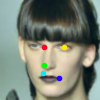}
    & \includegraphics[align=c, width=0.14\linewidth, height=0.14\linewidth]{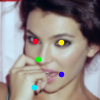}
    & \includegraphics[align=c, width=0.14\linewidth, height=0.14\linewidth]{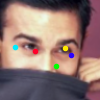} \rule[9.5ex]{0pt}{0pt}\rule[-8.5ex]{0pt}{0pt}\\
    LEAD
    & \includegraphics[align=c, width=0.14\linewidth, height=0.14\linewidth]{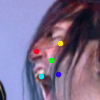}
    & \includegraphics[align=c, width=0.14\linewidth, height=0.14\linewidth]{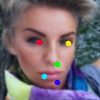}
    & \includegraphics[align=c, width=0.14\linewidth, height=0.14\linewidth]{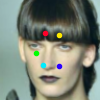}
    & \includegraphics[align=c, width=0.14\linewidth, height=0.14\linewidth]{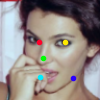}
    & \includegraphics[align=c, width=0.14\linewidth, height=0.14\linewidth]{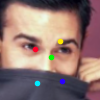} \rule[9.5ex]{0pt}{0pt}\rule[-8.5ex]{0pt}{0pt}\\
    Ours (DeiT-S)
    & \includegraphics[align=c, width=0.14\linewidth, height=0.14\linewidth]{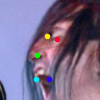}
    & \includegraphics[align=c, width=0.14\linewidth, height=0.14\linewidth]{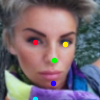}
    & \includegraphics[align=c, width=0.14\linewidth, height=0.14\linewidth]{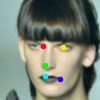}
    & \includegraphics[align=c, width=0.14\linewidth, height=0.14\linewidth]{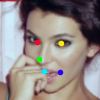}
    & \includegraphics[align=c, width=0.14\linewidth, height=0.14\linewidth]{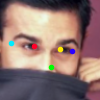} \rule[9.5ex]{0pt}{0pt}\rule[-8.5ex]{0pt}{0pt}\\
    Ours (DeiT-B)
    & \includegraphics[align=c, width=0.14\linewidth, height=0.14\linewidth]{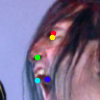}
    & \includegraphics[align=c, width=0.14\linewidth, height=0.14\linewidth]{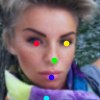}
    & \includegraphics[align=c, width=0.14\linewidth, height=0.14\linewidth]{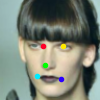}
    & \includegraphics[align=c, width=0.14\linewidth, height=0.14\linewidth]{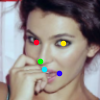}
    & \includegraphics[align=c, width=0.14\linewidth, height=0.14\linewidth]{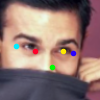} \rule[9.5ex]{0pt}{0pt}\rule[-8.5ex]{0pt}{0pt}\\
    \hline
    GT
    & \includegraphics[align=c, width=0.14\linewidth, height=0.14\linewidth]{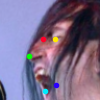}
    & \includegraphics[align=c, width=0.14\linewidth, height=0.14\linewidth]{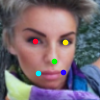}
    & \includegraphics[align=c, width=0.14\linewidth, height=0.14\linewidth]{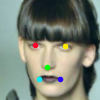}
    & \includegraphics[align=c, width=0.14\linewidth, height=0.14\linewidth]{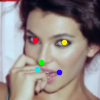}
    & \includegraphics[align=c, width=0.14\linewidth, height=0.14\linewidth]{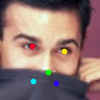} \rule[9.5ex]{0pt}{0pt}\rule[-8.5ex]{0pt}{0pt}\\
\end{tabular}
\centering
\caption{Failure cases of landmark matching.}
\label{fig failure cases of landmark matching}
\end{figure*}

Here we visualize some failure cases of landmark matching in Figure \ref{fig failure cases of landmark matching}. We find the main reason for these failure cases is occlusion. In some cases we can only see one side of the person's face in the image, thus the query or ground-truth landmark is occluded by other face parts. There are also cases where the landmarks are directly occluded by hand or cloth. In these cases, the query/test pixel representation at the landmark location may not effectively represent the landmark which leads to failure matching results. \vfill

\section{Trainable Components for Each Stage}
Our proposed method involves two training stages and the evaluation protocols for two downstream tasks are different as well. To offer a better understanding of how our framework is trained and evaluated, we detailed the trainable components for each stage and task in Table \ref{table supple. trainable components}. Note that \emph{only the component listed in the table is trained in the corresponding stage}, e.g., the Backbone (DeiT) is only trained in stage 1 and is frozen in all other stages. 

\begin{table}[t]
\caption{\textbf{Trainable components for each training stage and evaluation task.}}
\centering
\label{table supple. trainable components}
\begin{tabular}{c|ccc}
\hline
   & Stage 1       & Stage 2    & Evaluation               \\
\hline
Landmark Matching    & \multirow{2}{*}{Backbone} & \multirow{2}{*}{Projector} & None \\\cline{1-1} \cline{4-4}
Landmark Detection   &                       &                           & Regressor \\
\hline
\end{tabular}
\end{table}

\section{Limitations and Future Work}
In this work, we presented a two-stage framework to address self-supervised face landmark estimation tasks. Despite the significant performance gain, there are still some limitations of the proposed method. Firstly, the use of the cover-and-stride technique to expand feature map resolution and produce more fine-grained representations requires additional forward passes during inference. Secondly, our second stage refining relies on the similarity map generated by the CLS token. The CLS token may be distracted when there are other salient objects in the given image. However, since our method operates on face crops, the dependency on the CLS token is mostly offloaded onto the face crop generation algorithm. Considering the limitations above, future work may involve exploring more efficient methods to gain high-resolution fine-grained feature representation and more reliable algorithms to separate the landmark and non-landmark regions. We hope our work will inspire more research in this field.

\end{document}